\RequirePackage[l2tabu,orthodox]{nag}

\documentclass[headsepline,footsepline,footinclude=false,oneside,fontsize=11pt,paper=a4,listof=totoc,bibliography=totoc,DIV=12]{scrbook} 

\PassOptionsToPackage{table,svgnames,dvipsnames}{xcolor}

\usepackage[T1]{fontenc}
\usepackage[sc]{mathpazo}
\usepackage[ngerman,english]{babel} 
\usepackage[autostyle]{csquotes}
\usepackage[%
  backend=biber,
  url=true,
  natbib=true,
  style=alphabetic, 
  sorting=none, 
  maxnames=1,
  minnames=1,
  maxbibnames=99,
  useprefix=true,
  giveninits,
  uniquename=init]{biblatex} 
\usepackage{graphicx}
\usepackage{scrhack} 
\usepackage{listings}
\usepackage{lstautogobble}
\usepackage{tikz}
\usepackage{pgfplots}
\usepackage{pgfplotstable}
\usepackage{booktabs} 
\usepackage[final]{microtype}
\usepackage{caption}
\usepackage[hidelinks]{hyperref} 
\usepackage{ifthen} 
\usepackage{pdftexcmds} 
\usepackage{paralist} 
\usepackage{subcaption}
\usepackage{siunitx} 
\usepackage{multirow} 
\usepackage{array} 
\usepackage{makecell} 
\usepackage{pdfpages} 
\usepackage{adjustbox} 
\usepackage{tablefootnote} 
\usepackage{threeparttable} 

\usepackage{amssymb}
\usepackage{pifont}

\usepackage[acronym,xindy,toc]{glossaries} 
\makeglossaries

\newglossaryentry{r0}
{
  name=\glslink{R0}{\ensuremath{R_{0}}},
  text=F\"{o}rster distance,
  description={F\"{o}rster distance, where 50\% ...}, 
  sort=R
}

\newglossaryentry{kdeac}
{
  name=\glslink{R0}{\ensuremath{k_{DEAC}}},
  text=$k_{DEAC}$, 
  description={is the rate of deactivation from ... and emission)}, 
  sort=k
}

\newacronym[shortplural={TUM}, longplural={Technical University of Munich}]{tuma}{TUM}
{Technical University of Munich}

\newglossaryentry{tum}
{
  name = TUM,
  description = {A university in Germany, Bavaria, in the city of Munich},
  plural = TUM
}

\newglossaryentry{computer}
{
  name=computer,
  description={is a programmable machine that receives input,
               stores and manipulates data, and provides
               output in a useful format}
}

\newacronym{ebm}{EBM}{Energy-based model}
\newacronym{jem}{JEM}{Joint Energy-based model}
\newacronym{epn}{EPN}{Energy-Prior Network}
\newacronym{ood}{OOD}{out-of-distribution}
\newacronym{id}{ID}{in-distribution}
\newacronym{sgld}{SGLD}{Stochastic Gradient Langevin Dynamics}
\newacronym{pgd}{PGD}{Projected Gradient Descent}
\newacronym{fgm}{FGM}{Fast Gradient Method}
\newacronym{vera}{VERA}{Variational Entropy Regularized Approximate Maximum Likelihood}
\newacronym{kl}{KL divergence}{Kullback-Leibler divergence}
\newacronym{nce}{NCE}{Noise Contrastive Estimation}
\newacronym{ece}{ECE}{Expected Calibration Error}
\newacronym{rbf}{RBF}{Radial Basis Function}
\newacronym{mcmc}{MCMC}{Markov Chain Monte Carlo}
\newacronym{aucpr}{AUC-PR}{Area under the Precision-Recall curve}
\newacronym{dpn}{DPN}{Dirchlet Prior Network}
\newacronym{ssm}{SSM}{Sliced Score Matching}
\newacronym{msp}{MSP}{Maximum Softmax Probability}
\newacronym{bnn}{BNN}{Bayesian Neural Network}
\newacronym{mlp}{MLP}{Multi-Layer Perceptron}
\newacronym{cd}{CD}{Contrastive Divergence}

\bibliography{library}

\setkomafont{disposition}{\normalfont\bfseries} 
\BeforeTOCHead[toc]{{\cleardoublepage\pdfbookmark[0]{\contentsname}{toc}}}

\definecolor{TUMBlue}{HTML}{0065BD}
\definecolor{TUMSecondaryBlue}{HTML}{005293}
\definecolor{TUMSecondaryBlue2}{HTML}{003359}
\definecolor{TUMBlack}{HTML}{000000}
\definecolor{TUMWhite}{HTML}{FFFFFF}
\definecolor{TUMDarkGray}{HTML}{333333}
\definecolor{TUMGray}{HTML}{808080}
\definecolor{TUMLightGray}{HTML}{CCCCC6}
\definecolor{TUMAccentGray}{HTML}{DAD7CB}
\definecolor{TUMAccentOrange}{HTML}{E37222}
\definecolor{TUMAccentGreen}{HTML}{A2AD00}
\definecolor{TUMAccentLightBlue}{HTML}{98C6EA}
\definecolor{TUMAccentBlue}{HTML}{64A0C8}

\pgfplotsset{compat=newest}
\pgfplotsset{
  cycle list={TUMBlue\\TUMAccentOrange\\TUMAccentGreen\\TUMSecondaryBlue2\\TUMDarkGray\\},
}

\graphicspath{
    {logos/}
    {figures/}
}

\newcolumntype{P}[1]{>{\centering\arraybackslash}p{#1}} 
\newcolumntype{M}[1]{>{\centering\arraybackslash}m{#1}} 
\newcolumntype{L}[1]{>{\raggedright\arraybackslash}m{#1}} 
\newcolumntype{R}[1]{>{\raggedleft\arraybackslash}m{#1}} 

\usepackage{cleveref}
\usepackage{amsthm}
\usepackage{amsmath}
\usepackage{nicefrac}
\usepackage{parskip}
\usepackage{mathtools}
\usepackage{xurl}
\newtheorem{theorem}{Theorem}
\newtheorem{lemma}{Lemma}

\newcommand{\params}{\boldsymbol{\theta}}
\newcommand{\umarginal}{\tilde{p}_{\params}(\mathbf{x})}
\newcommand{\predictivedist}{p(y \mid \mathbf{x}, \params{})}
\newcommand{\model}{f_{\params}({\mathbf{x})}}
\newcommand{\efunc}{E_{\params{}}}
\newcommand{\data}{\mathbf{x}}
\newcommand*{\getUniversity}{Technische Universität München}
\newcommand*{\getFaculty}{Department of Informatics}
\newcommand*{\getTitle}{Out-of-distribution Detection with Energy-based Models}
\newcommand*{\getTitleGer}{Anomalieerkennung mit Energy-based Models}
\newcommand*{\getAuthor}{Sven Elflein}
\newcommand*{\getDoctype}{Master Thesis in Informatics}
\newcommand*{\getSupervisor}{Prof. Dr. Stephan Günnemann}
\newcommand*{\getAdvisor}{M. Sc. Daniel Zügner, M. Sc. Bertrand Charpentier}
\newcommand*{\getSubmissionDate}{15. August 2021}
\newcommand*{\getSubmissionLocation}{Munich}

\begin{document}

\selectlanguage{english}

\pagenumbering{alph}
\begin{titlepage}
  \oddsidemargin=\evensidemargin\relax
  \textwidth=\dimexpr\paperwidth-2\evensidemargin-2in\relax
  \hsize=\textwidth\relax

  \centering

  \IfFileExists{logos/tum.pdf}{%
    \includegraphics[height=20mm]{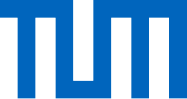}
  }{%
    \vspace*{20mm}
  }

  \vspace{5mm}
  {\huge\MakeUppercase{\getFaculty{}}}\\

  \vspace{5mm}
  {\large\MakeUppercase{\getUniversity{}}}\\

  \vspace{20mm}
  {\Large \getDoctype{}}

  \vspace{15mm}
  \makeatletter
  \ifthenelse{\pdf@strcmp{\languagename}{english}=0}
  {\huge\bfseries \getTitle{}}
  {\huge\bfseries \getTitleGer{}}
  \makeatother

  \vspace{15mm}
  {\LARGE \getAuthor{}}

  \IfFileExists{logos/faculty.png}{%
    \vfill{}
    \includegraphics[height=20mm]{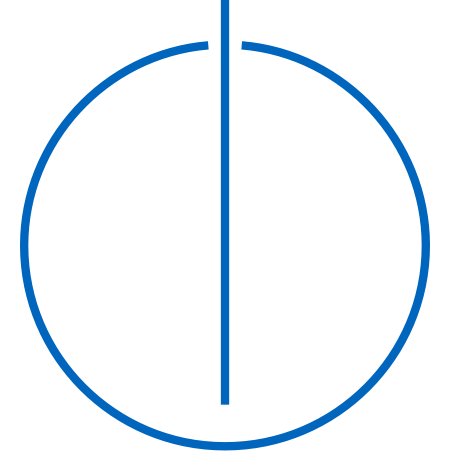}
  }{}
\end{titlepage}

\frontmatter{}

\begin{titlepage}
  \centering

  \IfFileExists{logos/tum.pdf}{%
    \includegraphics[height=20mm]{logos/tum.pdf}
  }{%
    \vspace*{20mm}
  }

  \vspace{5mm}
  {\huge\MakeUppercase{\getFaculty{}}}\\

  \vspace{5mm}
  {\large\MakeUppercase{\getUniversity{}}}\\

  \vspace{20mm}
  {\Large \getDoctype{}}

  \makeatletter
  \vspace{15mm}
  \ifthenelse{\pdf@strcmp{\languagename}{english}=0}
  {
  {\huge\bfseries \getTitle{}}

  \vspace{10mm}
  {\huge\bfseries \foreignlanguage{ngerman}{\getTitleGer{}}}
  }
  {
  {\huge\bfseries \getTitleGer{}}

  \vspace{10mm}
  {\huge\bfseries \foreignlanguage{english}{\getTitle{}}}
  }
  \makeatother

  \vspace{15mm}
  \begin{tabular}{l l}
    Author:          & \getAuthor{} \\
    Supervisor:      & \getSupervisor{} \\
    Advisor:         & \getAdvisor{} \\
    Submission Date: & \getSubmissionDate{} \\
  \end{tabular}

  \IfFileExists{logos/faculty.png}{%
    \vfill{}
    \includegraphics[height=20mm]{logos/faculty.png}
  }{}
\end{titlepage}

\cleardoublepage{}

\thispagestyle{empty}
\vspace*{0.8\textheight}
\noindent
\makeatletter
\ifthenelse{\pdf@strcmp{\languagename}{english}=0}
{I confirm that this \MakeLowercase{\getDoctype{}} is my own work and I have documented all sources and material used.}
{Ich versichere, dass ich diese \getDoctype{} selbstständig verfasst und nur die angegebenen Quellen und Hilfsmittel verwendet habe.}
\makeatother

\vspace{15mm}
\noindent
\getSubmissionLocation{}, \getSubmissionDate{} \hspace{50mm} \getAuthor{}

\cleardoublepage{}

\chapter{\abstractname}

Today, deep learning is increasingly applied in security-critical situations such as autonomous driving and medical diagnosis. Despite its success, the behavior and robustness of deep networks are not fully understood yet, posing a significant risk.

In particular, researchers recently found that neural networks are overly confident in their predictions, even on data they have never seen before.
To tackle this issue, one can differentiate two approaches in the literature. One accounts for uncertainty in the predictions, while the second estimates the underlying density of the training data to decide whether a given input is close to the training data, and thus the network is able to perform as expected.

For the latter approach, \glspl{ebm} can be used as density estimators. \glspl{ebm} are freely specified by a single function mapping from the data domain to a single value. This allows the usage of existing, powerful classification architectures.

In this thesis, we investigate the capabilities of \glspl{ebm} at the task of fitting the training data distribution to perform detection of \gls{ood} inputs. For evaluation, we consider tabular as well as image data. We find that on most datasets, vanilla \glspl{ebm} do not inherently outperform other density estimators at detecting \gls{ood} data despite their flexibility. Thus, we additionally investigate the effects of supervision, dimensionality reduction, and architectural modifications on the performance of \glspl{ebm}. 

Further, we propose \acrfull{epn} which enables estimation of various uncertainties within an \gls{ebm} for classification, bridging the gap between two approaches for tackling the \gls{ood} detection problem, i.e., fitting the data distribution and uncertainty estimation. We achieve this by identifying a connection between the concentration parameters of the Dirichlet distribution of a \gls{dpn} and the joint energy in an \gls{ebm}. 
Additionally, this allows optimization without a held-out \gls{ood} dataset, which might not be available or costly to collect in some applications.
Finally, we empirically demonstrate that \gls{epn} is able to detect \gls{ood} inputs, datasets shifts, and adversarial examples while preserving good performance at the classification task. Theoretically, \gls{epn} offers favorable properties for the asymptotic case when inputs are far from the training data.

\makeatletter
\ifthenelse{\pdf@strcmp{\languagename}{english}=0}
{\renewcommand{\abstractname}{Kurzfassung}}
{\renewcommand{\abstractname}{Abstract}}
\makeatother

\chapter{\abstractname}

\begin{otherlanguage}{ngerman} 
Deep Learning wird heute zunehmend in sicherheitskritischen Situationen wie dem autonomen Fahren und der medizinischen Diagnose eingesetzt. Jedoch ist dessen Verhalten und Robustheit noch nicht vollständig verstanden, was ein erhebliches Risiko darstellt.

Erst kürzlich haben Forscher herausgefunden, dass neuronale Netze übermä{\ss}ige Konfidenz bei ihren Vorhersagen zeigen, selbst bei Daten, die sie zuvor noch nie gesehen haben.
Um dieses Problem zu lösen, werden in der Literatur zwei Ansätze unterschieden. Der eine berücksichtigt die Unsicherheit bei der Vorhersage eines neuronalen Netzes, während der Zweite eine Wahrscheinlichkeitsverteilung über die Trainingsdaten approximiert, um zu entscheiden, ob eine Eingabe ähnlich zu den Trainingsdaten ist und das Netz somit die erwartete Leistung erbringen kann.

Für den letzteren Ansatz können \acrfull{ebm} als Dichteschätzer verwendet werden. \glspl{ebm} werden frei durch eine einzelne Funktion spezifiziert, die von den Daten auf ein Skalar abbildet. Diese Flexibilität ermöglicht die Nutzung bestehender, leistungsfähiger Architekturen von neuronaler Netzen.

In dieser Arbeit untersuchen wir \glspl{ebm} und deren Fähigkeit der Erkennung von anomalen (\gls{ood}) Eingaben. Wir stellen fest, dass \glspl{ebm}, trotz ihrer Flexibilität, nicht per se besser als andere Dichteschätzer bei der Erkennung von \gls{ood}-Daten funktionieren. Daher untersuchen wir zusätzlich den Fall, dass annotierte Daten zur Verfügung stehen, sowie die Auswirkungen von Dimensionalitätsreduktion und Änderungen an der Netzwerk-Architektur auf die Leistung von \glspl{ebm}. 

Des Weiteren präsentieren wir \acrfull{epn}, das die Schätzung verschiedener Unsicherheiten innerhalb eines \gls{ebm} ermöglicht und damit eine Brücke zwischen zwei Ansätzen zur \gls{ood}-Erkennung schlägt. Wir erreichen dies, indem wir eine Verbindung zwischen den Konzentrationsparametern der Dirichlet-Verteilung und der Energy in einem \gls{ebm} identifizieren. 
Diese Erkenntnis ermöglicht die Optimierung ohne eines \gls{ood} Datensatzes, der in manchen Anwendungen nicht verfügbar oder teuer zu beschaffen wäre.
Schlussendlich zeigen wir mit Experimenten, dass \gls{epn} in der Lage ist, diverse \gls{ood}-Eingaben zu erkennen. Weiterhin zeigen wir, dass \gls{epn} günstige theoretische Eigenschaften für den asymptotischen Fall von Eingaben weit entfernt von den Trainingsdaten, aufweist.
\end{otherlanguage}

\makeatletter
\ifthenelse{\pdf@strcmp{\languagename}{english}=0}
{\renewcommand{\abstractname}{Abstract}}
{\renewcommand{\abstractname}{Kurzfassung}}
\makeatother
\microtypesetup{protrusion=false}
\tableofcontents{}
\microtypesetup{protrusion=true}

\mainmatter{}


\chapter{Introduction}\label{chapter:introduction}

In recent years, the number of applications leveraging machine learning has risen significantly. In particular, deep learning has become the standard for many applications, outperforming traditional machine learning by large margins in the sub-fields of computer vision~\cite{omahonyDeepLearningVs2020} and natural language processing~\cite{torfiNaturalLanguageProcessing2021}. Nowadays, security-critical applications such as autonomous driving and medical diagnosis~\cite{grigorescuSurveyDeepLearning2020,shenDeepLearningMedical2017} leverage neural networks. However, their behavior and robustness is not fully understood yet~\cite{zhangUnderstandingDeepLearning2021}, posing a significant risk. 
One particular finding is that deep learning models often assign confident predictions to data on which they were not initially trained~\cite{heinWhyReLUNetworks2019,lakshminarayananSimpleScalablePredictive2017}, i.e., they make wrong predictions with high confidence. The overconfidence hinders the adoption of these models in real-world applications since one cannot be sure whether to trust the predictions. 

Recent research aims to solve this problem: on the one hand, by accounting for uncertainty in the prediction of a neural network~\cite{galUncertaintyDeepLearning2016} and, on the other hand, by estimating the underlying density of the training data to decide whether a given input is likely under the training data distribution and therefore the model will perform according to its specification~\cite{ruffUnifyingReviewDeep2021}.

For the latter task, recent work proposes increasingly powerful density estimators, e.g., Normalizing Flows~\cite{rezendeVariationalInferenceNormalizing2016}. Normalizing Flows transform a base distribution with a series of invertible transformations allowing the computation of exact likelihoods. However, \citet{nalisnickDeepGenerativeModels2019, kirichenkoWhyNormalizingFlows2020} recently showed that Normalizing Flows trained to estimate the density of the training data, assign higher likelihood to other data, termed out-of-distribution (\gls{ood}), than the \gls{id} data. This renders Normalizing Flows unsuitable for the task of detecting inputs that the models cannot confidently predict. One particular issue could be that Normalizing Flows rely on transformations for which the determinant of the Jacobian is easy to compute to be tractable, restricting the expressiveness of individual transformations~\cite{verineExpressivityBiLipschitzNormalizing2021}. 

Another promising class of density estimators are \glspl{ebm}~\cite{lecunTutorialEnergyBasedLearning}. \glspl{ebm} are not restricted to invertible transformations since \glspl{ebm} use a single, flexible function mapping from the data domain to the scalar energy value. This allows the usage of existing, powerful neural network architectures without modification. To this end, \citet{grathwohlYourClassifierSecretly2020} demonstrate good performance at detecting \gls{ood} samples using \glspl{ebm} based on the finding that discriminative models can be interpreted as \glspl{ebm}.

Overall, these findings encourage that \glspl{ebm} might be better suited for the task of \gls{ood} detection. Therefore, we aim to leverage \glspl{ebm} in order to improve \gls{ood} detection in this work. Further, we aim to study the main factors facilitating superior OOD detection of EBMs versus other generative models such as Normalizing Flows.

\section{Problem statement}
\label{ch:introduction:sec:problem_statement}
\begin{figure}
    \centering
    \includegraphics{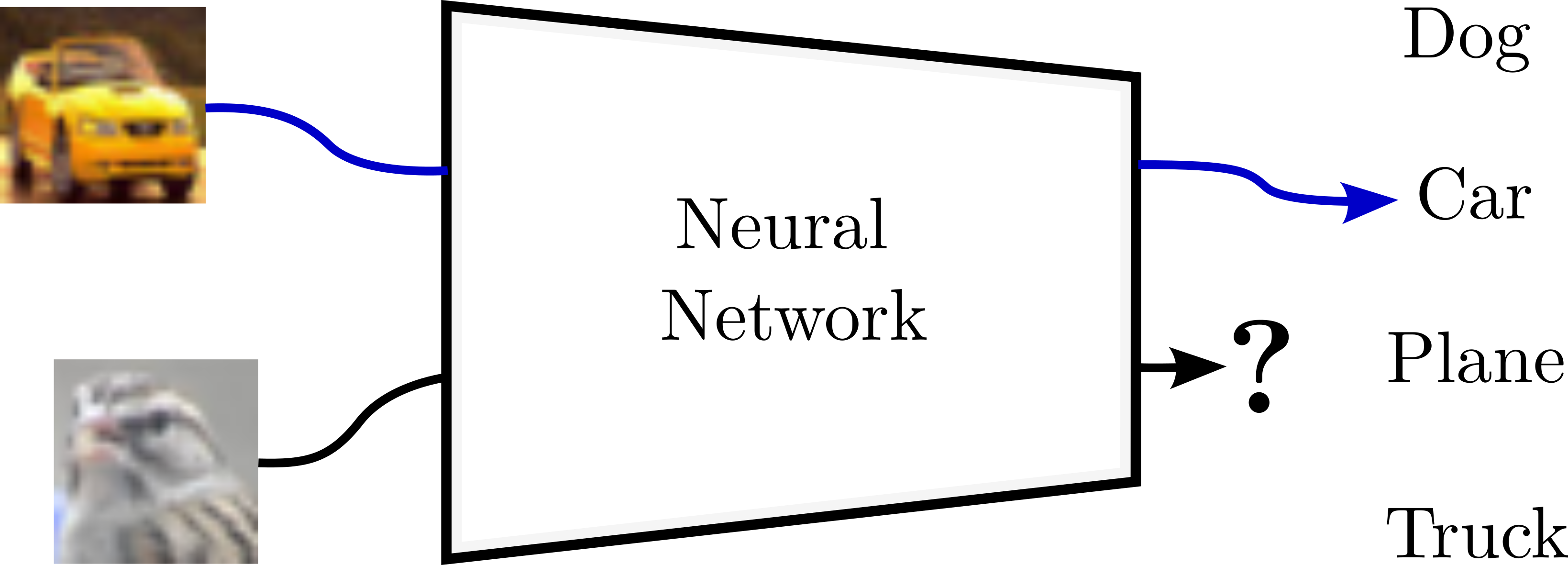}
    \caption{Inference in a classification model for an \gls{id} input (\textit{car}) and \gls{ood} input (\textit{bird}).}
    \label{ch:introduction:fig:motivation}
\end{figure}

Our focus in this thesis is on the classification setting. In more detail, we consider a neural network taking an input, e.g., an image, and classifying the input into a predefined set of classes. We illustratively visualize this setting in \Cref{ch:introduction:fig:motivation}. Here, the network is successfully able to predict the class of the image containing the car. 

However, the second image is not similar to the training data of the model. Intuitively, one would expect that in the case of an input that is not similar to any data the model has seen, the model would be maximally uncertain, i.e., it assigns every existing class the same probability. However, theoretical~\cite{heinWhyReLUNetworks2019} (for neural networks with ReLU activations) and empirical~\cite{lakshminarayananSimpleScalablePredictive2017,galDropoutBayesianApproximation2016} results show that the model assigns \textit{one of the existing classes} high probability. Thus, the model makes confident, wrong predictions which could lead to disastrous outcomes when deploying this model in security-critical applications.
The goal of \gls{ood} detection is to detect such inputs.

\section{Related Work}
As the main focus is reliable \gls{ood} detection, relevant prior work mainly revolves around the tasks of \gls{ood} detection and uncertainty estimation in the classification setting.  \paragraph{Uncertainty Estimation.} \citet{malininPredictiveUncertaintyEstimation2018, malininReverseKLDivergenceTraining2019} obtain uncertainty estimates for \gls{ood} detection by predicting parameters of a Dirichlet distribution for classification. \citet{charpentierPosteriorNetworkUncertainty2020} extended this approach to not rely on a separate \gls{ood} dataset by learning density-based pseudo counts. Other methods obtain uncertainty estimates based on a Bayesian framework by learning a distribution over model parameters given the data~\cite{blundellWeightUncertaintyNeural2015, maddoxSimpleBaselineBayesian2019, ritterScalableLaplaceApproximation2018,}, approximating that distribution with samples~\cite{lakshminarayananSimpleScalablePredictive2017} or with a variational approximation~\cite{galDropoutBayesianApproximation2016}.  Another line of approaches estimates uncertainty with a single deterministic neural network. \Citet{vanamersfoortUncertaintyEstimationUsing2020} leverage \gls{rbf} networks~\cite{broomheadRadialBasisFunctions1988} to provide uncertainty estimates by ensuring smoothness of the networks through a gradient penalty. Instead, \citet{liuSimplePrincipledUncertainty2020} use spectral normalization for enforcing smoothness and a Gaussian Process in the output layer enabling probabilistic reasoning. Finally, \citet{vanamersfoortFeatureCollapseDeep2021} improve this approach by using a point approximation of the Gaussian process~\cite{titsiasVariationalLearningInducing2009} enabling accuracy similar to modern discriminative methods.

\paragraph{Classifier-based \gls{ood} detection.}
\citet{hendrycksBaselineDetectingMisclassified2018} initially proposed to use the maximum softmax probability as \gls{ood} score. \citet{liangEnhancingReliabilityOutofdistribution2020, hsuGeneralizedODINDetecting2020} augment this approach by temperature scaling. Several methods add additional loss terms to the objective based on a separate \gls{ood} dataset~\cite{hendrycksDeepAnomalyDetection2019}, a generative model~\cite{leeTrainingConfidencecalibratedClassifiers2018, sricharanBuildingRobustClassifiers2018} or a uniformative noise distribution~\cite{heinWhyReLUNetworks2019} to encourage maximum entropy predictions for \gls{ood} inputs.

\paragraph{Density-based \gls{ood} detection.}
Other methods estimate the distribution over feature activations of a pre-trained network at multiple layers with class conditional Gaussian distributions~\cite{leeSimpleUnifiedFramework2018} or with Normalizing Flows~\cite{zisselmanDeepResidualFlow2020}.
Another set of methods aims to learn the density over the input data directly. \citet{nalisnickDeepGenerativeModels2019} discovered that the density learned by generative models cannot distinguish between in-distribution and \gls{ood} inputs but focus their analysis on flow-based models. Instead, \citet{renLikelihoodRatiosOutofDistribution2019} find that the likelihood in generative models is heavily affected by background statistics and instead propose a likelihood ratio-based score. Similarly,~\cite{serraInputComplexityOutofdistribution2020} find a significant influence of input complexity and propose a likelihood ratio based on a complexity estimate. \citet{nalisnickDetectingOutofDistributionInputs2019} discover a mismatch between the typical set and high-density regions which is tackled in \citet{choiWAICWhyGenerative2019} by estimating the epistemic uncertainty through an ensemble of generative models, in \citet{morningstarDensityStatesEstimation2020} through multiple summary statistics of the generative model, and in \citet{nalisnickDetectingOutofDistributionInputs2019} through a statistical test of typicality. However, these methods focus on flow-based and autoregressive density methods with tractable likelihood.

\paragraph{Energy-based models for \gls{ood} detection.}
Recently, there has been increasing interest in leveraging \glspl{ebm} as generative models for \gls{ood} detection.
\citet{duImplicitGenerationGeneralization2020} investigate the generative capabilities and generalization of \glspl{ebm} to \gls{ood} inputs based on approximate maximum-likelihood training.
\citet{zhaiDeepStructuredEnergy2016} introduces \gls{ebm} architectures for different types of data and a score matching objective for training \glspl{ebm} for anomaly detection.
\citet{grathwohlYourClassifierSecretly2020} and \citet{grathwohlNoMCMCMe2020} derive optimization procedures for hybrid \glspl{ebm} and investigate their \gls{ood} detection performance.

\paragraph{Differentiation to prior work.} 
In contrast to prior work, the research conducted in this thesis aims to connect the two aforementioned, seemingly unrelated research directions of \glspl{ebm} and Dirichlet-based models. While existing work on \glspl{ebm} reports great \gls{ood} detection results, it does not study the factors leading to improved OOD detection with \glspl{ebm} compared to other generative models. 

Most relevant to the latter is the work by \citet{kirichenkoWhyNormalizingFlows2020, schirrmeisterUnderstandingAnomalyDetection2020} which found that Normalizing Flows learn low-level features common to image datasets and thus struggle with detecting OOD inputs. We aim to provide similar insight for \glspl{ebm}.

\chapter{Background}\label{chapter:background}

In this section, we introduce key terminology and concepts which are required for comprehension of the work conducted in this thesis. Further, we expand on specific prior work and fundamental knowledge relevant for the proposed method introduced in \Cref{chapter:method} and comparisons in \Cref{ch:results}.

\section{Classification}
\label{ch:background:sec:classification}
In the classification setting, we consider a function \( f_{\params} : \mathbb{R}^D \mapsto \mathbb{R}^C \), which in our case is a neural network with parameters \(\params\), assigning real values, so called logits, for a fixed amount of classes \(C \in \mathbb{N}, C \geq 2\) given a datapoint \(\data \in \mathbb{R}^D\). 
The probabilities over the individual classes \(y \in \{1, \dots, C\}\) are defined using the softmax transfer function~\cite{Goodfellow-et-al-2016} as

\begin{equation}
    \label{ch:background:sec:classification:eq:predictive_distribution}
    p_{\params}(y \mid \data) = \frac{\exp(f_{\params}(\data)[y])}{\sum_{y^\prime} \exp(f_{\params}(\data)[y^\prime]}
\end{equation}

where we use \( f_{\params}(\data)[y] \) to denote the \(y\)-th logit. 

Traditionally, one uses a labeled dataset \(\mathcal{D} = \{(\data^{(i)}, y^{(i)})\}_{i=1}^N\) to optimize the parameters \(\params\) of the neural network \(f\) using the cross-entropy objective~\cite{Goodfellow-et-al-2016} 

\begin{equation}
    \min_{\params} \mathbb{E}_{(\data^{(i)}, y^{(i)}) \sim D} \left[- \sum_{c=1}^C  \delta_{y^{(i)}, c}  p_{\params}(c \mid \data) \right]
\end{equation}

where \(\delta\) denotes the Kronecker delta. 


\section{Uncertainty Estimation}
\label{ch:background:sec:uncertainty_estimation}

In security-critical applications, one often considers the notion of uncertainty. In the context of machine learning, uncertainty refers to the confidence of the model being correct. In particular, we expect that the model reports increasing uncertainty as the model performance degrades such that uncertainty is meaningful in practice.

For classification models describing a conditional density $p_{\params}(y \mid \data)$ over the classes $y$ given the data $\data$, uncertainty is divided into \textit{aleatoric} and \textit{epistemic} uncertainty.

\textit{Aleatoric uncertainty} or \textit{data uncertainty} refers to the uncertainty inherent to the data-generating process. Aleatoric uncertainty is \textit{irreducible}, i.e., collecting additional training data or increasing the capacity of the model does not reduce aleatoric uncertainty. Using the example of flipping a coin, we cannot predict the outcome of a coin toss with higher accuracy than $50\%$ - assuming a perfect coin - due to the inherent aleatoric uncertainty in the outcome.

On the other hand, \textit{epistemic uncertainty} or \textit{model uncertainty} describes the uncertainty in the model parameters given the training dataset. Therefore, one can reduce epistemic uncertainty by collecting more data or increasing the model capacity.

\paragraph{Bayesian view on uncertainty.}
Current approaches use a Bayesian view on uncertainty estimation~\cite{malininPredictiveUncertaintyEstimation2018, galDropoutBayesianApproximation2016}. For this, consider a finite dataset \(\mathcal{D} = \{\mathbf{x}_i, y_i\}_{i=1}^N\) and a classification model \(p(y|\mathbf{x}, \mathcal{D})\).

One defines the aleatoric uncertainty as the uncertainty in the posterior distribution over classes given the parameters of the model \(\params \), while uncertainty in the posterior distribution over the model parameters \(\params\) given the finite dataset \(\mathcal{D}\) represents the epistemic uncertainty:

\begin{equation}
    \label{chapter:background:eq:bayesian_uncertainty}
    p(y\mid \data, \mathcal{D}) = \int \underbrace{p(y|\data, \params)}_{\text{Data}} \underbrace{p(\params \mid \mathcal{D})}_{\text{Model}} d \params
\end{equation}

Since the true posterior distribution over parameters \(p(\params \mid \mathcal{D})\) is intractable, various approximations \(q(\params)\) have been proposed. \glspl{bnn}~\cite{blundellWeightUncertaintyNeural2015, maddoxSimpleBaselineBayesian2019, ritterScalableLaplaceApproximation2018} allow estimating a full distribution over weights, however, \glspl{bnn} are typically restricted to simple architectures. Two popular approaches which scale to complex tasks and show good performance in practice are Monte Carlo Dropout~\cite{galDropoutBayesianApproximation2016} and Deep Ensembles~\cite{lakshminarayananSimpleScalablePredictive2017}.

\subsection{Monte Carlo Dropout}
\citet{galDropoutBayesianApproximation2016} propose to use Dropout~\cite{srivastavaDropoutSimpleWay2014} in order to obtain a variational approximation of the true weight posterior \(p(\params \mid \mathcal{D}) \). In more detail, they define their approximation \( q(\params)\) as

\begin{align}
    W_i &= M_i \cdot \mathrm{diag}(\mathbf{z}_i) \\
    \mathbf{z}_{i} &\sim \mathrm{Bernoulli}(p_i) \quad \text{for} \quad i = 1, \dots, L
\end{align}

with \(\params = \{ \mathbf{W}_i \}_{i=1}^{L} \), the Dropout probability \( p_i \) and weight matrices \(M_i\) as variational parameters.

\citet{galDropoutBayesianApproximation2016} obtain the predictive distribution by plugging \(q(\params)\) into \Cref{chapter:background:eq:bayesian_uncertainty}, yielding

\begin{equation}
    q(y \mid \mathbf{x}) = \int p(y \mid  \mathbf{x}, \params) q(\params) d \params.
\end{equation}

The epistemic uncertainty is estimated as the empirical variance of \(q(y \mid \mathbf{x}) \) given by \(T\) stochastic forward passes with realizations of the Dropout configurations \( \{\mathbf{z}_{1}^t, \dots ,\mathbf{z}_{L}^t\}_{t=1}^T \).

\subsection{Deep Ensemble}
With a similar goal, \citet{lakshminarayananSimpleScalablePredictive2017} train an ensemble of neural networks to estimate uncertainty. From a probabilistic view, the ensemble approximates \(p(\params| \mathcal{D})\) through samples. Similar to Monte Carlo Dropout, \citet{lakshminarayananSimpleScalablePredictive2017} define the expected predictive distribution as 

\begin{equation}
    p(y \mid \mathbf{x}, \mathcal{D}) \approx M^{-1} p(y \mid \mathbf{x}, \params{}_m)
\end{equation}

where \(\{\params{}_i\}_{i=1}^{M}\) are the parameters of the \(M\) individual models of the ensemble. Deep Ensembles leverage the \gls{msp} of the expected predictive distribution \(\max p(y \mid \mathbf{x})\) to measure uncertainty.

\subsection{Dirichlet-based Models}
\label{ch:background:sec:uncertainty_estimation:subsec:dirichlet_based_models}

\citet{malininPredictiveUncertaintyEstimation2018} propose to decompose the \textit{epistemic uncertainty} by introducing \textit{distributional uncertainty}, which emerges due to mismatch between train and test data distributions. This decomposition allows to further differentiate between the sources of uncertainty: The \textit{aleatoric} uncertainty captures the uncertainty in situations of class overlap or noise, e.g., semantically similar classes such as cars and trucks, which means that the model knows both classes and is uncertain about the actual class (\textit{known-unknown}). On the other hand, \textit{distributional uncertainty} reflects an \textit{unknown-unknown}, e.g., the input does not belong to any of the classes of the training dataset, i.e., \gls{ood} inputs.

\citet{malininPredictiveUncertaintyEstimation2018} capture distributional uncertainty through a prior over categorical distributions. Incorporating the additional prior leads to the following predictive distribution:

\begin{equation}
    \label{ch:background:sec:dirichlet_methods:eq:predictive_distribution}
    p(y \mid \mathbf{x}, \mathcal{D}) = \int \int p(y|\boldsymbol{\mu}) p(\boldsymbol{\mu} \mid \mathbf{x}, \params) p(\params \mid \mathcal{D}) d \boldsymbol{\mu} d \params
\end{equation}

\begin{figure}
    \begin{subfigure}[t]{.3\textwidth}
    	\centering
    	\includegraphics[width=.9\linewidth]{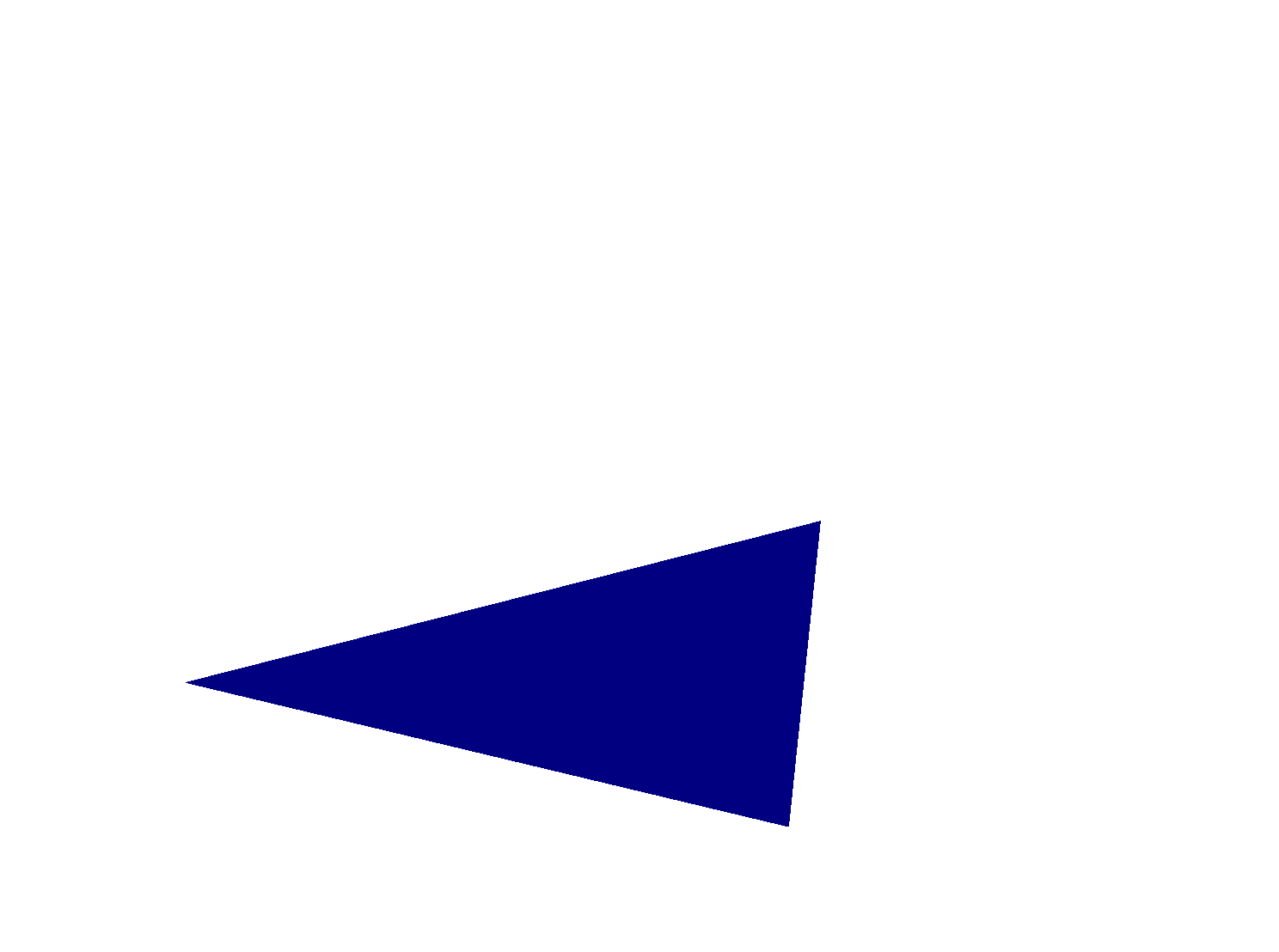}
    	\caption{Out-of-distribution}
    	\label{ch:background:sec:uncertainty_estimation:fig:flat_dirichlet}
    \end{subfigure}
    \hfill
    \begin{subfigure}[t]{.3\textwidth}
    	\centering
    	\includegraphics[width=.9\linewidth]{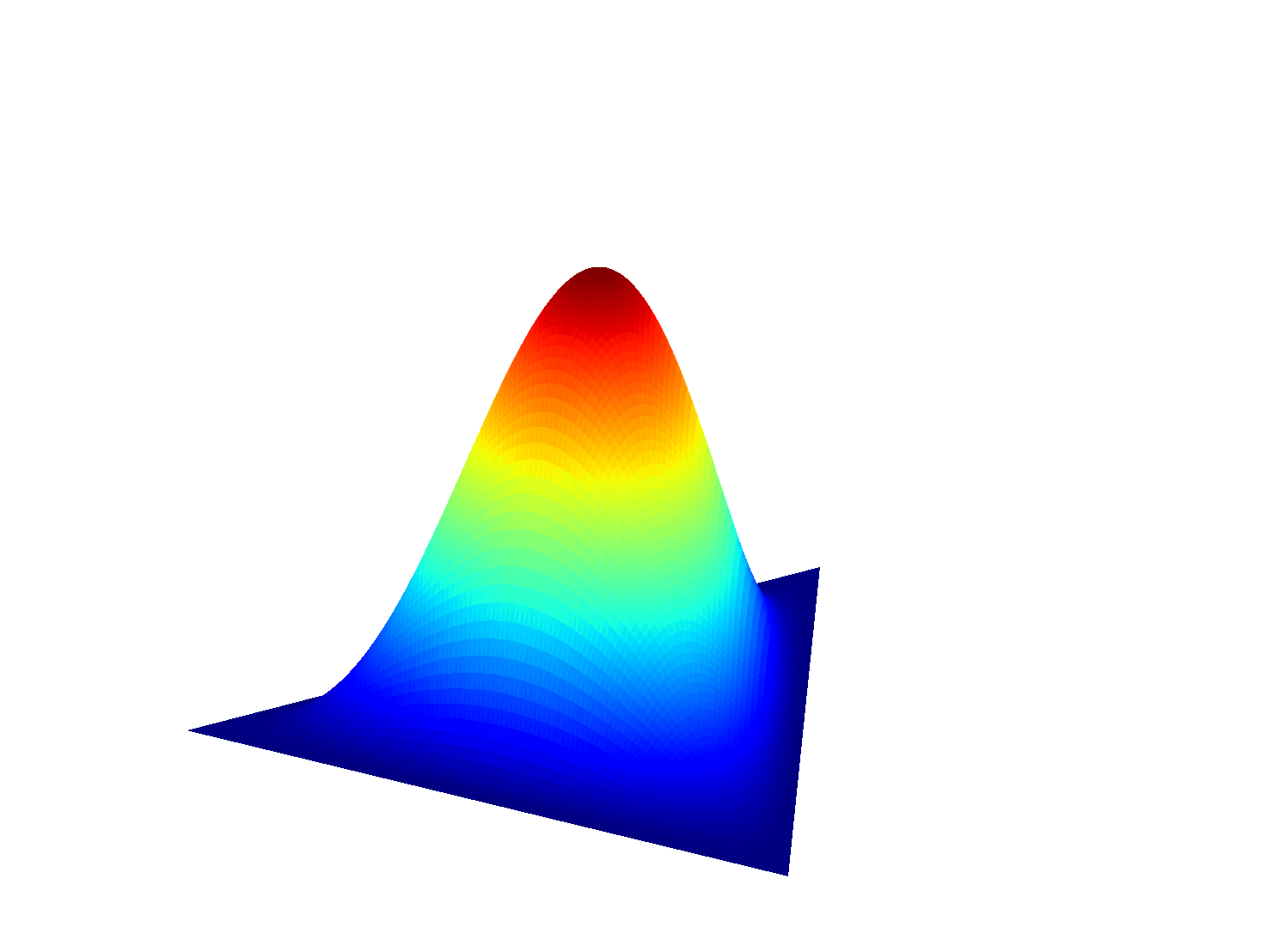}
    	\caption{High data uncertainty}
    	\label{ch:background:sec:uncertainty_estimation:fig:supported_dirichlet}
    \end{subfigure}
    \hfill
    \begin{subfigure}[t]{.3\textwidth}
    	\centering
    	\includegraphics[width=.9\linewidth]{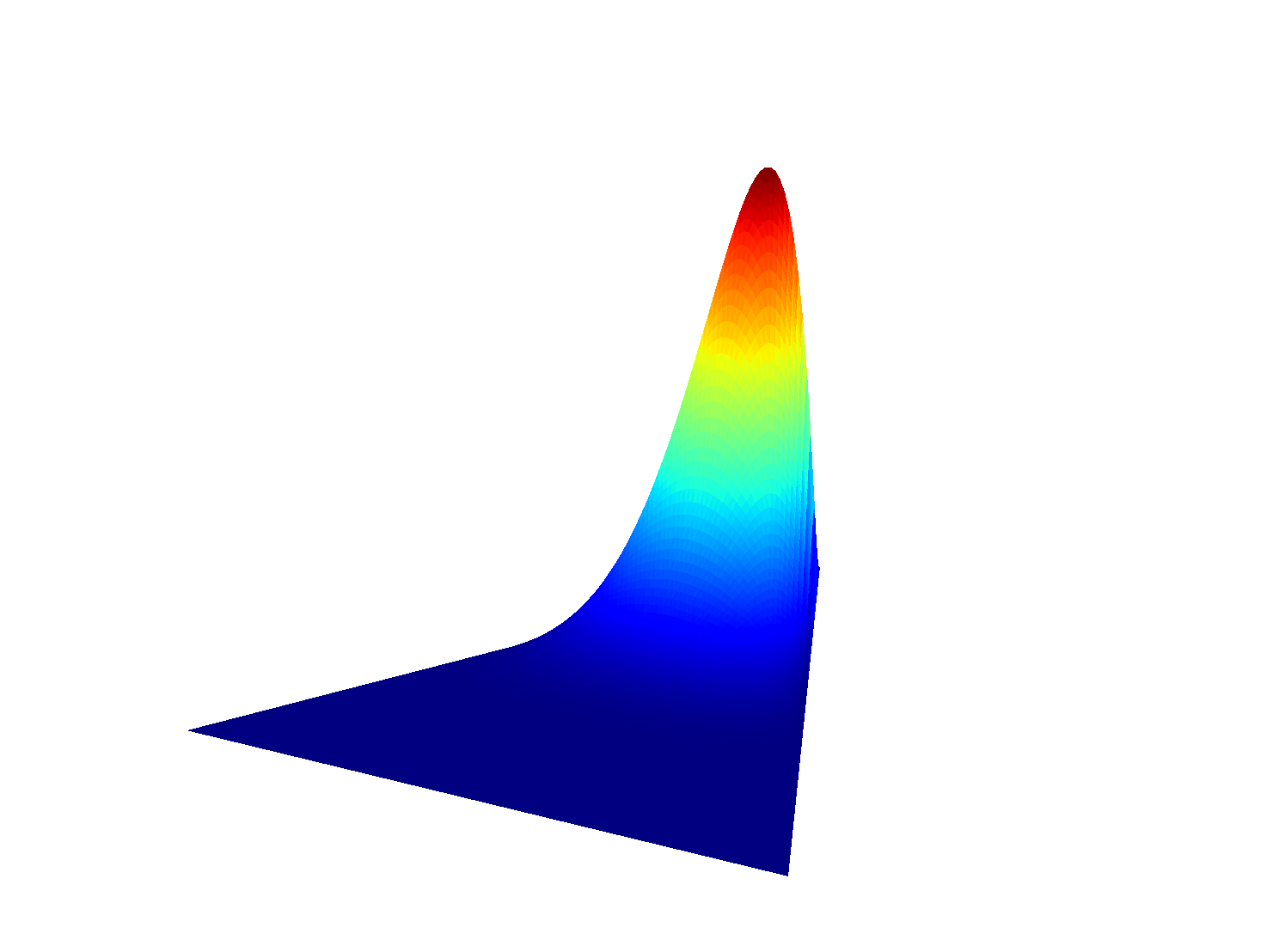}
    	\caption{Confident Prediction}
    	\label{ch:background:sec:uncertainty_estimation:fig:peaky_dirichlet}
    \end{subfigure}
    \caption{Posterior Dirichlet distributions \(p(\boldsymbol{\mu} \mid \mathbf{x}, \mathcal{D})\) with desired behavior.}
	\label{ch:background:sec:uncertainty_estimation:fig:dirichlet_vis}
\end{figure}

\paragraph{Dirichlet Prior Networks.}
A instance of models incorporating distributional uncertainty estimates at the task of classification are \glspl{dpn}~\cite{malininPredictiveUncertaintyEstimation2018, malininReverseKLDivergenceTraining2019}. They propose to parameterize the prior distribution \(p(\boldsymbol{\mu} \mid \mathbf{x}, \mathcal{D})\) with a Dirichlet distribution which is the conjugate prior~\cite{bishopPatternRecognitionMachine2006} to the categorical distribution and thus has nice analytic properties. A Dirichlet distribution is parameterized by a vector of concentration parameters \(\boldsymbol{\alpha} = [\alpha_1, \dots, \alpha_C]^T\).

\begin{equation}
\mathrm{Dir}(\boldsymbol{\mu} \mid \boldsymbol{\alpha}) = \frac{\Gamma(\alpha_0)}{\prod_{c=1}^{C} \Gamma(\alpha_c)} \prod_{c=1}^{C} \mu_c^{\alpha_c - 1}, \quad \alpha_C > 0
\end{equation}

where \(\alpha_0 = \sum_{c=1}^{C} \alpha_c\) is the \textit{precision} of the distribution and \(\Gamma\) denotes the Gamma function. Higher precision leads to sharper distributions as shown in \Cref{ch:background:sec:uncertainty_estimation:fig:dirichlet_vis}. 

In \gls{dpn}, one aims to obtain a flat Dirichlet distribution (\Cref{ch:background:sec:uncertainty_estimation:fig:flat_dirichlet}) where all categorical distributions are equiprobable (unknown-unknown) for \gls{ood} data, while for known-unknown inputs (aleatoric uncertainty) uniform categorical distributions should be more likely (\Cref{ch:background:sec:uncertainty_estimation:fig:supported_dirichlet}). Finally, in the case of a confident prediction with little uncertainty, the categorical distributions in the corner of the respective class should be likely under the Dirichlet prior as visualized in \Cref{ch:background:sec:uncertainty_estimation:fig:peaky_dirichlet}.

A DPN predicts the concentration parameters \(\boldsymbol{\alpha}\), parameterizing the Dirichlet prior as 

\begin{equation}
    p(\boldsymbol{\mu} \mid \mathbf{x}, \params) = \mathrm{Dir}(\boldsymbol{\mu} \mid \boldsymbol{\alpha} \\
     = \exp(f_{\params}(\mathbf{x}))
\end{equation}

where \(\mathbf{f}\) is a neural network. \citet{malininPredictiveUncertaintyEstimation2018} use the exponential as activation function to ensure \(\alpha_c > 0\).  

As a result, the posterior over classes is given by 

\begin{align}
    p(y_c \mid \mathbf{x}, \params) 
    &= \int p(y \mid \boldsymbol{\mu}) p(\boldsymbol{\mu} \mid \mathbf{x}, \params) d \boldsymbol{\mu} \\
    &= \frac{\alpha_c}{\alpha_0} \\
    &= \frac{\exp(f(\mathbf{x}; \params)[c])}{\sum_{c^\prime=1}^C \exp(f(\mathbf{x}; \params)[c^\prime])}
\end{align}

which is equivalent to the output of a neural network with Softmax activation. Thus, standard cross-entropy training, as described in \Cref{ch:background:sec:classification}, can be seen as optimizing the expected categorical distribution under a Dirichlet prior~\cite{malininPredictiveUncertaintyEstimation2018}. Unfortunately, the mean \(\frac{\alpha_c}{\alpha_0}\) is insensitive to arbitrary scaling of the concentration parameters, which means the precision \(\alpha_0\) and therefore uncertainty estimates are meaningless under cross-entropy training.

In order to achieve the expected behavior, \citet{malininPredictiveUncertaintyEstimation2018} propose to train the \gls{dpn} by minimizing the \gls{kl}~\cite{kullbackInformationSufficiency1951} between the predicted distribution and a sharp Dirichlet distribution with concentration parameters \(\hat{\boldsymbol{\alpha}}\) for the in-distribution dataset \(p_{in}\) and a flat Dirichlet with parameters \(\tilde{\mathbf{\alpha}} \) for out-of-distribution data \(p_{out}\) where all categorical distributions are equiprobable.

\begin{align}
    \label{ch:background:sec:dirichlet_methods:eq:priornet_objective}
    L(\params) = {}& \mathbb{E}_{p_{\mathrm{in}}(\mathbf{x})} \left[ D_{KL} \left[ \mathrm{Dir}(\boldsymbol{\mu} \mid \boldsymbol{\hat{\alpha}}) \mid\mid p(\boldsymbol{\mu} \mid \mathbf{x}, \params) \right]\right] \\
    &+ \mathbb{E}_{p_{\mathrm{out}}(\mathbf{x})} \left[ D_{KL} \left[ \mathrm{Dir}(\boldsymbol{\mu} \mid \boldsymbol{\tilde{\alpha}}) \mid\mid p(\boldsymbol{\mu} \mid \mathbf{x}, \params) \right]\right]
\end{align}

Training with \Cref{ch:method:sec:energy-priornet:eq:overall_objective} objective requires an \gls{ood} dataset \(p_{\mathrm{out}}\) which ideally covers the entire domain outside of the in-distribution dataset. Since this is unfeasible, \citet{malininPredictiveUncertaintyEstimation2018} propose to use a different, real dataset as \gls{ood} samples. Further, they fix the parameters of the target Dirichlet distributions \(\boldsymbol{\tilde{\alpha}} = [1, \dots, 1]\) and 

\begin{equation}
    \boldsymbol{\hat{\alpha}}_c = 
    \begin{cases}
        1,& \text{if } c \neq y, \\
        \alpha + 1, & \text{otherwise}
    \end{cases}
\end{equation}

where \(y\) is the index of the target class and \(\alpha\) is a manually tuned hyperparameter~\cite{malininReverseKLDivergenceTraining2019}.

Based on \Cref{ch:background:sec:dirichlet_methods:eq:predictive_distribution}, \citet{malininPredictiveUncertaintyEstimation2018} compute the \textit{Differential Entropy} of the Dirichlet distribution\footnote{Note that differential entropy is not the correct continuous analog to discrete entropy as it loses certain properties of discrete entropy~\cite{jaynesInformationTheoryStatistical1957}. We will use it since it has shown to yield consistent results~\cite{malininPredictiveUncertaintyEstimation2018}.} in order to measure \textit{distributional} uncertainty.

\begin{equation}
    \label{ch:background:sec:dirichlet_methods:eq:differential_entropy}
    \mathcal{H}[p(\boldsymbol{\mu} \mid \mathbf{x}, \mathcal{D})] = - \int_{\mathcal{S}^{C-1}} p(\boldsymbol{\mu} \mid \mathbf{x}, \mathcal{D}) \ln \left( p(\boldsymbol{\mu} \mid \mathbf{x}, \mathcal{D})\right) d\boldsymbol{\mu}
\end{equation}

Intuitively, differential entropy measures the flatness of the distribution and therefore indicates \gls{ood} samples in the \gls{dpn} framework.


\paragraph{Posterior Network.}
\citet{charpentierPosteriorNetworkUncertainty2020} use a Bayesian update of a categorical distribution based on a prior Dirichlet distribution \(\mathrm{Dir}(\boldsymbol{\beta}^{\mathrm{prior}})\) and learned pseudo counts \(\boldsymbol{\beta}\) to form the concentration parameters \(\boldsymbol{\alpha}\) of the Dirichlet distribution as 

\begin{equation}
    \boldsymbol{\alpha} = \boldsymbol{\beta}^{\mathrm{prior}} + \boldsymbol{\beta}
\end{equation}

Posterior Network defines the pseudo counts \(\boldsymbol{\beta} = [\beta_1, \dots, \beta_C]\) as

\begin{equation}
    \beta_c = N_c \cdot p(\boldsymbol{z}|c, \boldsymbol{\phi})
\end{equation}

where $N_c$ are the number of observations of class $c$ and \(p(\boldsymbol{z}|c, \boldsymbol{\phi})\) is the density of a Normalizing Flow with parameters \(\boldsymbol{\phi}\) at the position of the low-dimensional encoding \(\boldsymbol{z}\) which is the output of a encoder neural network \(f(\mathbf{x}, \params)\).

\subsection{Uncertainty measures.}
\label{ch:background:sec:uncertainty_estimation:subsec:uncertainty_measures}
Based on \Cref{chapter:background:eq:bayesian_uncertainty}, one can derive multiple quantities measuring the \textit{total}, \textit{aleatoric}, or \textit{epistemic} uncertainty.

To measure the overall uncertainty in the prediction, \citet{hendrycksBaselineDetectingMisclassified2018} use the \gls{msp} 

\begin{equation}
    \label{ch:background:sec:uncertainty_estimation:eq:msp}
    \max_y p(y \mid \mathbf{x}, \mathcal{D})
\end{equation}

and \citet{galDropoutBayesianApproximation2016} use the entropy of the predictive distribution

\begin{equation}
    \mathcal{H}(p(y \mid \mathbf{x}, \mathcal{D})) = -\sum_{c=1}^{C} p(c \mid \mathbf{x}, \mathcal{D}) \ln \left( p(c \mid \mathbf{x}, \mathcal{D})\right).
\end{equation}

For measuring \textit{model uncertainty}, \citet{depewegDecompositionUncertaintyBayesian2018} consider the \textit{Mutual Information (MI)}~\cite{shannonMathematicalTheoryCommunication1948} between the categorical label $t$ and the parameters of the model \(\params\) measuring the spread of an ensemble \(\{p(y \mid \mathbf{x}, \params_i)\}_{i=1}^M\)~\cite{galUncertaintyDeepLearning2016}. One can write the model uncertainty as the difference in total uncertainty and data uncertainty measured as the expected entropy of the members of the ensemble following~\cite{depewegDecompositionUncertaintyBayesian2018} as

\begin{equation}
    \label{chapter:background:eq:mutual_information_epistemic_uncertainty}
    \underbrace{\mathcal{I}[y, \params \mid \mathbf{x}, \mathcal{D}]}_{\text{Model Uncertainty}} = \underbrace{\mathcal{H}[\mathbb{E}_{p(\params \mid \mathcal{D})}[p(y \mid \mathbf{x}, \params)]]}_{\text{Total Uncertainty}} - \underbrace{\mathbb{E}_{p(\params \mid \mathcal{D})}[\mathcal{H}[p(y \mid \mathbf{x}, \params)] ]}_{\text{Data Uncertainty}}
\end{equation}

Further, one can consider another decomposition~\cite{malininPredictiveUncertaintyEstimation2018} based on marginalizing \(\params\) in \Cref{chapter:background:eq:bayesian_uncertainty} suitable for \gls{dpn} models

\begin{equation}
    \underbrace{\mathcal{I}[y, \boldsymbol{\mu} \mid \mathbf{x}, \mathcal{D}]}_{\text{Distributional Uncertainty}} = \underbrace{\mathcal{H}[\mathbb{E}_{p(\boldsymbol{\mu} \mid \data, \mathcal{D})}[p(y \mid \boldsymbol{\mu})]]}_{\text{Total Uncertainty}} - \underbrace{\mathbb{E}_{p(\boldsymbol{\mu} \mid \data, \mathcal{D})}[\mathcal{H}[p(y \mid \boldsymbol{\mu})] ]}_{\text{Data Uncertainty}}
\end{equation}

Here, the distributional uncertainty, as a part of epistemic uncertainty, is explicitly considered as the spread of categorical distributions under the Dirchlet prior similar to differential entropy in \Cref{ch:background:sec:dirichlet_methods:eq:differential_entropy}. For simplicity of notation, we will omit the dependence on the dataset \(\mathcal{D}\) in the following.

\section{Density Estimation}
\label{ch:background:sec:density_estimation}
Another approach to out-of-distribution detection is to estimate the underlying density of the training data \(p(\data)\). Then, one considers inputs with sufficiently low density to be \acrlong{ood}~\cite{pimentelReviewNoveltyDetection2014}.

Density estimation is the task of approximating the underlying distribution of a dataset while only observing samples from that particular distribution~\cite{silvermanDensityEstimationStatistics2017}. More formally, consider a finite set of data points \(\{\mathbf{x}^{(i)}\}_{i=1}^N, \mathbf{x}^{(i)} \in \mathbb{R}^D\) where the individual data points are sampled independently and identically from an underlying, unknown random variable \(\mathbf{x}_i \sim X\) with probability density function \(p(\data=\data)\). Then, the task is to recover a probability density function \(p_{\params}(\data)\) from the finite dataset \(\{\mathbf{x}^{(i)}\}_{i=1}^N\) which is close to the true \(p(\data)\).

Afterwards, we can evaluate the density \(p_{\params}(\hat{\data})\) for a particular, novel datapoint \(\hat{\data}\) together with a threshold to determine if that datapoint comes from the underlying distribution \(p(\data)\).

\begin{figure}[ht]
    \centering
    \includegraphics{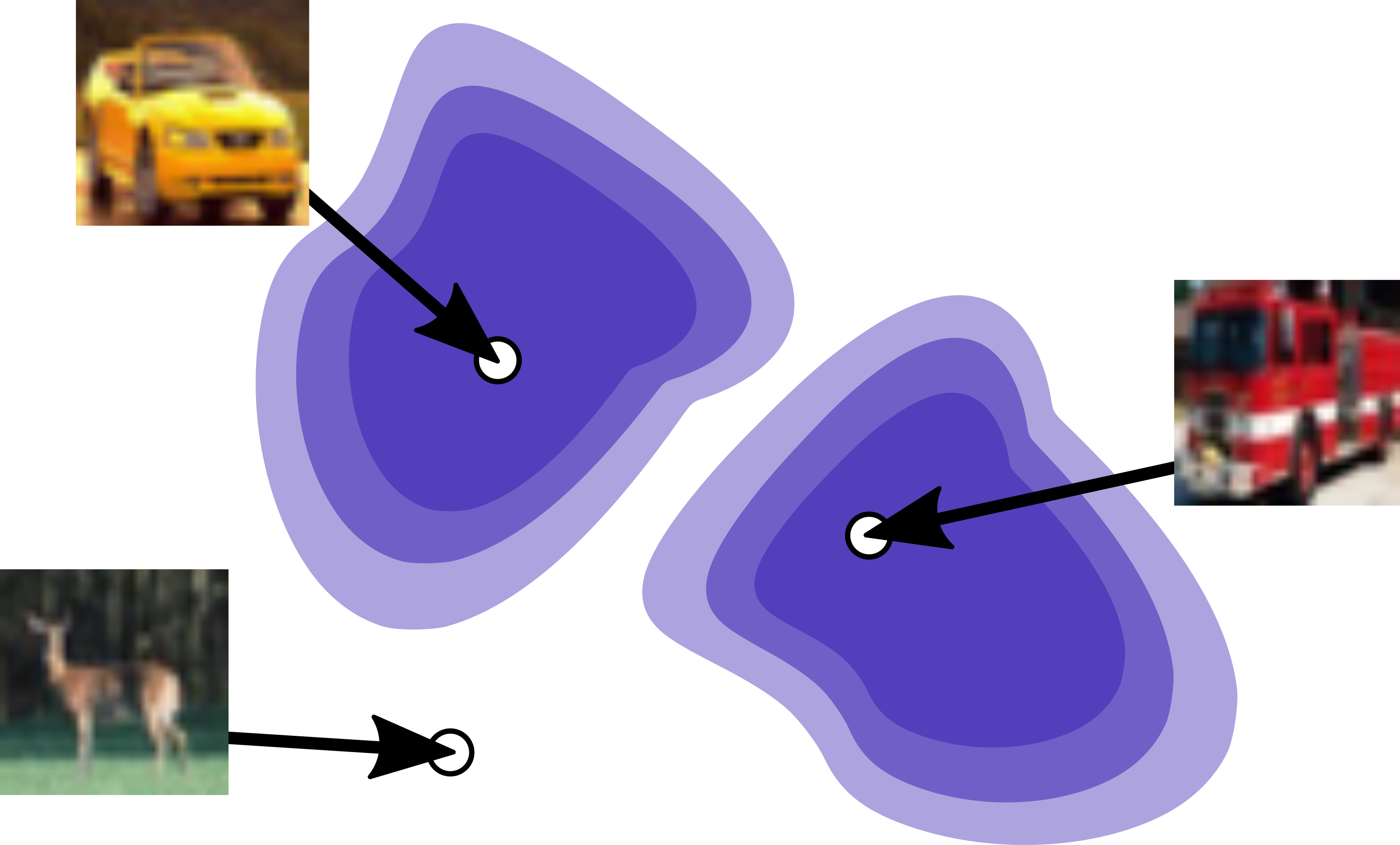}
    \caption{Example density estimate of dataset of images of cars and trucks. Darker shades of blue indicates higher density.}
    \label{ch:background:sec:density_estimation:fig:example_density}
\end{figure}

\Cref{ch:background:sec:density_estimation:fig:example_density} shows an illustrative example of a density estimate of a dataset of images containing cars and trucks. Consequently, regions of high density surround the samples of cars and trucks. On the other hand, samples not belonging to the dataset, e.g., the image of a deer, are assigned low density. Thus, one can differentiate between in-distribution data, i.e., cars and trucks, and out-of-distribution data by comparing the density evaluated at the position of the data points.

\subsection{Normalizing Flows}
\label{ch:background:sec:density_estimation:subsec:normalizing_flow}

A particular class of density estimators are Normalizing Flows introduced in \citet{rezendeVariationalInferenceNormalizing2016}. Normalizing Flows transform a base distribution \(p(\mathbf{z})\) into another distribution \(p(x)\) based on the change of variable formula 

\begin{equation}
    p_2(\data) = p_1(f^{-1}(\data)) \cdot \left| \det \left( \frac{\partial f^{-1}(\data)}{\partial \data} \right) \right |
\end{equation}

where $f: \mathbb{R}^D \mapsto \mathbb{R}^D, f(\mathbf{z}) = \data$ is an invertible and differentiable transformation.

These transformations can be stacked obtaining a series of differentiable and invertible transformations $f_1, \dots, f_n$ transforming a base density \(p_0(\mathbf{z}_0)\) into a potentially complex target density \(p_n(\data)\)

\begin{equation}
    \log p_n(\data) = p_0(\mathbf{z}_0) + \sum_{i=1}^{n} \log \left| \det \left( \frac{\partial f_i}{\partial \mathbf{z}_i} \right) \right| 
\end{equation}

where $\mathbf{z}_i = f_i(\mathbf{z}_{i-1})$.

In order to speed up the computation of the determinant, various transformations with triangular Jacobian structure have been proposed~\cite{rezendeVariationalInferenceNormalizing2016} at the cost of reducing the expressiveness of individual transformations.

\paragraph{Affine Coupling Layers.}
A popular instance of the transformations are Affine Coupling layers~\cite{dinhDensityEstimationUsing2017, kingmaGlowGenerativeFlow2018, dinhNICENonlinearIndependent2015} which allow efficient determinant computation while being expressive. For this, the input \(\data\) is split into two disjoint subsets $\data_a, \data_b$ and the output $\mathbf{y} = [\mathbf{y}_a, \mathbf{y}_b]$ is defined as

\begin{align} 
    (\mathbf{s}, \mathbf{t}) &= \mathrm{NN}(\data_b) \\
    \mathbf{y}_a &= \mathbf{s} \odot \mathbf{x}_a  + \mathbf{t} \\
    \mathbf{y}_b &= \mathbf{x}_b
\end{align}

where \(\mathrm{NN}\) can be an arbitrary neural network. As a result of this choice of transformation the Jacobian

\begin{equation}
    \frac{\partial f}{\partial \data} = 
    \begin{pmatrix}
    \mathbb{I}_d & 0 \\
    \frac{\partial \mathbf{y}_a}{\partial \data_a} & \mathrm{diag}\left(\exp (\mathbf{s}) \right) \\
    \end{pmatrix}
\end{equation}

has triangular structure. Thus, the determinant can be computed efficiently as

\begin{equation}
    \det\left(\frac{\partial f}{\partial \data}\right) = \prod_{i=1}^{D} \left( \frac{\partial f}{\partial \data} \right)_{ii}.
\end{equation}

\paragraph{Training.}
In order to optimize the parameters \(\params{}\) of the transformations $f_1, \dots, f_n$, one uses maximum likelihood estimation~\cite{rezendeVariationalInferenceNormalizing2016}. That is, one optimizes

\begin{equation}
    \max_{\params{}} \sum_{i=1}^{N} \log p_{\params{}}(\data^{(i)})
\end{equation}

given a set of data points \(\{\data^{(i)}\}_{i=1}^N\)

We compare against Normalizing Flows based on affine coupling layers in the experimental section.

\subsection{Energy-based Models}
\label{ch:background:sec:density_estimation:subsec:ebm}

\glspl{ebm}~\cite{lecunTutorialEnergyBasedLearning} use the energy-function \( \efunc \) which defines a density over the data \(\data\) as

\begin{equation}
    \label{ch:background:sec:density_estimation:subsec:ebm:eq:ebm}
    p_{\params}(\data) = \frac{\exp(-\efunc(\data))}{Z(\params)}
\end{equation}

where \(\params\) are learnable parameters and \(Z(\params) = \int \exp(-\efunc(\data)) d\data \) is the normalizing constant of the \gls{ebm}. \(Z_{\params{}}\) ensures that the induced density function \Cref{ch:background:sec:density_estimation:subsec:ebm:eq:ebm} integrates to \(1\). However, \(Z_{\params}\) is often hard to compute or even approximate since it is a high-dimensional integral.  On the positive side, \( \efunc \) can be any function \( \efunc : \mathbb{R}^D \mapsto \mathbb{R} \) placing no restrictions on the transformations compared to Normalizing Flows. 

Note that we can directly use the energy for \gls{ood} detection. That is, since a data point having high probability is equivalent to having low energy as \(p_{\params}(\data) \propto -\efunc(\data)\). This means we are not required to estimate the normalizing constant \(Z(\params)\) in practice by considering a decision threshold \(\tau\) and binary classifer \(G\) for \gls{ood} data as

\begin{equation}
    G(\data, \tau, \params) = 
    \begin{cases}
    0, \quad \text{if } -\efunc(\data) \leq \tau, \\
    1, \quad \text{if } -\efunc(\data) > \tau
    \end{cases}
\end{equation}

where \(1\) corresponds to \gls{id} and \(0\) to \gls{ood} data. 

\paragraph{Joint Energy-based model.}
We additionally consider \glspl{jem} for discriminative EBMs~\cite{grathwohlYourClassifierSecretly2020}. Suppose \( f_{\params} : \mathbb{R}^D \mapsto \mathbb{R}^C \) is a classifier assigning logits for \(C\) classes for a data point \(\data \in \mathbb{R}^D\) as in \Cref{ch:background:sec:classification}.
\citet{grathwohlYourClassifierSecretly2020} interpret the logits as unnormalized probabilities of the joint distribution

\begin{equation}
\label{eq:ebm}
    p_{\params}(\data, y) = \frac{\exp(f(\data)[y])}{Z(\params)}
\end{equation}

yielding the marginal distribution over \(\data\) as

\begin{equation}
\label{eq:jem_p_x}
    p_{\params}(\data) = \sum_y p_{\params} (\data, y)
\end{equation}

For training, \citet{grathwohlYourClassifierSecretly2020} use the factorization

\begin{equation}
\log p_{\params}(\data, y) = \log p_{\params}(\data) + \log p_{\params}(y \mid \data)
\end{equation}

using \Cref{ch:background:sec:classification:eq:predictive_distribution} and \Cref{eq:jem_p_x}. In particular, \citet{grathwohlYourClassifierSecretly2020} use a standard Cross Entropy objective to optimize \(p_{\params}(y \mid \data) \) weighted with hyperparameter \(\gamma\).

\subsubsection{Training EBMs}
\label{ch:background:sec:density_estimation:subsec:ebm:subsubsec:ebm_training}
Since the normalizing constant \(Z(\params)\) is intractable, plain maximum likelihood training is not feasible. As a result, research proposes different methods to optimize EBMs.
For optimizing \(p_{\params}(\data)\), we consider several approaches that scale to high-dimensional data. In the following, we briefly discuss these approaches.

\paragraph{Sliced Score Matching.}
\citet{hyvarinenEstimationNonNormalizedStatistical2005} propose to learn an unnormalized density by approximating the gradient of the distribution \(\boldsymbol{\psi}(\data) = \nabla_\data p(\data)\), i.e., the score of the distribution, instead of the distribution itself by minimizing

\begin{align}
    \label{ch:background:sec:ebm:eq:fisher_divergence}
    L({\params}) &= \frac{1}{2} \mathbb{E}_{p(\data)} \left[ \| \nabla_{\data} \log p_{\params}(\data) - \nabla_\data p(\data)  \|^2 \right]
\end{align}

Note that this objective does not depend on the normalizing constant \( Z(\params) \) as

\begin{equation}
    \nabla_{\data} \log p_{\params}(\data)
    = \nabla_{\data} E_{\params}(\mathbf{x}) - \underbrace{\nabla_{\mathbf{x}} Z(\params)}_{=0}
\end{equation}

However, \Cref{ch:background:sec:ebm:eq:fisher_divergence} relies on the unknown groundtruth density \(p(\mathbf{x})\) we are trying to recover. Instead,\citet{hyvarinenEstimationNonNormalizedStatistical2005} propose an optimization scheme requiring only samples from the groundtruth distribution, leveraging integration by parts, yielding 

\begin{align}
    L(\params) =& \mathbb{E}_{p(\mathbf{x})} \left[ \sum_{i=1}^D \left( \partial_i \psi_i(\mathbf{x}, \params) + \frac{1}{2} \psi_i(\mathbf{x}, \params) ^2 \right) \right] + \mathrm{constant} \\
    =& \mathbb{E}_{p(\mathbf{x})} \left[ \frac{1}{2} \sum_{i=1}^D \left( \frac{\partial^2 E_{\params}}{(\partial x_i)^2} + \frac{1}{2} \left( \frac{\partial E_{\params}}{\partial x_i}\right)^2 \right) \right] + \mathrm{constant}
    \label{ch:background:sec:ebm:eq:score_matching_obj}
\end{align}

The optimization of \Cref{ch:background:sec:ebm:eq:score_matching_obj} involves computing matrix Hessian-products which are expensive for high dimensional data. 

In order to circumvent this problem, \citet{songSlicedScoreMatching2019} propose \gls{ssm}, an alternative update formula based on a random projection. The update only involves vector-Hessian products which one can compute in constant time w.r.t. the dimensionality of the data. This change allows the training of models on high-dimensional data such as images.

\begin{equation}
    \mathbb{E}_{p_{\mathbf{v}}} \mathbb{E}_{p(\data)} \left[\mathbf{v}^T \nabla_\mathbf{x} \boldsymbol{\psi}_{\params}(\data)\mathbf{v} + \frac{1}{2} \lVert \boldsymbol{\psi}_{\params}(\data) \rVert^2_2 \right]
\end{equation}

where \(\mathbf{v} \sim p_\mathbf{v}\) is a simple distribution of random vectors, e.g. Gaussian.

\paragraph{Contrastive Divergence.}
As mentioned before, direct maximum likelihood estimation is not possible with \glspl{ebm} due to the intractable normalizing constant \(Z(\params)\).
However, it is possible to estimate the gradient of the maximum likelihood objective.

\begin{equation}
    \nabla_{\params} \log p_{\params}(\mathbf{x}) = -\nabla_{\params} E_{\params}(\mathbf{x}) - \nabla_{\params} \log Z(\params)
\end{equation}

\(\nabla_{\params} \log Z(\params)\) is intractable to compute directly, though, one can rewrite the expressions as the following expectation~\cite{songHowTrainYour2021}:

\begin{align}
    \label{ch:background:sec:ebm:eq:approx_nabla_z_theta}
    \nabla_{\params} \log Z(\params) = \mathbb{E}_{\mathbf{x} \sim p_{\params(\mathbf{x})}} \left[ \nabla_{\params} E_{\params}(\mathbf{x}) \right]
\end{align}

Following recent literature~\cite{duImplicitGenerationGeneralization2020}, we approximate the expectation in \Cref{ch:background:sec:ebm:eq:approx_nabla_z_theta} using samples obtained through \gls{sgld}~\cite{wellingBayesianLearningStochastic2011}. 

\citet{wellingBayesianLearningStochastic2011} use an Langevin diffusion process with step size \(\alpha\) and \(K\) steps in order to sample from \(p_{\params}(\data)\):

\begin{equation}
    x^{(k + 1)} = x^{(k)} + \frac{\alpha^2}{2} \nabla_x \log p_{\params}(\mathbf{x}^{(k)}) + \alpha \mathbf{z}^{(k)}, \quad \mathbf{z}^{(k)} \sim N(\mathbf{0}, \mathbf{I}), \quad k = 0, \dots, K-1
\end{equation}

Under some regularity conditions, \(\mathbf{x}^{(K)}\) is guaranteed to be a sample from \(p_{\params}(\mathbf{x})\) as $\alpha \mapsto 0$ and $K \mapsto \infty$.

Constrastive Divergence~\cite{hintonTrainingProductsExperts2002} approximates the gradient of the maximum likelihood objective using the \gls{sgld} \gls{mcmc} sampler with a finite $K$ by

\begin{equation}
\label{eq:cd}
    \nabla_{\params} p_{\params}(\mathbf{x}) = \mathbb{E}_{\mathbf{x}^\prime \sim p_{\params}(\data)} \left[ \nabla_{\params} E_{\params}(\mathbf{x}^\prime) \right] - \nabla_{\params} E_{\params}(\mathbf{x})
\end{equation}

Further, \(\mathbf{x}^{(0)}\) is initialized from a replay buffer~\cite{tielemanTrainingRestrictedBoltzmann2008}. Overall, this leads to samples which are only approximately distributed according to \(p_{\params}(\mathbf{x})\). However,~\cite{grathwohlYourClassifierSecretly2020, duImplicitGenerationGeneralization2020} show this approximation to be sufficient for training EBMs in practice. 

\paragraph{Variational Entropy Regularized Approximate Maximum Likelihood.}
Lastly, we consider the recently proposed \gls{vera} training~\cite{grathwohlNoMCMCMe2020}. \citet{grathwohlNoMCMCMe2020} propose to learn the parameters \(\boldsymbol{\phi}\) of an auxiliary distribution $q_{\boldsymbol{\phi}}$ as the optimum of

\begin{equation}
    \log Z(\params) = \max_{q_{\boldsymbol{\phi}}} \mathbb{E}_{q_{\boldsymbol{\phi}}(\data)} \left[ f_{\params}(\data) \right] + \mathcal{H}(q_{\boldsymbol{\phi}})
\end{equation}

which one can plug into \Cref{eq:ebm} to obtain a alternative method for training \glspl{ebm}. \citet{grathwohlNoMCMCMe2020} choose a generator distribution of the form \(q_{\boldsymbol{\phi}}(\data) = \int_{\mathbf{z}} q_{\boldsymbol{\phi}} (\data \mid \mathbf{z}) q_{\boldsymbol{\phi}}(\mathbf{z}) d\mathbf{z} \). To estimate the entropy term \(\mathcal{H}(q_{\boldsymbol{\phi}})\), they propose a variational approximation circumventing the need for sampling as in prior work~\cite{diengPrescribedGenerativeAdversarial2019, titsiasUnbiasedImplicitVariational2019} and thus speed up training.

Finally, note that there exist more approaches~\cite{gutmannNoisecontrastiveEstimationNew, ceylanConditionalNoiseContrastiveEstimation2018} for training \glspl{ebm}. However, they do not scale to high-dimensional data as they make strong assumptions about the data. We conduct initial experiments with these approaches in \Cref{ch:appendix:sec:other_directions}.

\subsection{Energy-based out-of-distribution detection}
\label{ch:background:sec:density_estimation:subsec:energy_ood}
In contrast to training full \glspl{ebm}, \citet{liuEnergybasedOutofdistributionDetection2020} found the energy \(E_{\params}(\mathbf{x})\) of a model, trained with cross-entropy loss only, to be effective for \gls{ood} detection. In particular, they find that the energy score outperforms other scores in the literature such as the \gls{msp} \(\max_y p_{\params}(y \mid \mathbf{x})\)~\cite{hendrycksBaselineDetectingMisclassified2018}. To motivate this, \citet{liuEnergybasedOutofdistributionDetection2020} show that the \gls{msp} score loses relevant information in the scale of the logits as

\begin{align}
    \max_y p_{\params}(y \mid \mathbf{x})
    &= \max_y \frac{\exp(\model[y]))}{\sum_{y^\prime} \exp(\model[y^\prime])} \\ \nonumber
    \label{ch:background:sec:density_estimation:subsec:energy_ood:eq:stable_softmax}
    &= \frac{1}{\sum_{y^\prime} \exp(\model[y^\prime] - \max_y \model[y])}
\end{align}

\citet{liuEnergybasedOutofdistributionDetection2020} proceed to rewrite \Cref{ch:background:sec:density_estimation:subsec:energy_ood:eq:stable_softmax} using \Cref{eq:jem_p_x} and obtain 

\begin{align}
    \log \max_y p_{\params}(y \mid \data)
    &= \efunc(\data) + \max_y \model[y] \\
    &= -\log p_{\params}(\data) - \log Z(\params) + \underbrace{\max_y \model[y]}_{\text{Not constant w.r.t. }\data} \\
    &\not\propto - \log p_{\params}(\data),
\end{align}

i.e., the \gls{msp} is not aligned with the marginal density of the model. This motivates the usage of the energy score \(\efunc(\data)\) for \gls{ood} detection instead.

To further improve the \gls{ood} detection, they propose a margin loss based on a separate dataset with \gls{ood} data \(\mathcal{D}_{\text{OOD}}\) as

\begin{align}
    \label{ch:background:sec:density_estimation:subsec:energy_ood:eq:regularization}
    L_{\text{Energy}}
    &= \mathbb{E}_{(\data, y) \sim \mathcal{D}} \left[\max(0, \efunc(\data) - m_{\text{in}})\right]^2 \\ \nonumber
    &+ \mathbb{E}_{(\data, y) \sim \mathcal{D}_{\text{OOD}}} \left[\max(0, m_{\text{out}} - \efunc(\data)) \right]^2
\end{align}

Intuitively, \Cref{ch:background:sec:density_estimation:subsec:energy_ood:eq:regularization} regularizes the energy function \(\efunc\) to output energy of at most \(m_{\text{in}}\) for \gls{id} data and at least \(m_{\text{out}}\) for \gls{ood} data. In other words, the objective encourages high \(p_{\params}(\data)\) for \gls{id} and low \(p_{\params}(\data)\) for \gls{ood} datapoints. \(m_{\text{in}}\), \(m_{\text{out}}\), and the weighting \(\lambda\) of the loss term \(L_{\text{Energy}}\) are hyperparameters that need to be tuned.

\chapter{Method}
\label{chapter:method}

In this chapter, we introduce our model which builds upon existing work discussed in \Cref{chapter:background}. In particular, we establish a connection between \glspl{ebm} and \acrlong{dpn} in \Cref{ch:method:sec:combined_view_ebm_dirichlet}. Based on our findings, we introduce \acrlong{epn} together with a training objective in \Cref{ch:method:sec:energy-priornet}. Finally, we discuss the properties of \gls{epn} in \Cref{ch:method:sec:epn:properties} where we also highlight theoretical advantages our model compared to prior work.  

\section{Derivation of Joint View}
\label{ch:method:sec:combined_view_ebm_dirichlet}

In the following, we derive a connection between a \acrfull{jem} \cite{grathwohlYourClassifierSecretly2020} and a \acrfull{dpn} \cite{malininPredictiveUncertaintyEstimation2018}.
For this, we start from a discriminative model \(f_{\params}: \mathbb{R}^D \mapsto \mathbb{R}^C\) mapping datapoints \(\mathbf{x} \in \mathbb{R}^D\) to logits. The categorical distribution over classes is given by the softmax function as

\begin{equation}
    p_{\params}(y \mid \mathbf{x}) = \frac{\exp(f_{\params}(\mathbf{x})[y])}{\sum_{y^\prime}\exp(f_{\params}(\mathbf{x})[y^\prime])}
\end{equation}

Next, let us shortly recall the definition of \gls{jem} and \gls{dpn} from \Cref{ch:background:sec:density_estimation:subsec:ebm} and \Cref{ch:background:sec:uncertainty_estimation:subsec:dirichlet_based_models}, respectively.

\paragraph{\acrlong{jem}.}
In \gls{jem}, we consider the logits to introduce a joint distribution over data \(\data\) and labels \(y\) as
\begin{equation*}
	p_{\params} (\data, y) = \frac{\exp(f_{\params}(\data)[y])}{Z ({\params})}.
\end{equation*}

We can marginalize \(y\) using \(p_{\params}(\data) = \sum_y p_{\params}(\data, y)\) and obtain for the unnormalized marginal data distribution
\begin{align}
    \label{ch:method:sec:combined_view_ebm_dirichlet:eq:marginal_px}
	\tilde{p}_{\params}(\data) &= \sum_y \tilde{p}_{\params}(\data, y) \\
	                    &= \sum_y \exp(f_{\params}(\data)[y]) 
\end{align}

where we use \(\tilde{p}\) to denote that the distribution is unnormalized.

\paragraph{\acrlong{dpn}.}
For \gls{dpn}, we follow \citet{malininPredictiveUncertaintyEstimation2018} and use the exponential function for predicting the concentration parameters of the Dirichlet distribution to ensure \(\alpha_c > 0, c \in \{1, \dots, C\}\)

\begin{equation*}
	\alpha_c = \exp(f_{\params} (\data)[c])
\end{equation*}

We then use the predicted concentration parameters \(\boldsymbol{\alpha} = [\alpha_1, \dots, \alpha_C]^T\) to parameterize the Dirichlet prior over categorical distributions \(p(\boldsymbol{\mu} \mid \mathbf{x}, \params) = Dir(\boldsymbol{\alpha})\).

As a result, the \textit{precision} of the Dirichlet distribution is given by
\begin{align*}
	\alpha_0 &= \sum_y \alpha_y \\
			 &= \sum_y \exp(f_{\params}(\data)[y])
\end{align*}

\paragraph{Joint view.}
Observe that we arrive at the same quantity \(\sum_y \exp(f_{\params}(\data)[y])\) in \gls{dpn} and \gls{jem} with different interpretations based on the considered model.
In \gls{jem}, \(\sum_y \exp(f_{\params}(\data)[y])\) corresponds to the unnormalized, marginal data density \(\tilde{p}_{\params}(\data)\), while in \gls{dpn} the same quantity corresponds to the \textit{precision} \(\alpha_0\) of the Dirchlet prior over categorical distributions \(p(\boldsymbol{\mu} \mid \mathbf{x}, \params)\): 

\begin{equation}
    \tilde{p_{\params}}(\data) = \sum_y \exp(f_{\params}(\data)[y]) = \alpha_0 
    \label{ch:method:sec:combined_view_ebm_dirichlet:eq:precision_marginal}
\end{equation}

From \Cref{ch:background:sec:uncertainty_estimation:subsec:dirichlet_based_models}, we remember that the \textit{precision} \(\alpha_0\) influences the sharpness of Dirichlet distribution which measures \textit{distributional uncertainty}. We recall that distributional uncertainty is supposed to increase in the case of \gls{ood} inputs. On the other hand, the same quantity interpreted in the scope of \gls{jem} describes the unnormalized, marginal density \(\tilde{p_{\params}}(\mathbf{x})\) which is expected to decrease for \gls{ood} data.

We conclude that both models leverage the same quantity in order to perform \gls{ood} detection. Further note that this joint view is consistent: When the precision \(\alpha_0\) is low in \gls{dpn}, the distributional uncertainty becomes high indicating \gls{ood} input. In that case, the marginal density \(\tilde{p}_{\params}(\data)\) of \gls{jem} is low as well by \Cref{ch:method:sec:combined_view_ebm_dirichlet:eq:precision_marginal} which indicates \gls{ood} samples from the perspective of \glspl{ebm}. We build upon this identification in the following section to construct a novel model.


\section{Energy-PriorNet}
\label{ch:method:sec:energy-priornet}

Based on our insights in \Cref{ch:method:sec:combined_view_ebm_dirichlet}, we derive our model, \acrfull{epn}, and an optimization objective leveraging the connection. Coincidentally, this enables us to naturally combine both methods to alleviate the disadvantages present in the individual approaches: While \glspl{ebm} have shown to be powerful density estimators enabling \gls{ood} detection based on the marginal data distribution, their discriminative formulation \gls{jem} \cite{grathwohlYourClassifierSecretly2020} lacks a Bayesian decomposition of uncertainty estimates. We find that we can use the Dirichlet-based perspective to equip an \gls{jem} with additional uncertainty estimates. 

Similarly, Dirichlet Prior Networks \cite{malininPredictiveUncertaintyEstimation2018} suffer issues w.r.t. data availability and hyperparameter choice. The energy-based view on Prior Networks allows us to perform training using energy-based optimization resolving two issues in the optimization of DPN in \Cref{ch:background:sec:dirichlet_methods:eq:priornet_objective}, i.e.,

\begin{itemize}
    \item \textbf{Requirement of separate out-of-distribution dataset \(p_{\mathrm{OOD}}\).} A dataset consisting of inputs considered as \gls{ood} might not be readily available. Further, sometimes it might even be hard to define what constitutes OOD inputs. Thus, it becomes hard to cover the entire domain of out-of-distribution inputs.
    \item \textbf{Hyperparameter choice of \(\beta_y\).} \citet{malininPredictiveUncertaintyEstimation2018} use manual specification of the concentration parameter corresponding to the target class of Dirichlet distribution \(\beta_y\) which is set based on empirical results in the experiments.
\end{itemize}

In the following, we derive the model based on our prior insights on \glspl{dpn} and \glspl{ebm} from \Cref{ch:method:sec:combined_view_ebm_dirichlet} in order to tackle aforementioned issues.

We again consider a discriminative model \(f_{\params}: \mathbb{R}^D \mapsto \mathbb{R}^C\). In particular, we do not require any architectural changes since our insights from \Cref{ch:method:sec:combined_view_ebm_dirichlet} simply rely on different interpretations of \(\exp(f_{\params}(\data)[y])\).

In a first step, we incorporate an uninformative prior \(\text{Dir}(\boldsymbol{\alpha}^{\text{prior}})\) with \(\boldsymbol{\alpha}^{\text{prior}} = [1, \dots, 1]^T\) following \citet{charpentierPosteriorNetworkUncertainty2020, malininPredictiveUncertaintyEstimation2018}. 
The setting of the prior corresponds to a flat Dirichlet distributions where all categorical distributions are equally likely which is a reasonable assumption in the presence of no observations. To incorporate the prior, remember that we can interpret the exponentiated logits \(\{e^{f_{\params}(\mathbf{x})[c]}\}_{c=1}^C\) as the concentration parameters \(\hat{\boldsymbol{\alpha}} = [e^{f_{\params}(\mathbf{x})[1]}, \dots, e^{f_{\params}(\mathbf{x})[C]}]^T\) of a Dirichlet distribution \(\text{Dir}(\hat{\boldsymbol{\alpha}})\) based on \Cref{ch:method:sec:combined_view_ebm_dirichlet}.

We can now use a Bayesian update of the prior distribution which is given in closed form as \(\text{Dir}(\boldsymbol{\alpha}^{\text{prior}} + \hat{\boldsymbol{\alpha}})\) \cite{charpentierPosteriorNetworkUncertainty2020}, i.e, we have for the concentration parameters

\begin{align}
    \label{ch:method:sec:energy-priornet:eq:pred_dirichlet}
    \alpha_c 
    &= \alpha^{\text{prior}}_c + \hat{\alpha}_c \\
    &=  1 + \exp(f_{\params}(\mathbf{x})[c])
\end{align}

to parameterize \(p(\boldsymbol{\mu} \mid \mathbf{x}, \params{}) = \mathrm{Dir}(\boldsymbol{\alpha})\). In the context of \citet{charpentierPosteriorNetworkUncertainty2020}, we treat the exponentiated logit \(\exp(f_{\params}(\mathbf{x})[c])\) as learned \textit{pseudo-count} for class \(c\) given an input \(\data\).

Next, note that \gls{epn} still embeds an \acrlong{ebm} similar to \gls{jem}.
For this, let \(f_{\params}(\data)[c]\) be the classification logit for class \(c\). We can interpret \(f_{\params}(\data)[y]\) as the (negative) joint energy \(E_{\params}(\data, y)\) over data and labels yielding the joint distribution of \gls{epn} as 

\begin{equation}
p_{\params}(\data, y) = \frac{\exp(f_{\params}(\data)[y])}{Z(\params)}
\end{equation}

The marginal distribution \(p_{\params}(\data)\) is given by marginalizing \(y\). Thus, we obtain 

\begin{equation}
    \label{ch:method:sec:energy-priornet:eq:marginal_energy}
    p_{\params}(\data) = \sum_c p_{\params}(\data, c) = \frac{\sum_c \exp(f_{\params}(\data)[c])}{Z(\params)}
\end{equation}

where the energy function is now given as \(E_{\params}(\data) = - \log \sum_c \exp(f_{\params}(\data)[c])\).

\subsection{Training}
\label{ch:method:sec:energy-priornet:subsec:training}

After specifying the model, we introduce a method for training \gls{epn}. For this, we take inspiration from \citet{malininPredictiveUncertaintyEstimation2018} for optimizing the predictive performance of our model using the \gls{kl} between the Dirichlet distribution parameterized using \Cref{ch:method:sec:energy-priornet:eq:pred_dirichlet} and a target Dirichlet distribution we specify in the following. Further, we leverage EBM training as discussed in \Cref{ch:background:sec:density_estimation:subsec:ebm}. Overall, we consider the following objective

\begin{equation}
\label{ch:method:sec:energy-priornet:eq:overall_objective}
\min -\underbrace{p_{\params}(\data)}_{\text{Density optimization}} + \underbrace{\mathrm{KL}( \mathrm{Dir}(\boldsymbol{\beta}) \mid\mid \mathrm{Dir}(\boldsymbol{\alpha}))}_{\text{Classification term}} + \lambda \underbrace{\mathcal{H}(\mathrm{Dir}(\boldsymbol{\hat{\alpha}}))}_{\text{Smoothness Regularizer}}
\end{equation}

where \(\lambda\) is a weighting hyperparameter.

Next, we define the concentration parameters of the target Dirichlet distribution \(\boldsymbol{\beta}^{(y)} = [\beta_1, \dots, \beta_C]\) as

\begin{equation}
    \label{ch:method:sec:energy-priornet:eq:target_dirichlet}
    \beta_{c}^{(y)} =
    \left\{
        \begin{array}{ll}
            1,  &   c \neq y \\
             \hat{\beta}_0 = \sum_{i=1}^C \hat{\beta}_i = \sum_{i=1}^C 1 + \exp(f_{\params}(\data)[i]) = C + \tilde{p}_{\params}(\data), &  c = y
        \end{array}
    \right.
\end{equation}

where \(y\) is label of the target class and \(C\) is the number of classes.

This objective consists of two main terms which overall contribute to \gls{epn} having good predictive performance while retaining reliable uncertainty estimates even far from the training data. In the following, we will analyze both terms and how they are optimized within \gls{epn}.

\paragraph{Optimizing the marginal data distribution \(p_{\params}(\mathbf{x})\).}
To optimize the marginal data density \(p_{\params}(\mathbf{x})\), we can leverage \gls{ebm} training techniques based on our findings in \Cref{ch:method:sec:combined_view_ebm_dirichlet}, i.e., that \gls{epn} defines a density over \(\params\) in \Cref{ch:method:sec:energy-priornet:eq:marginal_energy}.

That means we can optimize the marginal energy \(E_{\params}(\mathbf{x})\) with methods introduced in \Cref{ch:background:sec:density_estimation:subsec:ebm}. In particular, we consider two versions of our model: \textit{EPN-V} uses \gls{vera} for optimizing \(p_{\params}(\mathbf{x})\), while \textit{EPN-M} uses contrastive divergence instead.

\begin{figure}[h]
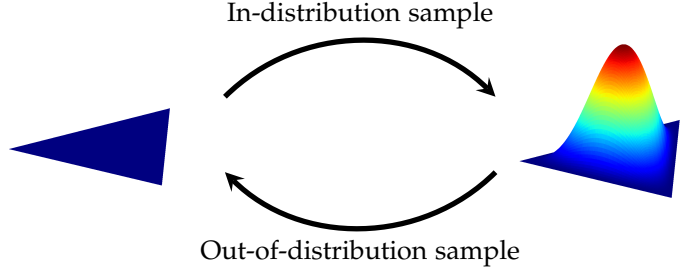

    \hspace*{\fill}%
    \begin{subfigure}[t]{.3\textwidth}
    	\centering
    	\tikz[remember picture]\node[inner sep=0pt,outer sep=0pt] (a){\includegraphics[width=.9\linewidth]{images/flat_dirichlet.png}};
    \end{subfigure}%
    \hfill
    \begin{subfigure}[t]{.3\textwidth}
    	\centering
    	\tikz[remember picture]\node[inner sep=0pt,outer sep=0pt] (b){\includegraphics[width=.9\linewidth]{images/supported_dirichlet.png}};
    \end{subfigure}
    \tikz[remember picture,overlay]\draw[line width=2pt,-stealth,black] ([xshift=-8mm]a.east) to[out=45,in=135] node[above,font=\small] {In-distribution sample} ([xshift=3mm]b.west);
    \tikz[remember picture,overlay]\draw[line width=2pt,-stealth,black] ([yshift=-1cm,xshift=3mm]b.west) to[out=225,in=315] node[below,font=\small] {Out-of-distribution sample} ([yshift=-1cm,xshift=-8mm]a.east);
    \hspace*{\fill}%
    \vspace*{10mm}
    \caption{Illustrative example on how the \gls{ebm} training acts on the posterior \(p(\boldsymbol{\mu} \mid \mathbf{x}, \mathcal{D})\) for \gls{id} and \gls{ood} samples.}
    \label{ch:method:sec:energy-priornet:subsec:training:fig:px_optim}
\end{figure}

In \gls{epn}, \gls{ebm} training allows the interpretation of pushing down on the precision of the Dirichlet distribution \(p(\boldsymbol{\mu} \mid \mathbf{x}, \params{})\) for out-of-distribution datapoints while increasing the precision for \gls{id} data points as shown in \Cref{ch:method:sec:energy-priornet:subsec:training:fig:px_optim}.

\begin{figure}[h]
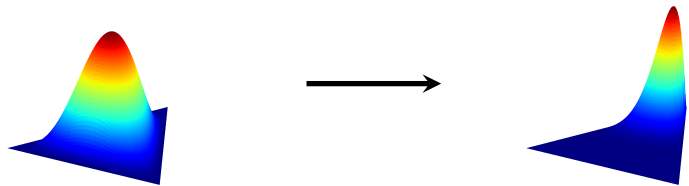

    \hspace*{\fill}%
    \begin{subfigure}[t]{.3\textwidth}
    	\centering
    	\tikz[remember picture]\node[inner sep=0pt,outer sep=0pt] (a){\includegraphics[width=.9\linewidth]{images/supported_dirichlet.png}};
    \end{subfigure}
    \hfill
    \begin{subfigure}[t]{.3\textwidth}
    	\centering
    	\tikz[remember picture]\node[inner sep=0pt,outer sep=0pt] (b){\includegraphics[width=.9\linewidth]{images/peaky_dirichlet.png}};
    \end{subfigure}
    \tikz[remember picture,overlay]\draw[line width=2pt,-stealth,black] ([xshift=3mm]a.east) --  ([xshift=-5mm]b.west);
    \hspace*{\fill}%
    \caption{Illustrative example on how the \gls{kl} term acts on the posterior \(p(\boldsymbol{\mu} \mid \mathbf{x}, \mathcal{D})\) for a particular sample with target class in the corner of the Dirichlet distribution.}
    \label{ch:method:sec:energy-priornet:subsec:training:fig:pyx_optim}
\end{figure}

\paragraph{Optimizing classification performance \(p_{\params}(y \mid \mathbf{x})\).}
With the second term of \Cref{ch:method:sec:energy-priornet:eq:overall_objective}, we aim to optimize the discriminative performance of \gls{epn}.
Intuitively, our objective forces the model to assign all probability mass of categorical distributions under the Dirichlet prior to be in the corner of the target class \(y\) as visualized in \Cref{ch:method:sec:energy-priornet:subsec:training:fig:pyx_optim}.

For further analysis, consider the predicted \(q = \mathrm{Dir}(\boldsymbol{\alpha})\) and target \(p = \mathrm{Dir}(\boldsymbol{\beta})\) with concentration parameters
\(\boldsymbol{\alpha} = \left[ \alpha_1, \dots, \alpha_C \right]^T\) and
\(\boldsymbol{\beta} = \left[ \beta_1, \dots, \beta_C \right]^T\) as defined in \Cref{ch:method:sec:energy-priornet:eq:pred_dirichlet} and \Cref{ch:method:sec:energy-priornet:eq:target_dirichlet}, respectively. Further, let
\(\alpha_0 = \sum_{c=1}^C \alpha_c\) and
\(\beta_0 = \sum_{c=1}^C \beta_c\) denote the precision of the distributions.

Since the KL-divergence between two Dirichlet distribution is given in closed form, we can expand the term:

\begin{align}
    \mathrm{KL}&( \mathrm{Dir}(p \mid\mid q) = \nonumber \\
    &\overset{(1)}{=} \log \Gamma(\alpha_0) - \sum_{c=1}^{C} \log \Gamma(\alpha_c) - \underbrace{\log \Gamma(\beta_0)}_{=\log \Gamma(\alpha_0)} + \sum_{c=1}^C \log \Gamma(\beta_k) + \sum_{c=1}^C (\alpha_c - \beta_c) (\psi(\alpha_c) - \psi(\alpha_0)) \nonumber \\
    &\overset{(2)}{=} - \sum_{c=1}^{C} \log \Gamma(\alpha_c) + \sum_{\substack{c=1 \\ c \neq t}}^C \underbrace{\log \Gamma(1)}_{=0} +  \log \Gamma(\alpha_0 - K - 1) + \sum_{\substack{c=1 \\ c \neq t}}^C (\alpha_c - 1) (\psi(\alpha_c - \alpha_0)) \nonumber \\
    &\qquad + (\alpha_t - \alpha_0 - K - 1) (\psi(\alpha_t - \alpha_0)) \nonumber \\ 
    &\overset{(3)}{=} - \sum_{c=1}^{C} \log \Gamma(\alpha_c)a + \log \Gamma(\alpha_0 - K - 1) + \sum_{\substack{c=1 \\ c \neq t}}^C (\alpha_c - 1) (\psi(\alpha_c) - \psi(\alpha_0)) \nonumber \\
    &\qquad + \sum_{\substack{c=1 \\ c \neq t}}^C \alpha_c(\psi(\alpha_0) - \psi(\alpha_t)) + (K-1)\underbrace{(\psi(\alpha_0) - \psi(\alpha_t))}_{\text{Uncertainty cross-entropy}}
\end{align}

where \(\psi\) denotes the Digamma function. In more detail, (1) uses \(\beta_0 = C + \sum_{i=1}^C \exp(f_{\params}(\mathbf{x}[i]) = \alpha_0\), (2) uses definition of \(\beta_i\), and (3) uses \(\alpha_t - \alpha_0 = - \sum_{c=1, c \neq t}^C \alpha_c\). 

Overall, we arrive at an equation where one term corresponds to the uncertainty cross-entropy loss \cite{bilosUncertaintyAsynchronousTime2020} instantiated for two Dirichlet distributions. In contrast to the cross-entropy loss, this objective incorporates uncertainty by taking the variance of the Dirichlet prior distribution into account. We find that this further motivates the usage of our objective in \Cref{ch:method:sec:energy-priornet:eq:overall_objective}. 

As an alternative, we also consider the forward \gls{kl} \(\mathrm{KL}(q \mid\mid p)\). When expanding the terms, we arrive at

\begin{equation}
\begin{aligned}
\mathrm{KL}(q \mid\mid p) = 
& - \log \Gamma(\alpha_0 - (K-1)) \\
& + \sum_{k=1}^K \log \Gamma (\alpha_k) \\
& + \sum_{k=1, k \neq t}^K (1 - \alpha_k)(\psi(1) - \psi(\alpha_0)) \\
& + \left(\left(\sum_{k=1, k \neq t}^K \alpha_k\right) - (K-1)\right)((\psi\left(\alpha_0 - (K-1)\right) - \psi(\alpha_0)))
\end{aligned}
\end{equation}

which does not allow the interpretation of minimizing the uncertainty cross-entropy objective. 

Furthermore, we can directly consider results from \citet{malininReverseKLDivergenceTraining2019} as the only difference is the parameterization of the value of the concentration parameter corresponding to the target class \(t\) in \Cref{ch:method:sec:energy-priornet:eq:target_dirichlet}. Instead, \citet{malininReverseKLDivergenceTraining2019} use a fixed value for the in-distribution dataset as discussed in \Cref{ch:background:sec:dirichlet_methods:eq:priornet_objective}. From results in\citet{malininReverseKLDivergenceTraining2019}, we know that the \gls{kl} induces a \textit{mixture of Dirichlet distributions} in expectation

\begin{equation}
    \mathcal{L}(\params, \mathcal{D}) = \mathbb{E}_{(\mathbf{x}, y) \sim \mathcal{D}} \left[ \mathrm{KL}\bigg[ \sum_{c=1}^C (\delta_{y, c}) p(\boldsymbol{\pi} \mid \boldsymbol{\beta}^{(c)}) \mid \mid p_{\params}(\boldsymbol{\pi} \mid \mathbf{x}) \bigg] \right]
\end{equation}

\begin{figure}
    \centering
    \includegraphics[width=0.3\linewidth]{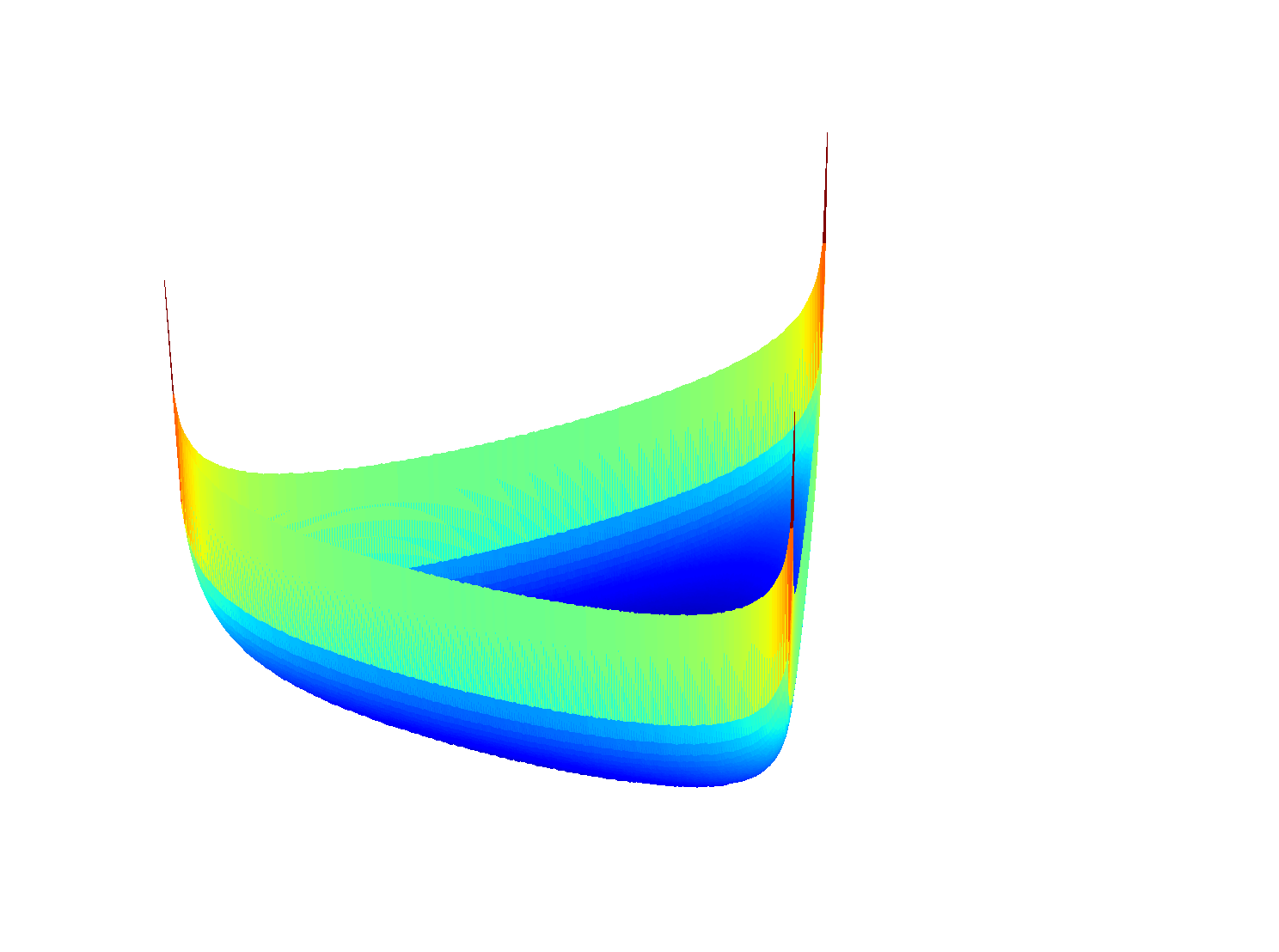}
    \caption{Induced target Dirichlet distribution for the forward KL-divergence.}
    \label{ch:method:sec:energy-priornet:fig:induced_target_dirichlet}
\end{figure}

Due to the \textit{forward} KL-divergence being \textit{zero-avoiding}, \citet{malininReverseKLDivergenceTraining2019} conclude that forward \gls{kl} objective induces a target distribution with probability mass distributed over all modes in the presence of significant data uncertainty, i.e., when multiple classes are feasible for an data point. Thus, the induced target becomes a Dirichlet distribution with "inverted" shape having low precision as shown in \Cref{ch:method:sec:energy-priornet:fig:induced_target_dirichlet}. However, we aim for \gls{epn} to assign density to a single mode within the Dirichlet similar to \gls{dpn} in \Cref{ch:background:sec:uncertainty_estimation:fig:dirichlet_vis} to enable consistent uncertainty estimation.

%

Finally, the entropy term in \Cref{ch:method:sec:energy-priornet:eq:overall_objective} acts as a regularizer ensuring smoothness of the predicted Dirichlet distribution \(p_{\params}(\boldsymbol{\mu} \mid \mathbf{x})\).

\subsection{Properties of Energy-Prior Network}
\label{ch:method:sec:epn:properties}
After specifying the model and investigating its training, we explore the characteristics of \gls{epn}. 
\citet{heinWhyReLUNetworks2019} discovered that neural networks with ReLU activations converge to high confidence predictions far from the training data. Further, \citet{ulmerKnowYourLimits2021} prove based on results in \citet{heinWhyReLUNetworks2019} that various uncertainty measures discussed in \Cref{ch:background:sec:uncertainty_estimation:subsec:uncertainty_measures}, i.e., the class variance, predictive entropy, and approximate mutual information, converge to fixed scores. In the following, we demonstrate that \gls{epn} does not suffer such issues even for data far from the training data distribution. 

To show this, consider a trained \gls{epn} \(E_{\params}: \mathbb{R}^D \mapsto \mathbb{R}^C \) where the energy increases for inputs \(\gamma \odot \mathbf{x}, \gamma \in \mathbb{R}\)

\begin{equation}
    \label{ch:method:sec:epn:properties:eq:increasing_energy}
    \lim_{\gamma \mapsto \infty} E_{\params}(\gamma \odot \mathbf{x}) = \infty
\end{equation}

This is a reasonable assumption since the objectives used for training the underlying \gls{ebm} embedded within \gls{epn} in \Cref{ch:background:sec:density_estimation:subsec:ebm} are consistent in the limit \cite{songHowTrainYour2021}.

As a consequence of the definition of an \gls{ebm}, the marginal density \(p_{\params}(\mathbf{x})\) learned by the model decreases as one moves from the training data:

\begin{align}
    \lim_{\gamma \mapsto \infty} \exp(-E_{\params}(\gamma \odot \mathbf{x})) &= \lim_{\gamma \mapsto \infty} \tilde{p}_{\params}(\gamma \odot \mathbf{x})  \\
    &= 0
\end{align}

Thus, one obtains for the concentration parameters \(\boldsymbol{\alpha} = [\alpha_1, \dots, \alpha_C]\) of the Dirichlet distribution in \Cref{ch:method:sec:energy-priornet} parameterizing the posterior \(p(\boldsymbol{\mu} \mid \mathbf{x}, \params)\) 

\begin{align}
    \lim_{\gamma \mapsto \infty} \alpha_c &= \lim_{\gamma \mapsto \infty} \alpha^{\text{prior}}_c + \hat{\alpha}_c \\
    &= \alpha^{\text{prior}}_c + \underbrace{\lim_{\gamma \mapsto \infty} \exp(f_{\params}(\gamma \odot \mathbf{x})[c])}_{=0} \\
    &= \alpha^{\text{prior}}_i 
\end{align}

which corresponds to a flat Dirichlet distribution for our choice of prior \(\boldsymbol{\alpha}^{\text{prior}} = \mathbf{1}\). Thus, the \textit{distributional uncertainty} as measured by differential entropy is maximal, as desired for \gls{ood} inputs.

Furthermore, for the expected predictive distribution, we obtain

\begin{align}
    \label{ch:method:sec:epn:properties:eq:expected_predictive_distribution}
    p(y \mid \mathbf{x}, \params) &= \int p(y \mid \boldsymbol{\mu}) p(\boldsymbol{\mu} \mid \mathbf{x}, \params) d\boldsymbol{\mu} \\
    &= \frac{\alpha_y}{\sum_i \alpha_i} \\
    &= \frac{1}{C}
\end{align}

Thus, \gls{epn} converges to a uniform prediction over the classes far from the training data. This allows reliable uncertainty estimates for \gls{ood} data since, in contrast to ReLU networks for classification \cite{heinWhyReLUNetworks2019}, the predictive distribution converges to a maximally uncertain prediction. For example, the \textit{maximum softmax probability} score \cite{lakshminarayananSimpleScalablePredictive2017,hendrycksBaselineDetectingMisclassified2018} for \gls{epn} becomes \(\max p_{\params}(y \mid \data) = \nicefrac{1}{C}\) which is the lowest possible, indicating maximum uncertainty.

\subsection{Enforcing asymptotic behavior}
\label{ch:method:sec:epn:subsec:asymptotic_behavior}
In \Cref{ch:method:sec:epn:properties:eq:increasing_energy}, we assume that the energy increases far from the training data based on the consistency of the \gls{ebm} training approaches \cite{songHowTrainYour2021}. However, we can consider a slight adaptation of the neural network architecture to enforce this behavior. This change provably ensures that the energy \(\efunc(\data)\) increases and thus the density \(p_{\params}(\data)\) defined by the \gls{ebm} decreases moving away from the training data. For this, we consider ReLU-networks \cite{heinWhyReLUNetworks2019}, i.e., neural networks using ReLU or LeakyReLU activations only. Originally, \citet{meinkeProvablyRobustDetection2021} introduce the considered parameterization to ensure that the classifier converges to uniform predictions over classes far from the training data.

\begin{figure}
    \centering
    \includegraphics{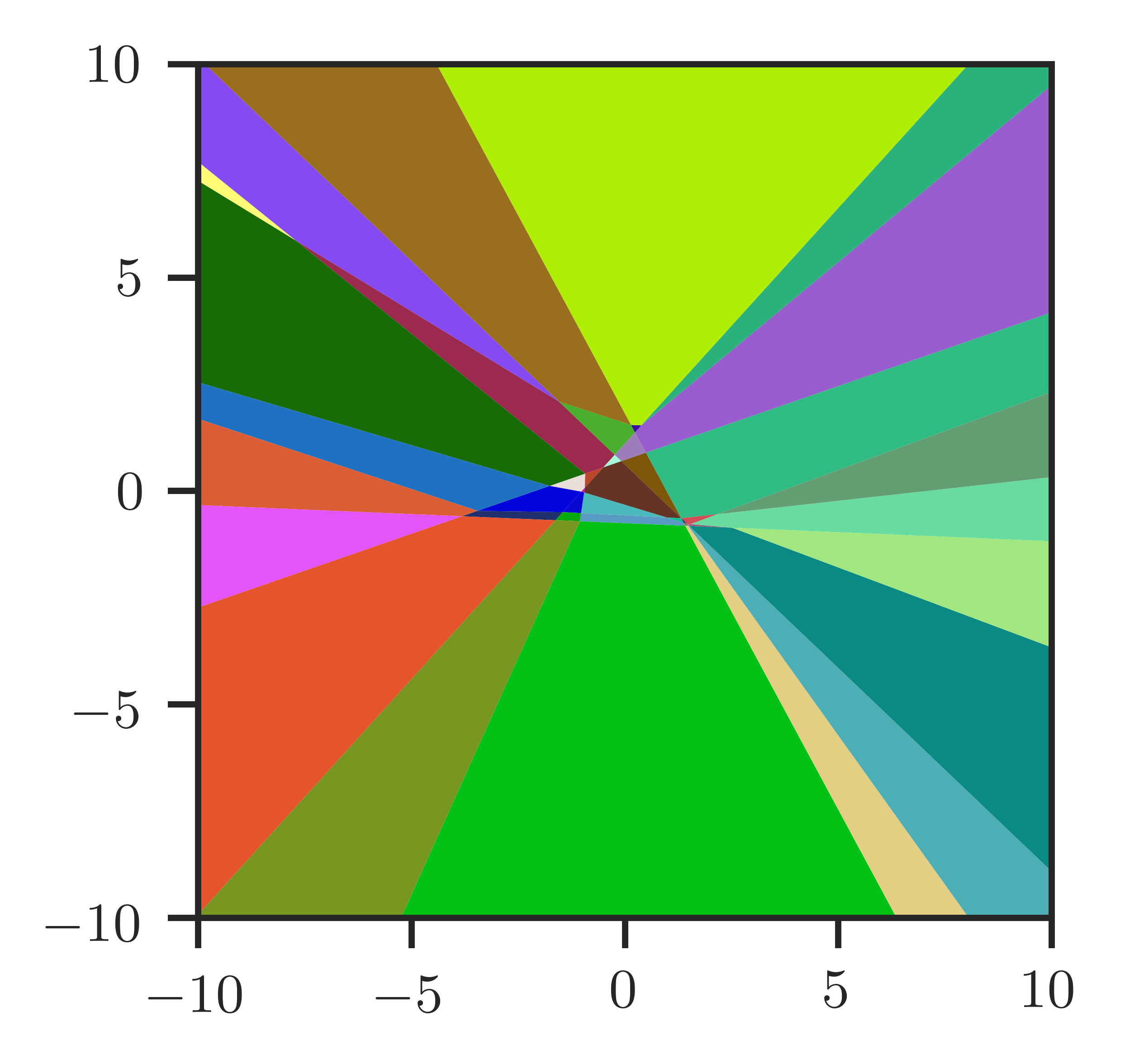}
    \caption{Visualization of the regions in \(\mathbb{R}^2\) on which the output of a 2-layer ReLU-network is an affine transformation of the input. We use code from \citet{jordanProvableCertificatesAdversarial2019} for plotting.}
    \label{ch:method:sec:energy-priornet:fig:polytopes}
\end{figure}

Similar to \citet{meinkeProvablyRobustDetection2021}, we use an established result by \citet{heinWhyReLUNetworks2019} which shows that by scaling a datapoint, one arrives at a region (convex polytope), extending to infinity, on which the output of the ReLU-network is an affine transformation of the input. For a visualization of the polytopes of a simple 2-layer neural network, see \Cref{ch:method:sec:energy-priornet:fig:polytopes}. Note that the outermost polytopes extend to infinity.

\begin{lemma}[\cite{heinWhyReLUNetworks2019}]
\label{lemma1}
Let \(\{Q_r\}_{r=1}^R\) be the set of convex polytopes associated with the ReLU-network \(f : \mathbb{R}^D \mapsto \mathbb{R}^C\) such that for every \(k \in \{1, \dots, R \}\) and \(\data \in Q_k\), there exists \(V^k \in \mathbb{R}^{C\times D}\) and \(\mathbf{d}^k \in \mathbb{R}^C\) with \(f(x) = V^k \data + \mathbf{d}^k\). For any \(\data \in \mathbb{R}^D, \data \neq 0\) there exists \(\alpha \in \mathbb{R}\) and \(t \in \{1, \dots, R\}\) such that \(\beta \data \in Q_t\) for all \(\beta \geq \alpha\).
\end{lemma}

Based on \Cref{lemma1}, the following theorem shows that the marginal energy \(\efunc(\data)\) for a \gls{jem} with a slightly adapted architecture increases as one moves away from the training data. Correspondingly, this means that the marginal data density \(p_{\params}(\data)\) decreases as one would expect from any reasonable density estimate.

\clearpage
\setcounter{theorem}{0}
\begin{theorem}
\label{theorem1}
Let \(\data \in \mathbb{R}^D, \data \neq \mathbf{0}\) and let \(g:\mathbb{R}^D \mapsto \mathbb{R}^K\) be a ReLU-network with parameters \(\boldsymbol{\phi}\) where the last layer explicitly contains a ReLU activation. By \Cref{lemma1}, there exists a finite set of polytopes \(\{Q_r\}_r\) on which \(g\) is affine. Denote by \(Q_t\) the polytope such that \(\beta \data \in Q_t\) for all \(\beta \geq \alpha\) with \(\alpha \in \mathbb{R}\) and by \(g(\mathbf{z}) = U\mathbf{z}+\mathbf{d}\) with \(U \in \mathbb{R}^{K\times D}\) and \(\mathbf{d} \in \mathbb{R}^K\) be the output of \(g\) for \(\mathbf{z} \in Q_t\). Finally, let \(f:\mathbb{R}^D \mapsto \mathbb{R}^C, f(\data) = W g(\data) + \mathbf{b}\) 
be the Energy-based model where the last weight matrix \(W \in \mathbb{R}^{C\times K}\) is component-wise negative and \(\mathbf{b} \in \mathbb{R}^C\) .
If \(U\data \neq \mathbf{0}\), then

\begin{equation}
    \lim_{\beta \mapsto \infty} \efunc(\beta \data) = \lim_{\beta \mapsto \infty} -\log \sum_{c=1}^C e^{(f(\beta \data))_c} = \infty \nonumber
\end{equation}

where \(\params = \{\boldsymbol{\phi}, W, \mathbf{b}\}\) are the learnable parameters.

Furthermore, \(\lim_{\beta \mapsto \infty} p_{\params}(\beta \data) = 0\).
\end{theorem}

We proof this statement in \Cref{ch:appendix:sec:proof}. Based on \Cref{theorem1}, we can achieve that the data density \(p_{\params}(\data)\) defined by the \gls{ebm} decays by ensuring that the last weight matrix of the ReLU-network is component-wise negative.

We find that leveraging this result is not necessary in our model since \gls{epn} always provided increasing energy in our experiments. However, we discover that \Cref{theorem1} is useful for \glspl{jem} \cite{grathwohlYourClassifierSecretly2020} in \Cref{ch:results:sec:ood_detection_ebms:subsec:supervision_improves_ood} to ensure that the density decreases.
\chapter{Results}
\label{ch:results}
One focus of this chapter is to provide experimental evaluation of \acrfull{epn} introduced in \Cref{chapter:method}. We provide empirical analysis in \Cref{ch:results:sec:eval_energy_priornet}. Secondly, we dive deeper into the density estimation capabilities of \glspl{ebm} in \Cref{ch:results:sec:ood_detection_ebms} where we compare against Normalizing Flows as a density estimation baseline. Moreover, we investigate the reasons for  \gls{ood} detection capabilities of \glspl{ebm} in recent work~\cite{grathwohlYourClassifierSecretly2020, grathwohlNoMCMCMe2020}.

\section{Evaluation of Energy-Prior Network}
\label{ch:results:sec:eval_energy_priornet}
In this section, we provide an experimental study of the properties of \gls{epn}. In particular, we demonstrate the characteristics of our model, established in \Cref{ch:method:sec:epn:properties}, on a toy dataset. Further, we compare \gls{epn} against a model trained with cross-entropy loss as a baseline. 

\subsection{Toy Dataset}
\label{ch:results:sec:eval_energy_priornet:subsec:toy_dataset}
\begin{figure}[ht]
    \centering
    \includegraphics{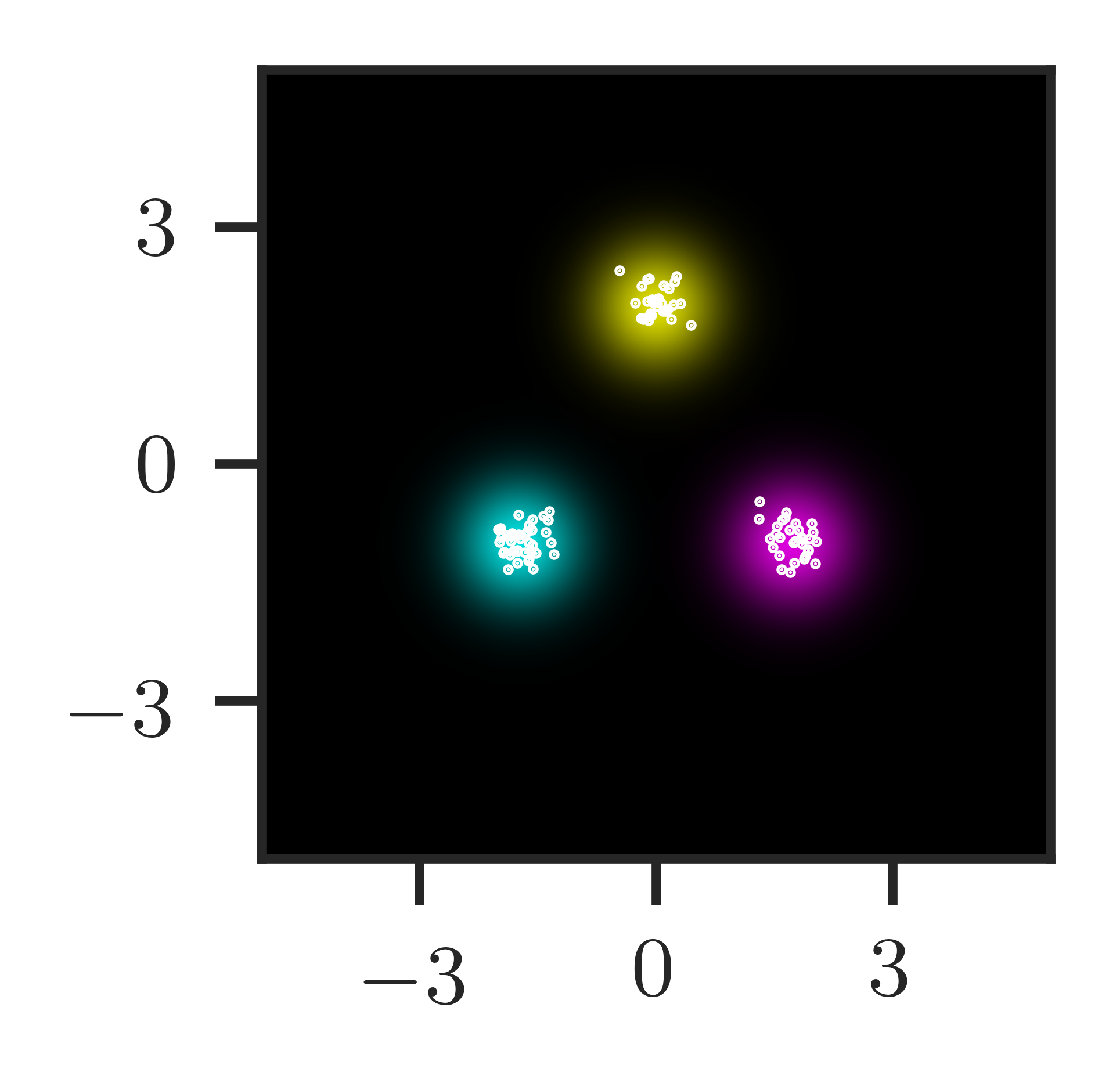}
    \caption{Visualization of the 3-Gaussian dataset with a subset of the dataset samples plotted in white.}
    \label{ch:results:sec:eval_energy_priornet:fig:3gaussian_gt}
\end{figure}

To investigate and visualize the properties of our model discussed in \Cref{ch:method:sec:epn:properties}, we first consider a small toy dataset. This dataset consists of samples drawn from three Gaussian distributions as shown in \Cref{ch:results:sec:eval_energy_priornet:fig:3gaussian_gt}. The class labels correspond to the index of the Gaussian that produced a particular sample. The task is to classify the data points into the correct classes. To demonstrate the issues of the conventional approach for classification, we additionally compare against a model trained with cross-entropy objective introduced in \Cref{ch:background:sec:classification} as a baseline.
While this task is simple since the dataset is only two-dimensional, it allows us to visualize the behavior of a classification model in terms of its confidence and out-of-distribution detection capabilities.

\paragraph{Setup.}
We use an existing dataset from \citet{charpentierPosteriorNetworkUncertainty2020} with \(1500\) samples uniformly generated from each Gaussian distribution \(N(\boldsymbol{\mu}, \sigma \mathbb{I}_2)\). The means of the Gaussians \(\boldsymbol{\mu} \) are \([0, 2]^T\), \([\sqrt{3}, -1]^T\), and \([-\sqrt{3}, -1]^T\), respectively, while the standard deviation is \(\sigma = 0.2\). Each model uses \(5\) fully connected, hidden layers using ReLU activation with dimensionality \(100\), input size \(2\) and output size \(3\) corresponding to the three classes.

\paragraph{Cross-entropy baseline.}
We use a softmax activation on the logits and train the model using the cross-entropy objective. As expected, the accuracy converges to \(100\%\) quickly due to the simplicity of the task. 
We proceed to investigate the uncertainty estimates and \gls{ood} detection scores.

\begin{figure}[ht]
    \centering
    \includegraphics{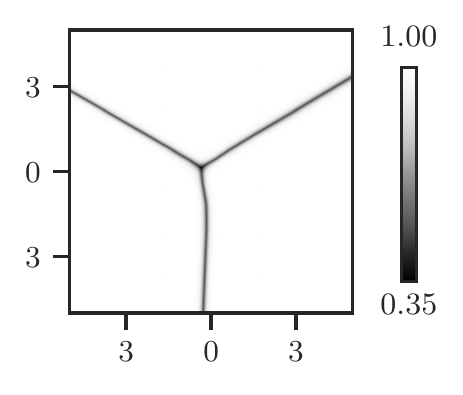}
    \caption{Visualization of the confidence of the cross-entropy baseline.}
    \label{ch:results:sec:eval_energy_priornet:fig:ce_baseline_toy_confidence}
\end{figure}

\Cref{ch:results:sec:eval_energy_priornet:fig:ce_baseline_toy_confidence} shows the confidence \(\max_y p_{\params}(y \mid \data)\) of the classifier. As recent theoretical results~\cite{heinWhyReLUNetworks2019} predict, the confidence increases when moving away form the training data, while confidence is low on the decision boundary only. When comparing with the dataset in \Cref{ch:results:sec:eval_energy_priornet:fig:3gaussian_gt}, the model predicts confidently on data it was not trained on, which potentially leads to confident, wrong predictions.

\begin{figure}[ht]
    \centering
    \hspace*{\fill}%
    \begin{subfigure}[t]{0.45\linewidth}
        \centering
        \includegraphics{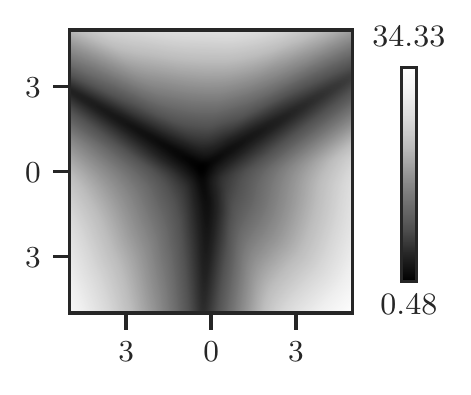}
        \caption{Unnormalized, marginal data density \(\tilde{p}_{\params}(\data{})\).}
        \label{ch:results:sec:eval_energy_priornet:subfig:ce_baseline_toy_marginal}
    \end{subfigure}%
    \hfill
    \begin{subfigure}[t]{0.45\linewidth}
        \centering
        \includegraphics{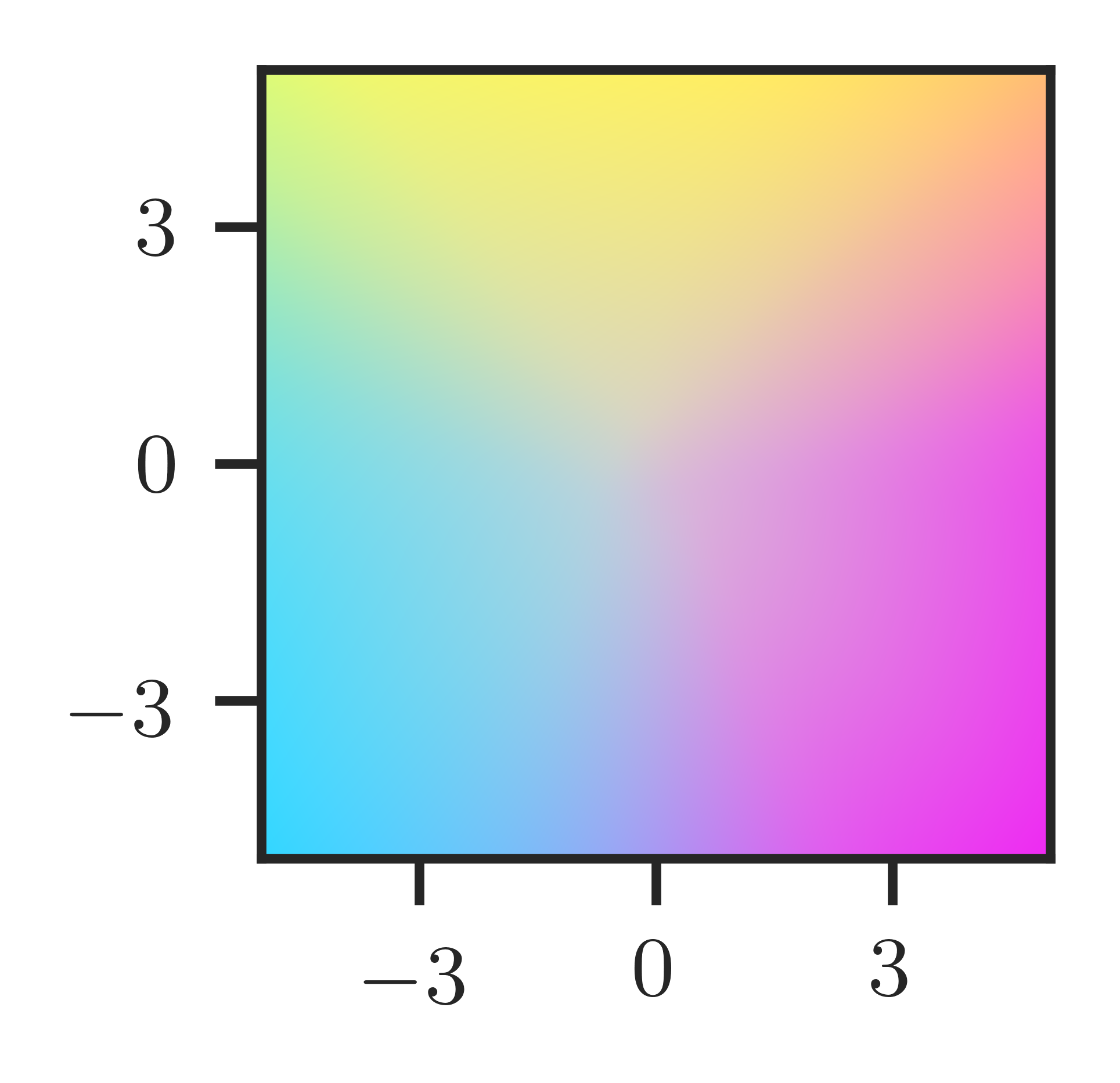}
        \caption{Unnormalized, class-conditional densities \(\tilde{p}_{\params}(y \mid \data{})\).}
        \label{ch:results:sec:eval_energy_priornet:subfig:ce_baseline_toy_class_conditionals}
    \end{subfigure}%
    \hspace*{\fill}%
    \caption{Visualization of distributions of the cross-entropy baseline using the interpretation of logits in \citet{grathwohlYourClassifierSecretly2020}. Colors correspond to the different classes.}
    \label{ch:results:sec:eval_energy_priornet:fig:ce_baseline_toy}
\end{figure}

Further, \Cref{ch:results:sec:eval_energy_priornet:fig:ce_baseline_toy} shows the unnormalized marginal and class-conditional densities for the cross-entropy baseline interpreted in the framework of \citet{grathwohlYourClassifierSecretly2020}. Note that the marginal data distribution \(\umarginal\) increases with larger distance to the training data. As a result, one is not able to detect \gls{ood} data based on the density. Note that this is also holds for the recently proposed energy score \(E_{\params}(\data{})\)~\cite{liuEnergybasedOutofdistributionDetection2020} in \Cref{ch:background:sec:density_estimation:subsec:energy_ood} which relates to the visualized quantity as \(\tilde{p}_{\params}(\data{}) = \exp(-E_{\params}(\data{}))\).

\paragraph{Energy-Prior Network.}

\begin{figure}[ht]
    \centering
    \includegraphics{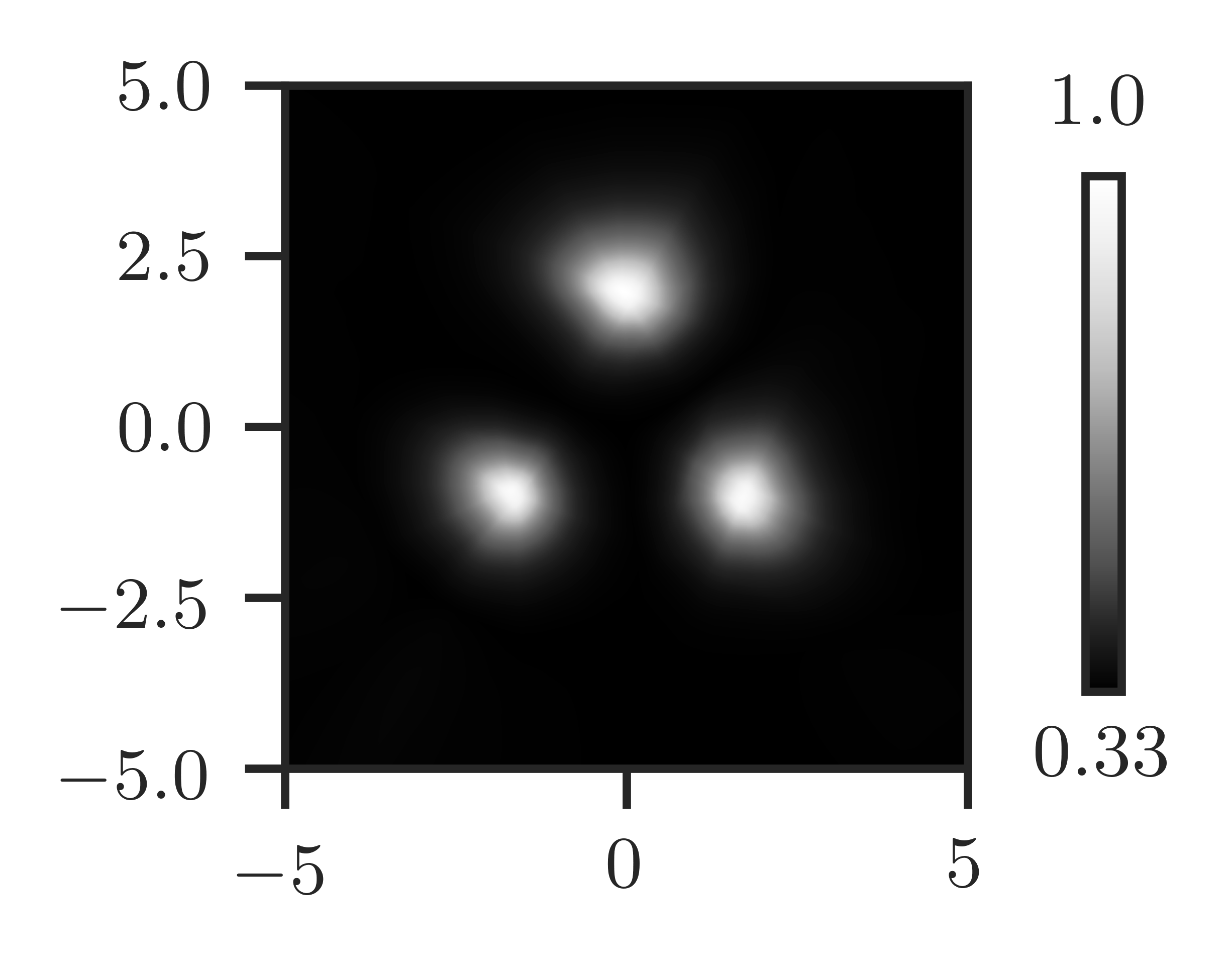}
    \caption{Visualization of the confidence of \gls{epn}.}
    \label{ch:results:sec:eval_energy_priornet:fig:epn_toy_confidence}
\end{figure}

On the other hand, we consider \gls{epn}. In contrast to the cross-entropy baseline, the confidence decreases for \gls{epn} with increasing distance to the training data as shown in \Cref{ch:results:sec:eval_energy_priornet:fig:epn_toy_confidence}. Thus, the confidence, as a measure of total uncertainty, allows \gls{epn} to differentiate \gls{id} from \gls{ood} data. Moreover, the confidence is only high in the regions of the individual classes. As expected, the confidence is low in between the classes since it is unknown whether a sample in this region belongs to either class. 

\begin{figure}[ht]
    \centering
    \hspace*{\fill}%
    \begin{subfigure}[t]{0.45\linewidth}
        \centering
        \includegraphics{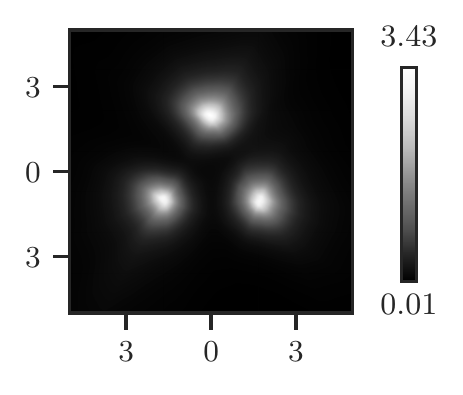}
        \caption{Unnormalized, marginal data density \(\tilde{p}_{\params}(\data)\) and equivalently the precision \(\alpha_0\) of the Dirichlet.}
        \label{ch:results:sec:eval_energy_priornet:subfig:epn_toy_marginal}
    \end{subfigure}%
    \hfill
    \begin{subfigure}[t]{0.45\linewidth}
        \centering
        \includegraphics{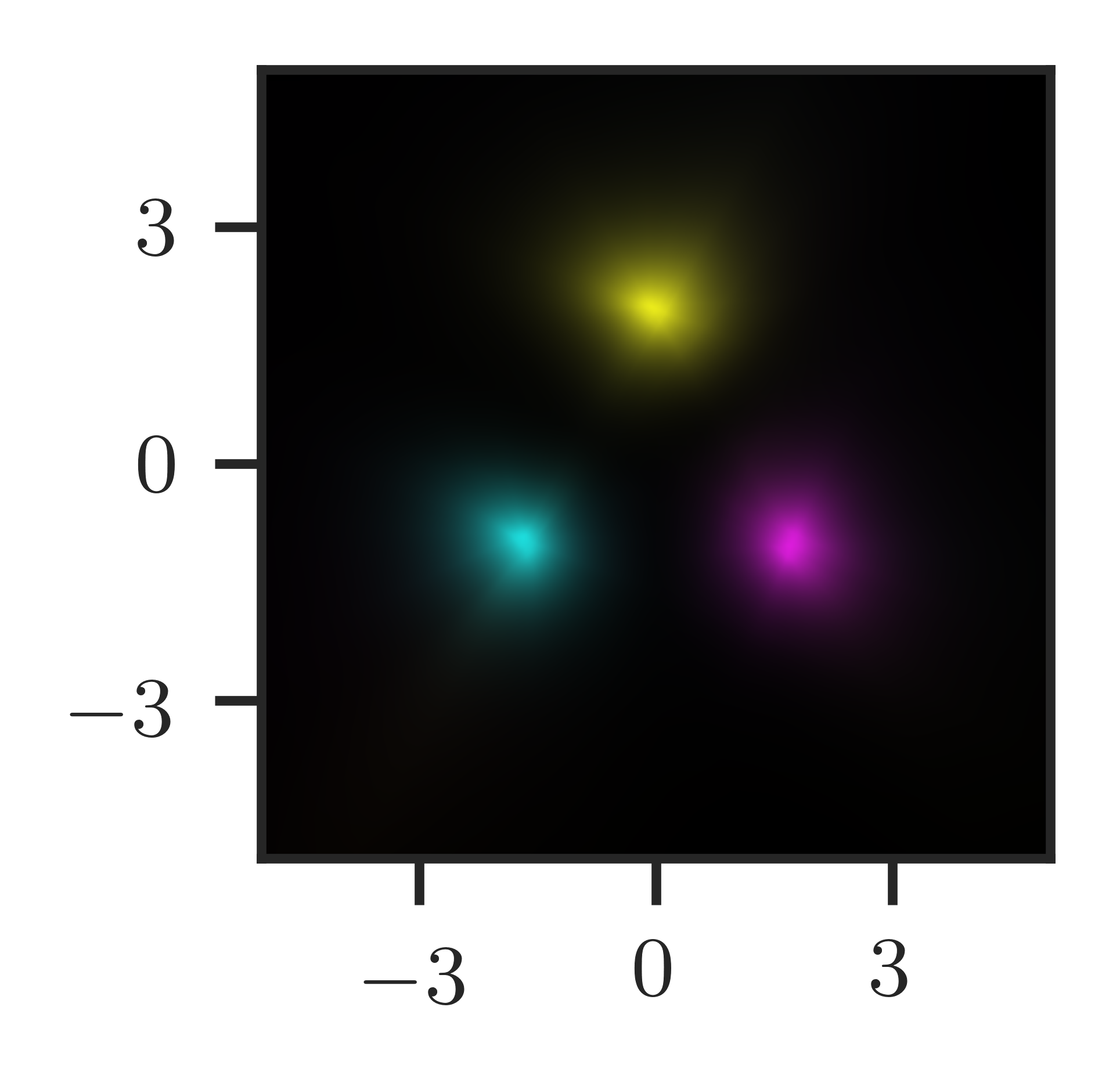}
        \caption{Unnormalized, class-conditional densities \(\tilde{p}_{\params}(y \mid \data{})\) and equivalently the concentration parameters \(\boldsymbol{\hat{\alpha}}\) of the Dirichlet.}
        \label{ch:results:sec:eval_energy_priornet:subfig:epn_toy_class_conditionals}
    \end{subfigure}%
    \hspace*{\fill}%
    \caption{Visualization of distributions of \gls{epn} which also corresponds to the parameters of the predictive Dirichlet distribution as shown in \Cref{ch:method:sec:combined_view_ebm_dirichlet}. Colors correspond to the different classes.}
    \label{ch:results:sec:eval_energy_priornet:fig:epn_toy}
\end{figure}

Similar results hold for the marginal and class-conditional distributions shown in \Cref{ch:results:sec:eval_energy_priornet:fig:epn_toy}. The estimated distributions follow closely the underlying data generating distribution of the 3-Gaussians dataset in \Cref{ch:results:sec:eval_energy_priornet:fig:3gaussian_gt}. The result demonstrates that \gls{epn} produces satisfactory density estimates useful for \gls{ood} detection.

\begin{figure}[ht]
    \centering
    \includegraphics{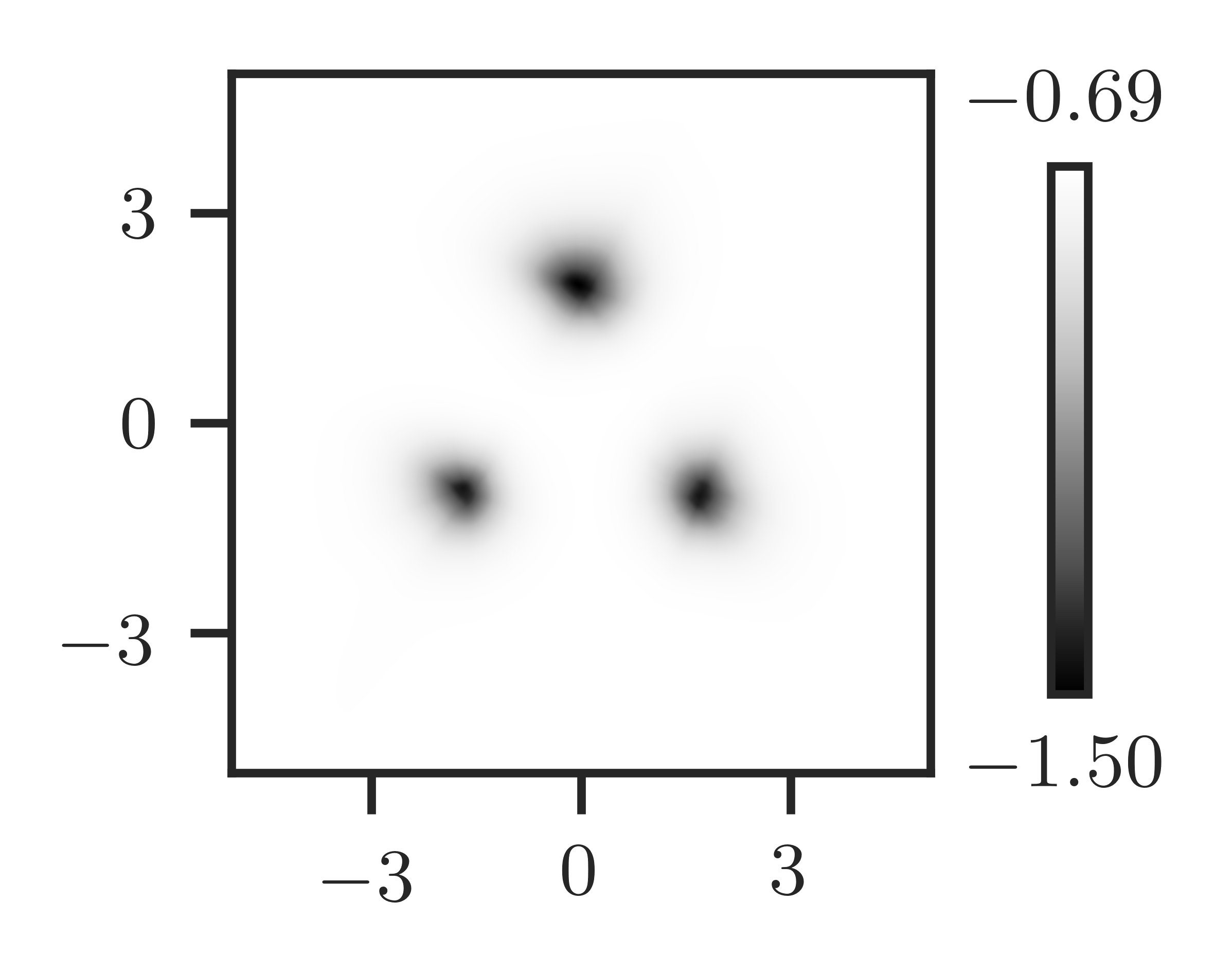}
    \caption{Visualization of the differential entropy of \gls{epn} on the 3-Gaussians dataset.}
    \label{ch:results:sec:eval_energy_priornet:fig:epn_toy_diff_entropy}
\end{figure}

Additionally, we can leverage the view on \gls{epn} predicting the parameters of a Dirichlet distribution derived in \Cref{ch:method:sec:combined_view_ebm_dirichlet} in order to estimate uncertainty. Therefore, we can, similar to \gls{dpn}~\cite{malininPredictiveUncertaintyEstimation2018}, consider the differential entropy to measure distributional uncertainty. In \Cref{ch:results:sec:eval_energy_priornet:fig:epn_toy_diff_entropy}, we observe that \gls{epn} is able to produce distributional uncertainty estimates which are low for regions with training data and increasing uncertainty for \gls{ood} data as desired.

\begin{figure}[ht]
    \centering
    \includegraphics{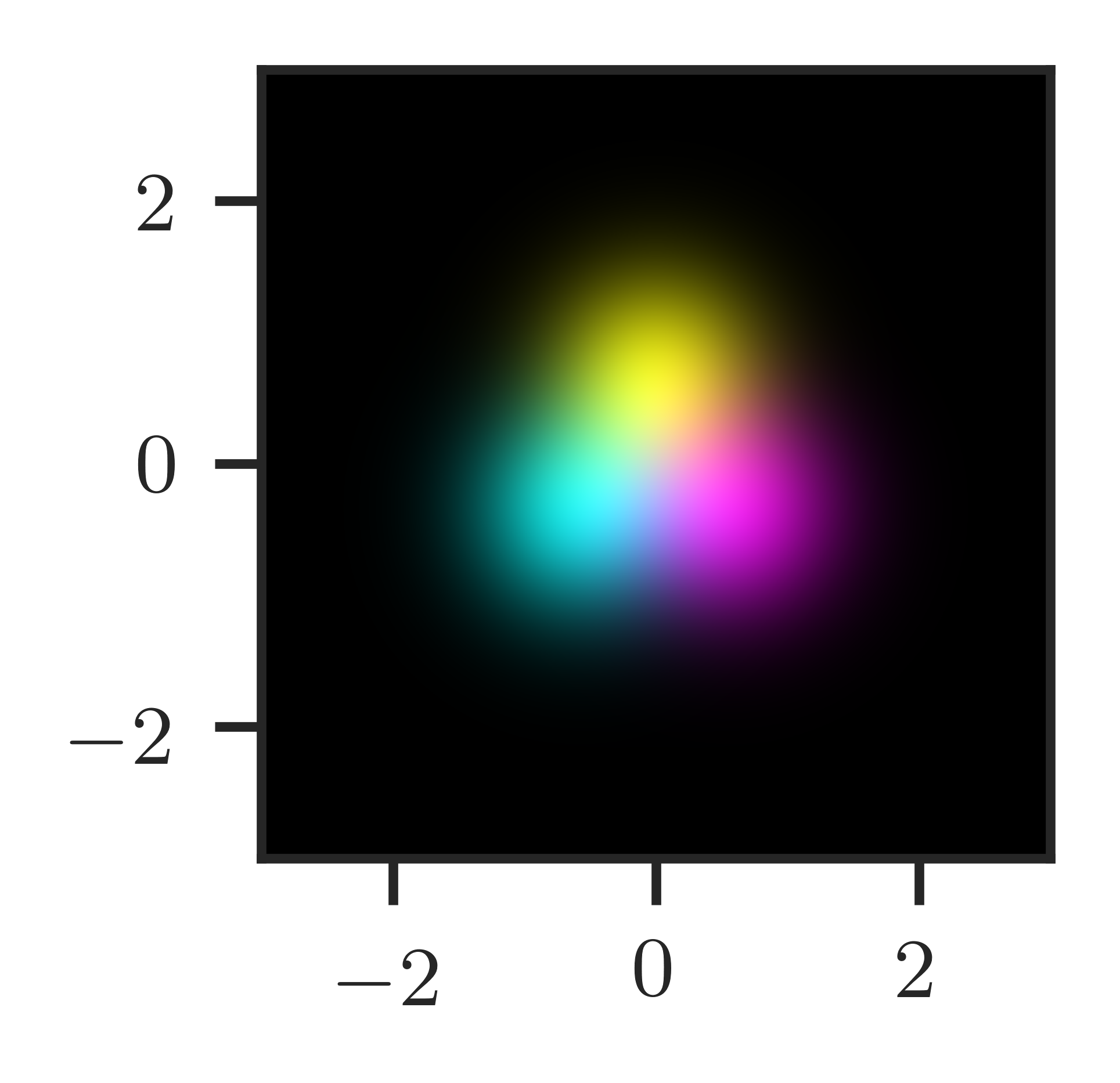}
    \caption{Variant of the 3-Gaussians dataset with overlapping classes.}
    \label{ch:results:sec:eval_energy_priornet:fig:3gaussian_overlap_gt}
\end{figure}

Finally, we consider another version of the 3-Gaussian dataset in \Cref{ch:results:sec:eval_energy_priornet:fig:3gaussian_overlap_gt}. We move the Gaussian distributions closer together such that the Gaussians overlap on the borders, which introduces \textit{data uncertainty} within the overlapping regions. As a result, we can use this dataset to investigate whether \gls{epn} can differentiate between different sources of uncertainty. 

\begin{figure}[ht]
    \centering
    \begin{subfigure}[t]{0.30\linewidth}
        \centering
        \includegraphics{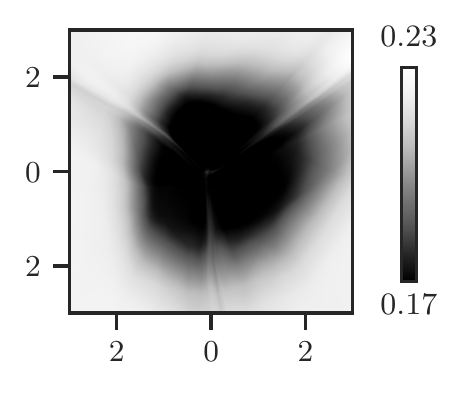}
        \caption{Epistemic Uncertainty.}
        \label{ch:results:sec:eval_energy_priornet:subfig:overlap_mutual_information}
    \end{subfigure}%
    \hfill
    \begin{subfigure}[t]{0.30\linewidth}
        \centering
        \includegraphics{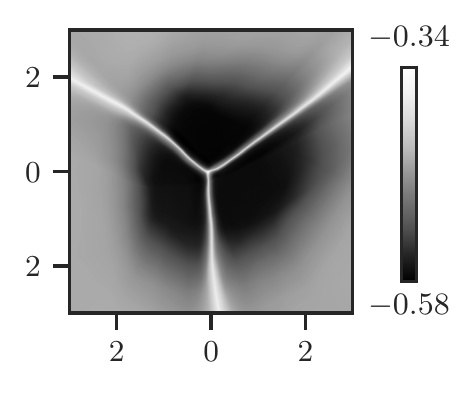}
        \caption{Aleatoric Uncertainty.}
        \label{ch:results:sec:eval_energy_priornet:subfig:overlap_aleatoric_uncert}
    \end{subfigure}%
    \hfill
    \begin{subfigure}[t]{0.30\linewidth}
        \centering
        \includegraphics{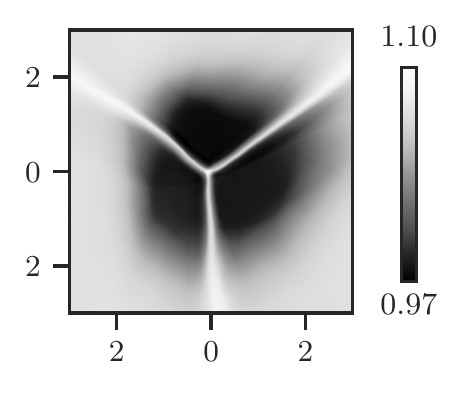}
        \caption{Overall Uncertainty.}
        \label{ch:results:sec:eval_energy_priornet:subfig:overlap_entropy_predictive}
    \end{subfigure}%
    \caption{Different types of uncertainty estimated by \gls{epn}.}
    \label{ch:results:sec:eval_energy_priornet:fig:epn_toy_overlap_uncertainties}
\end{figure}

In \Cref{ch:results:sec:eval_energy_priornet:fig:epn_toy_overlap_uncertainties}, we visualize the different types of uncertainty as introduced in \Cref{chapter:background:eq:bayesian_uncertainty}. For measuring \textit{model uncertainty}, we use the mutual information between the categorical label and the Dirichlet prior \(\mathcal{I}[y, \boldsymbol{\mu} \mid \mathbf{x}]\), for \textit{data uncertainty}, we use the expected entropy under the Dirichlet prior \(\mathbb{E}_{p(\boldsymbol{\mu} \mid \data, \mathcal{D})}[\mathcal{H}[p(y \mid \boldsymbol{\mu})] ]\), and to measure the total uncertainty, we use the entropy of the predictive distribution \(\mathcal{H}[p_{\params}(y \mid \data)]\).

We observe in \Cref{ch:results:sec:eval_energy_priornet:subfig:overlap_aleatoric_uncert} that \gls{epn} is able to assign high data uncertainty to the overlapping regions as desired. Furthermore, the model uncertainty is low in the overlapping regions and high outside the data as shown in \Cref{ch:results:sec:eval_energy_priornet:subfig:overlap_mutual_information}. This behavior is desired since the cause of uncertainty in the overlapping regions is the data generating process, i.e., the uncertainty is irreducible even when observing more data points. Overall, we conclude that \gls{epn} can estimate different types of uncertainties on the toy dataset.

\subsection{Out-of-distribution detection}
\label{ch:results:sec:eval_energy_priornet:subsec:ood_detection}
After demonstrating that \gls{epn} can provide accurate density and uncertainty estimates on the toy dataset, we evaluate the capability of \gls{epn} at detecting \gls{ood} inputs and compare with other methods in the literature on more realistic datasets. For evaluation, we train models on one dataset and then test the capability of the model to detect samples from other datasets. We consider \gls{ood} detection as a binary classification problem with labels \(1\) for \gls{id} and \(0\) for \gls{ood} and report the \gls{aucpr} as commonly done in the literature~\cite{hendrycksBaselineDetectingMisclassified2018}.

\paragraph{Datasets.}
Following~\cite{charpentierPosteriorNetworkUncertainty2020}, we consider the tabular datasets \textit{Sensorless drive}~\cite{UCIMachineLearning2021} and \textit{Segment}~\cite{UCIMachineLearning2021}, with dimensionality \(18\) and \(49\), and \(4\) and \(11\) classes, respectively. To obtain a representative \gls{ood} dataset, we remove one class (\textit{sky}) from the Segment and two classes (\textit{10}, \textit{11}) from the Sensorless drive dataset.

Further, we evaluate on image datasets. We use FMNIST \cite{xiaoFashionMNISTNovelImage2017} as \gls{id} dataset and MNIST \cite{lecunGradientbasedLearningApplied1998}, NotMNIST \cite{bulatovMachineLearningEtc2011}, KMNIST \cite{clanuwatDeepLearningClassical2018} as \gls{ood} datasets. Further, we train on CIFAR-10 \cite{krizhevskyLearningMultipleLayers2009} and use LSUN \cite{yuLSUNConstructionLargescale2016}, Textures~\cite{huangCompactConvolutionalNeural2020}, CIFAR-100 \cite{krizhevskyLearningMultipleLayers2009}, SVHN \cite{netzerReadingDigitsNatural2011} and Celeb-A~\cite{liu2015faceattributes} as \acrlong{ood}.

\begin{figure}[ht]
    \centering
    \hspace*{\fill}%
    \begin{subfigure}[t]{0.3\linewidth}
        \centering
        \includegraphics{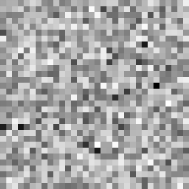}
    \end{subfigure}%
    \hfill
    \begin{subfigure}[t]{0.3\linewidth}
        \centering
        \includegraphics{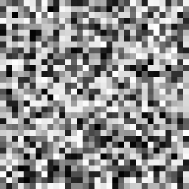}
    \end{subfigure}%
    \hfill
    \begin{subfigure}[t]{0.3\linewidth}
        \centering
        \includegraphics{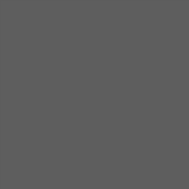}
    \end{subfigure}%
    \hspace*{\fill}%
    \caption{Examples for the Gaussian noise, Uniform noise, and Constant images.}
    \label{ch:results:sec:eval_energy_priornet:subsec:ood_detection:fig:non_natural_data}
\end{figure}

Finally, we generate \gls{ood} datasets with \textit{Noise}, \textit{Constant} samples and an \textit{OODomain} dataset as proposed by \citet{charpentierPosteriorNetworkUncertainty2020}, where the input data is not normalized into the range \( [0, 1] \). We visualize samples in \Cref{ch:results:sec:eval_energy_priornet:subsec:ood_detection:fig:non_natural_data}.
For the \textit{Noise} dataset, we use an equal amount of samples from a standard normal distribution \(N(0, 1)\) and an uniform distribution \(\mathcal{U}(-1, 1)\). The \textit{Constant} dataset is sampled by drawing a scalar from \(\mathcal{U}(-1, 1)\) and then filling a tensor with the same shape as the input data with that value. Finally, \textit{OODomain} inputs are the SVHN dataset and KMNIST dataset, where we apply no normalization for the in-distribution datasets of CIFAR-10 and FMNIST, respectively. As a result, the data is in the range \([0, 255]\).

\paragraph{Baselines.}
We compare \gls{epn} with the baseline of a classifier trained with cross-entropy objective (\textit{CE Baseline}). Further, we compare against several state-of-the-art approaches for \gls{ood} detection also considered in \Cref{chapter:background}: \textit{EnergyOOD}~\cite{liuEnergybasedOutofdistributionDetection2020}, \gls{jem}~\cite{grathwohlYourClassifierSecretly2020} (\textit{JEM}), Outlier-Exposure~\cite{hendrycksDeepAnomalyDetection2019} \textit{OE}, Reverse Prior Networks~\cite{malininReverseKLDivergenceTraining2019} (\textit{R-PriorNet}), \textit{Ensemble}~\cite{lakshminarayananSimpleScalablePredictive2017}, Monte Carlo Dropout~\cite{galDropoutBayesianApproximation2016} (\textit{MC Dropout}), and Posterior Network~\cite{charpentierPosteriorNetworkUncertainty2020} (\textit{PostNet}).

To perform \gls{ood} detection, we use the (inverse) variance of the predictions for sampling-based models (\textit{Ensemble}, \textit{MC Dropout}). \textit{OE} and \textit{PostNet} use the confidence \(\max_y \predictivedist\). For \textit{EnergyOOD}, \textit{JEM} and \gls{epn}, we use the marginal energy \(E_{\params}(\mathbf{x})\) which is equal to the unnormalized probability in \glspl{ebm} \(\umarginal\). Finally, \textit{R-PriorNet} uses the differential entropy \(\mathcal{H}(p(\boldsymbol{\mu} \mid \mathbf{x}, \params))\) to determine distributional uncertainty. We consider differential entropy for \gls{epn} as well to demonstrate that \gls{epn} provides consistent distributional uncertainty estimates.

For all models, we use 5-layer \glspl{mlp} with ReLU activations and hidden layers with dimensionality \(100\) on the tabular datasets and WideResNet-16-8~\cite{zagoruykoWideResidualNetworks2017} for the image datasets. We train each approach with \(5\) different random seeds except for the Ensemble model where we use \(M = 5\) models to form the Ensemble. For \textit{PostNet}, we use aforementioned encoder networks mapping to a latent space of dimensionality \(10\). We then fit Normalizing Flow models with Radial transforms~\cite{rezendeVariationalInferenceNormalizing2016} of depth \(8\) on the latent space. We use a Dropout rate of \(p = 0.3\) for the \textit{MC Dropout} models as recommended in \citet{galDropoutBayesianApproximation2016}. For the models which require an \gls{ood} dataset for training (\textit{EnergyOOD}, \textit{OE}, \textit{R-PriorNet}), we use uniform noise \(\mathcal{U}(\mathbf{0}, \mathbf{1})\). Uniform noise as \gls{ood} dataset ensures a fair comparison since a real \gls{ood} dataset might not be available in practice. We provide additional training details in \Cref{ch:appendix:sec:training_details}.

\begin{table}
    \caption{\gls{aucpr} for \gls{ood} detection on the Segment dataset.}
    \label{ch:results:sec:eval_energy_priornet:subsec:ood_detection:tab:segment_ood}
    \centering
    \begin{tabular}{lllll}
\toprule
           & OOD dataset &                                     Constant &                                       Noise &                                 Segment OOD \\
Model & Score &                                              &                                             &                                             \\
\midrule
CE Baseline & $\max p(y \mid \mathbf{x})$ &             51.27 {\footnotesize $\pm$ 3.88} &            41.05 {\footnotesize $\pm$ 2.49} &             50.6 {\footnotesize $\pm$ 7.04} \\
EnergyOOD & $p_{\boldsymbol{\theta}}(\mathbf{x})$ &             97.74 {\footnotesize $\pm$ 0.37} &  \bfseries{100.0 {\footnotesize $\pm$ 0.0}} &            99.98 {\footnotesize $\pm$ 0.02} \\
Ensemble & Variance &                                       92.95  &             81.5 {\footnotesize $\pm$ 0.47} &                                      74.31  \\
JEM & $p_{\boldsymbol{\theta}}(\mathbf{x})$ &             99.71 {\footnotesize $\pm$ 0.05} &            99.99 {\footnotesize $\pm$ 0.03} &            99.94 {\footnotesize $\pm$ 0.08} \\
MC Dropout & Variance &             93.05 {\footnotesize $\pm$ 0.58} &            55.92 {\footnotesize $\pm$ 2.76} &           52.23 {\footnotesize $\pm$ 11.46} \\
OE & $\max p(y \mid \mathbf{x})$ &             96.46 {\footnotesize $\pm$ 0.05} &            99.99 {\footnotesize $\pm$ 0.01} &            91.44 {\footnotesize $\pm$ 1.21} \\
PostNet & $\max p(y \mid \mathbf{x})$ &  \bfseries{100.0 {\footnotesize $\pm$ 0.01}} &  \bfseries{100.0 {\footnotesize $\pm$ 0.0}} &  \bfseries{100.0 {\footnotesize $\pm$ 0.0}} \\
R-PriorNet & $\mathcal{H}(p(\boldsymbol{\mu} \mid \mathbf{x}, \boldsymbol{\theta}))$ &             98.96 {\footnotesize $\pm$ 0.25} &  \bfseries{100.0 {\footnotesize $\pm$ 0.0}} &             99.93 {\footnotesize $\pm$ 0.1} \\
\midrule
\multirow{2}{*}{EPN-M} & $\mathcal{H}(p(\boldsymbol{\mu} \mid \mathbf{x}, \boldsymbol{\theta}))$ &             99.64 {\footnotesize $\pm$ 0.24} &  \bfseries{100.0 {\footnotesize $\pm$ 0.0}} &            99.51 {\footnotesize $\pm$ 0.69} \\
           & $p_{\boldsymbol{\theta}}(\mathbf{x})$ &             99.97 {\footnotesize $\pm$ 0.03} &  \bfseries{100.0 {\footnotesize $\pm$ 0.0}} &            99.96 {\footnotesize $\pm$ 0.05} \\
\midrule
\multirow{2}{*}{EPN-V} & $\mathcal{H}(p(\boldsymbol{\mu} \mid \mathbf{x}, \boldsymbol{\theta}))$ &   \bfseries{100.0 {\footnotesize $\pm$ 0.0}} &  \bfseries{100.0 {\footnotesize $\pm$ 0.0}} &             93.2 {\footnotesize $\pm$ 0.94} \\
           & $p_{\boldsymbol{\theta}}(\mathbf{x})$ &   \bfseries{100.0 {\footnotesize $\pm$ 0.0}} &  \bfseries{100.0 {\footnotesize $\pm$ 0.0}} &            97.49 {\footnotesize $\pm$ 0.97} \\
\bottomrule
\end{tabular}
\end{table}

\begin{table}
    \caption{\gls{aucpr} for \gls{ood} detection on the FMNST dataset.}
    \label{ch:results:sec:eval_energy_priornet:subsec:ood_detection:tab:fashionmnist_ood}
    \centering
    \setlength{\tabcolsep}{1mm}
    \resizebox{\linewidth}{!}{%
    \begin{tabular}{llllllll}
\toprule
                 & OOD dataset &                                      KMNIST &                                        MNIST &                                     NotMNIST &                                    Constant &                                        Noise &                                    OODomain \\
Model & Score &                                             &                                              &                                              &                                             &                                              &                                             \\
\midrule
CE Baseline & $\max p(y \mid \mathbf{x})$ &            97.54 {\footnotesize $\pm$ 0.17} &             88.35 {\footnotesize $\pm$ 0.38} &             88.56 {\footnotesize $\pm$ 0.86} &             98.15 {\footnotesize $\pm$ 0.5} &              87.07 {\footnotesize $\pm$ 1.6} &            33.71 {\footnotesize $\pm$ 0.24} \\
EnergyOOD & $p_{\boldsymbol{\theta}}(\mathbf{x})$ &  \bfseries{100.0 {\footnotesize $\pm$ 0.0}} &             88.81 {\footnotesize $\pm$ 1.62} &             93.89 {\footnotesize $\pm$ 1.66} &  \bfseries{100.0 {\footnotesize $\pm$ 0.0}} &             82.67 {\footnotesize $\pm$ 1.86} &             30.69 {\footnotesize $\pm$ 0.0} \\
Ensemble & Variance &                                      49.35  &                                       53.03  &                                       48.64  &            65.17 {\footnotesize $\pm$ 20.3} &                                       53.54  &                                      97.98  \\
JEM & $p_{\boldsymbol{\theta}}(\mathbf{x})$ &            48.94 {\footnotesize $\pm$ 3.66} &             78.75 {\footnotesize $\pm$ 9.06} &             79.24 {\footnotesize $\pm$ 6.95} &           45.33 {\footnotesize $\pm$ 17.65} &              67.4 {\footnotesize $\pm$ 11.0} &            95.14 {\footnotesize $\pm$ 6.88} \\
MC Dropout & Variance &            92.12 {\footnotesize $\pm$ 0.79} &             90.87 {\footnotesize $\pm$ 0.74} &             90.54 {\footnotesize $\pm$ 0.25} &            95.82 {\footnotesize $\pm$ 0.52} &             90.88 {\footnotesize $\pm$ 0.56} &             57.12 {\footnotesize $\pm$ 4.6} \\
OE & $\max p(y \mid \mathbf{x})$ &            97.46 {\footnotesize $\pm$ 0.87} &              83.1 {\footnotesize $\pm$ 0.24} &              86.3 {\footnotesize $\pm$ 5.15} &  \bfseries{100.0 {\footnotesize $\pm$ 0.0}} &             90.52 {\footnotesize $\pm$ 1.08} &             32.65 {\footnotesize $\pm$ 0.2} \\
PostNet & $\max p(y \mid \mathbf{x})$ &             97.25 {\footnotesize $\pm$ 0.2} &  \bfseries{95.14 {\footnotesize $\pm$ 1.16}} &             93.18 {\footnotesize $\pm$ 0.76} &            98.93 {\footnotesize $\pm$ 0.72} &             93.45 {\footnotesize $\pm$ 0.14} &  \bfseries{100.0 {\footnotesize $\pm$ 0.0}} \\
R-PriorNet & $\mathcal{H}(p(\boldsymbol{\mu} \mid \mathbf{x}, \boldsymbol{\theta}))$ &             96.7 {\footnotesize $\pm$ 0.67} &             94.52 {\footnotesize $\pm$ 0.64} &  \bfseries{95.98 {\footnotesize $\pm$ 0.28}} &  \bfseries{100.0 {\footnotesize $\pm$ 0.0}} &             93.91 {\footnotesize $\pm$ 2.65} &            71.12 {\footnotesize $\pm$ 5.31} \\
\midrule
\multirow{2}{*}{EPN-M} & $\mathcal{H}(p(\boldsymbol{\mu} \mid \mathbf{x}, \boldsymbol{\theta}))$ &             94.2 {\footnotesize $\pm$ 0.55} &              93.4 {\footnotesize $\pm$ 1.92} &             87.51 {\footnotesize $\pm$ 0.31} &  \bfseries{100.0 {\footnotesize $\pm$ 0.0}} &  \bfseries{97.37 {\footnotesize $\pm$ 0.35}} &  \bfseries{100.0 {\footnotesize $\pm$ 0.0}} \\
                 & $p_{\boldsymbol{\theta}}(\mathbf{x})$ &              79.9 {\footnotesize $\pm$ 0.1} &             91.84 {\footnotesize $\pm$ 3.45} &             78.36 {\footnotesize $\pm$ 3.18} &  \bfseries{100.0 {\footnotesize $\pm$ 0.0}} &             97.03 {\footnotesize $\pm$ 0.37} &  \bfseries{100.0 {\footnotesize $\pm$ 0.0}} \\
\midrule
\multirow{2}{*}{EPN-V} & $\mathcal{H}(p(\boldsymbol{\mu} \mid \mathbf{x}, \boldsymbol{\theta}))$ &            94.97 {\footnotesize $\pm$ 1.23} &             85.46 {\footnotesize $\pm$ 2.57} &              90.51 {\footnotesize $\pm$ 3.1} &  \bfseries{100.0 {\footnotesize $\pm$ 0.0}} &             91.46 {\footnotesize $\pm$ 0.35} &            \bfseries{100.0 {\footnotesize $\pm$ 0.0}} \\
                 & $p_{\boldsymbol{\theta}}(\mathbf{x})$ &            94.31 {\footnotesize $\pm$ 0.28} &             85.53 {\footnotesize $\pm$ 1.32} &             89.69 {\footnotesize $\pm$ 1.72} &  \bfseries{100.0 {\footnotesize $\pm$ 0.0}} &             91.26 {\footnotesize $\pm$ 0.25} &             \bfseries{100.0 {\footnotesize $\pm$ 0.0}} \\
\bottomrule
\end{tabular}
    }
\end{table}

\paragraph{Results.}
We show the AUC-PR for \gls{ood} detection on the Segment dataset in \Cref{ch:results:sec:eval_energy_priornet:subsec:ood_detection:tab:segment_ood} and for the FMNIST dataset in \Cref{ch:results:sec:eval_energy_priornet:subsec:ood_detection:tab:fashionmnist_ood}. We observe that \gls{epn} is able to perform similarly to other methods on these datasets using the marginal density as an \gls{ood} score as well as using differential entropy. 
We provide additional results on the CIFAR-10 and Sensorless dataset in \Cref{ch:appendix:sec:additonal_results}.

Overall, we conclude that indeed the (distributional) uncertainty estimates provided by \gls{epn} are consistent with its density estimates for detecting \gls{ood} samples. This result encourages that the connection, we derived between \glspl{ebm} and \acrlong{dpn} in \Cref{ch:method:sec:combined_view_ebm_dirichlet:eq:marginal_px}, is meaningful. Further, note that \gls{epn} is able to outperform \gls{jem} which does use \gls{ebm} training while lacking the interpretation of logits as concentration parameters of a Dirichlet distribution. Interestingly, \textit{R-PriorNet} is very competitive with \gls{epn} despite using uniform noise as \gls{ood} training data. We expect \gls{epn} to clearly outperform \textit{R-PriorNet} based on using the guided process of \gls{ebm} training instead of randomly sampling. However, in our experiments this advantage shows only on the \textit{OODomain} dataset.

Next, we like to raise attention to the results of \textit{EnergyOOD} in \Cref{ch:results:sec:eval_energy_priornet:subsec:ood_detection:tab:fashionmnist_ood}. \textit{EnergyOOD} uses the same score as \gls{epn}, i.e., the marginal unnormalized probability \(\umarginal\), and is able to perform similarly to \gls{epn} except on the \textit{OODomain} dataset.

\begin{figure}
    \hspace*{\fill}%
    \begin{subfigure}[t]{0.45\linewidth}
        \includegraphics{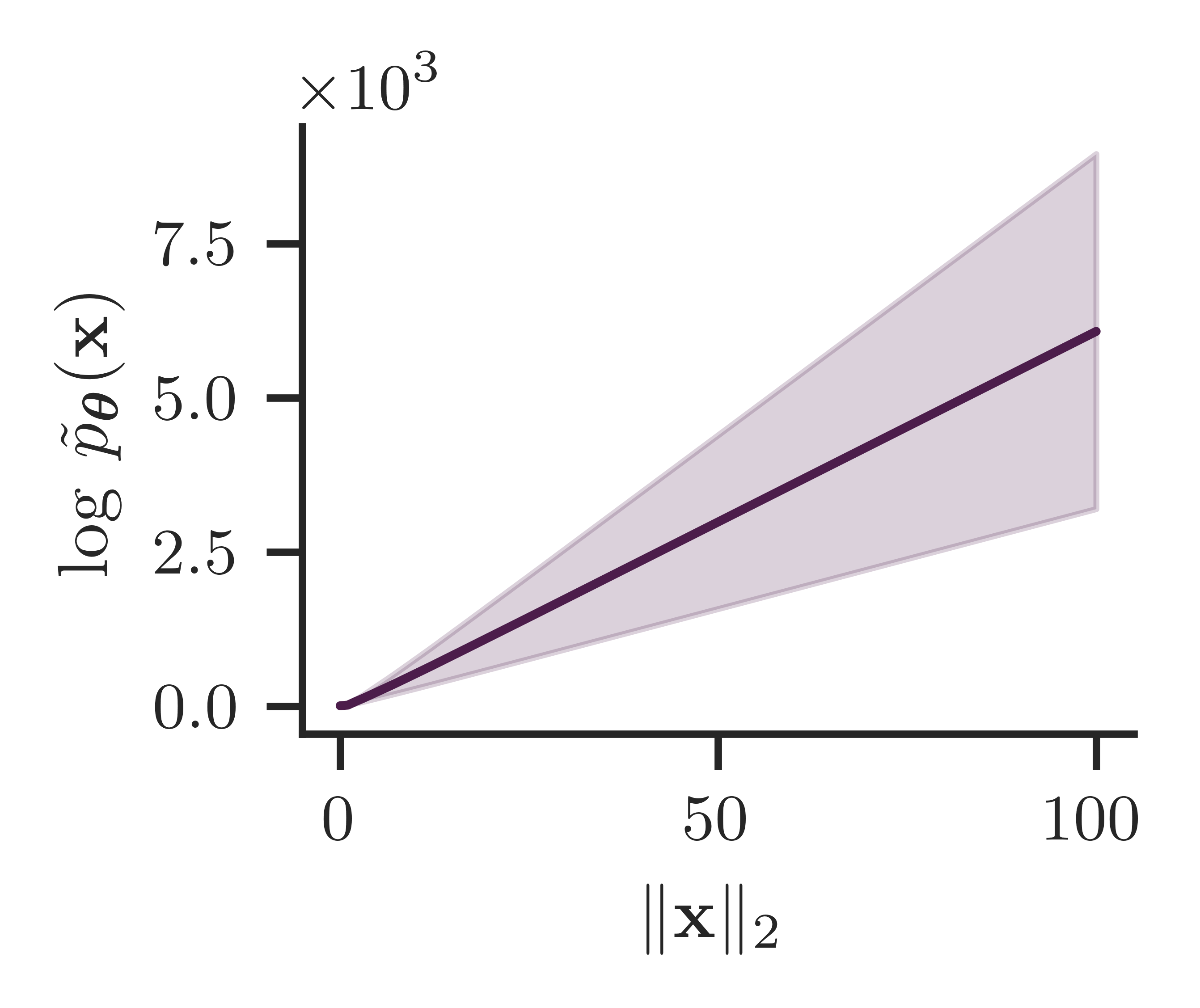}
        \caption{EnergyOOD}
        \label{ch:results:sec:eval_energy_priornet:subsec:ood_detection:fig:energy_ood_behavior:subfig:energy_ood}
    \end{subfigure}%
    \hfill
    \begin{subfigure}[t]{0.45\linewidth}
        \includegraphics{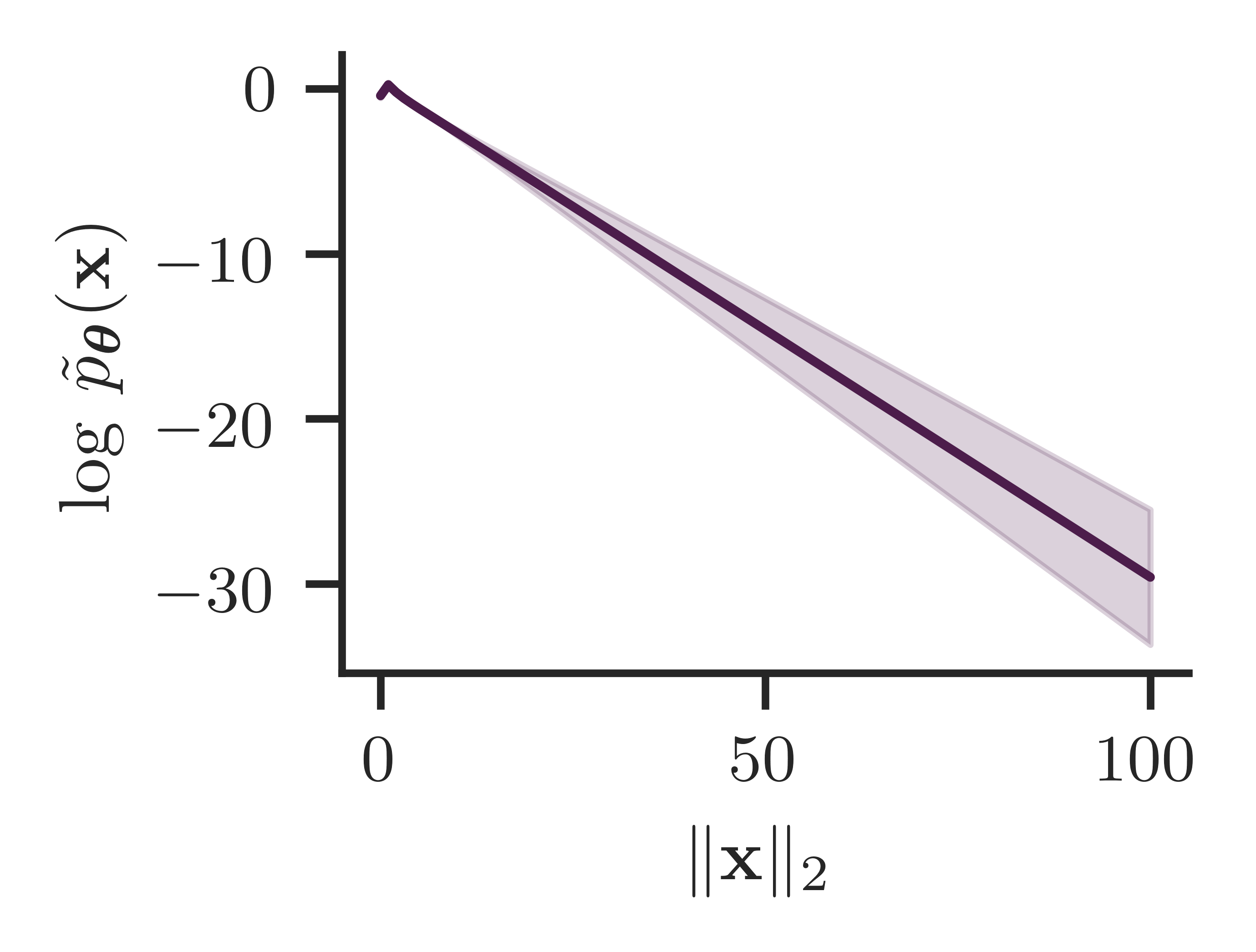}
        \caption{\gls{epn}}
    \end{subfigure}%
    \hspace*{\fill}%
    \caption{\(\umarginal\) for inputs \(\mathbf{x}\) moving away from the training data.}
    \label{ch:results:sec:eval_energy_priornet:subsec:ood_detection:fig:energy_ood_behavior}
\end{figure}

In \Cref{ch:results:sec:eval_energy_priornet:subsec:ood_detection:fig:energy_ood_behavior}, we find that \textit{EnergyOOD} assigns increasing density for inputs moving away from the training data, causing bad performance on the \textit{OODomain} dataset. We hypothesize that the objective in \Cref{ch:background:sec:density_estimation:subsec:energy_ood:eq:regularization} of \textit{EnergyOOD} does not ensure that the energy function converges to the groundtruth density. That is since their objective is motivated from a contrastive learning point of view and is not provably consistent as the \gls{ebm} training approaches in \Cref{ch:background:sec:density_estimation:subsec:ebm:subsubsec:ebm_training}. Further, since the unnormalized probability for \textit{EnergyOOD} increases exponentially in the observed domain in \Cref{ch:results:sec:eval_energy_priornet:subsec:ood_detection:fig:energy_ood_behavior:subfig:energy_ood}, we can assume that \(\umarginal\) increases indefinitely. This means that \textit{EnergyOOD} does not obtain a valid distribution in terms of the learned energy function since the normalizing constant \(Z(\params)\) becomes \(\infty\). Thus, our finding conflicts with the probabilistic arguments in \citet{liuEnergybasedOutofdistributionDetection2020} which assume that a valid distribution \(p_{\params}(\data)\) is learned. 

\begin{figure}
    \centering
    \hspace*{\fill}%
    \begin{subfigure}[t]{0.5\linewidth}
        \centering
        \includegraphics[height=0.5\linewidth]{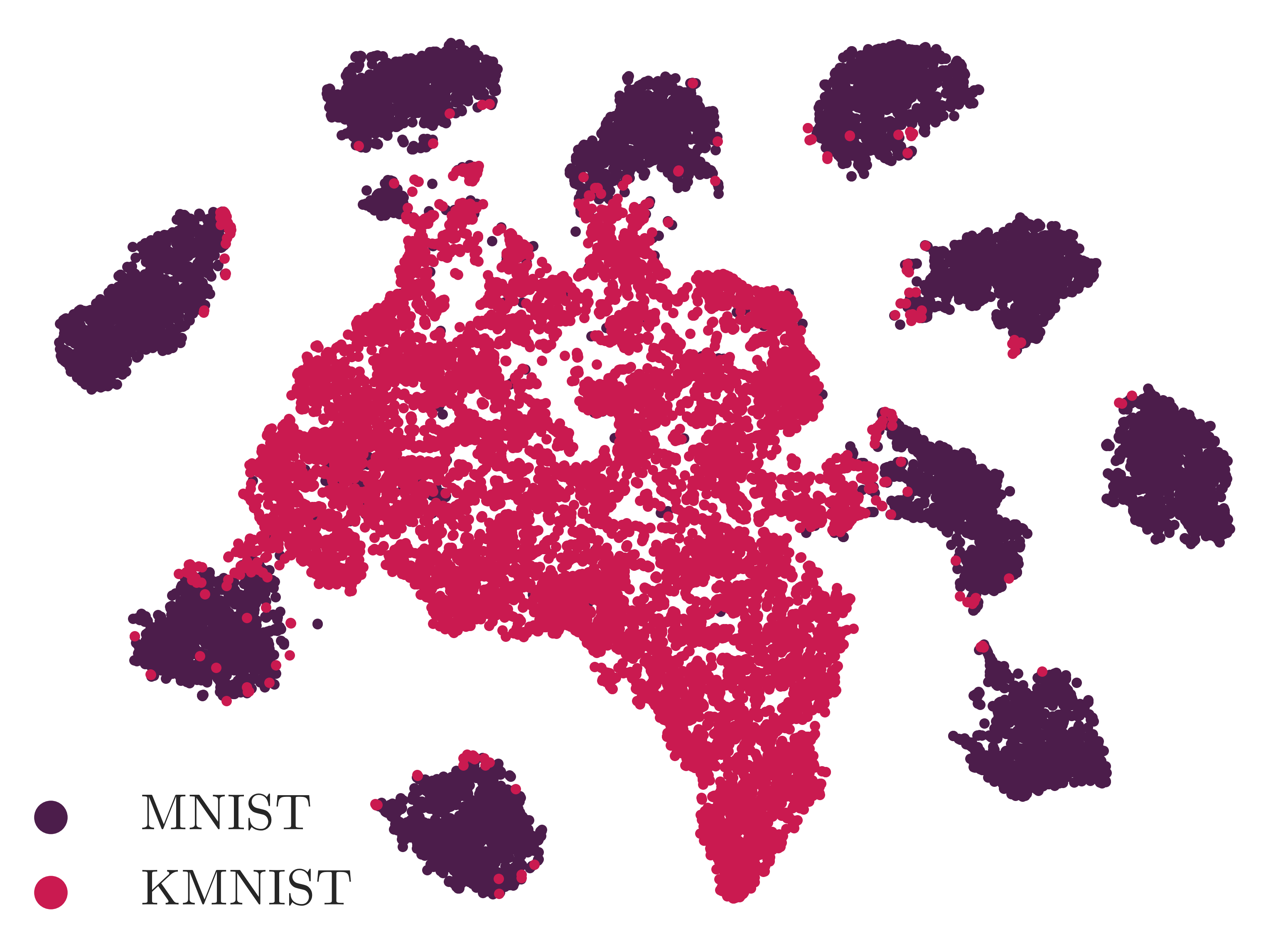}
        \caption{t-SNE embeddings of \gls{epn} features.}
    \end{subfigure}%
    \hfill
    \begin{subfigure}[t]{0.5\linewidth}
        \centering
        \includegraphics[height=0.5\linewidth]{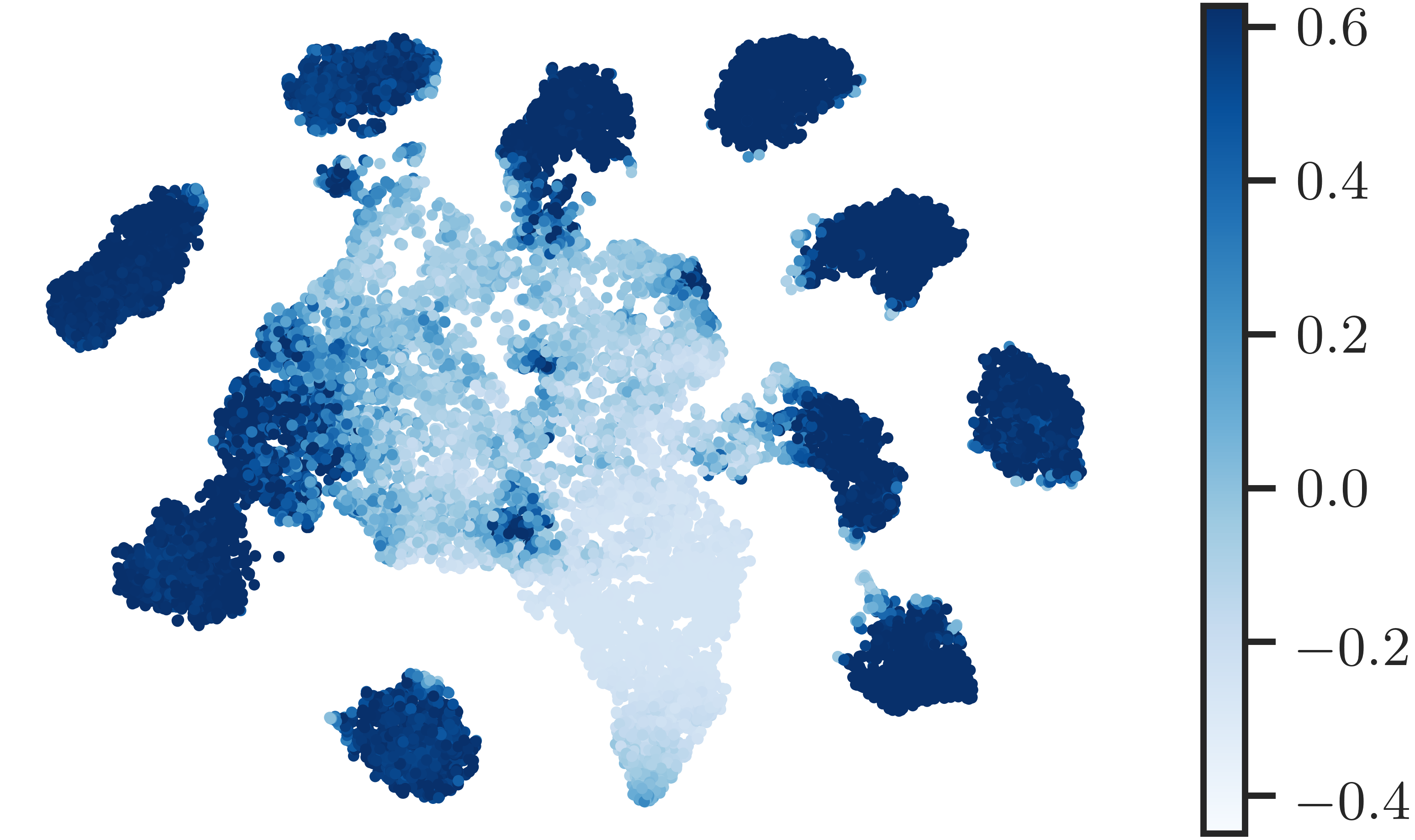}
        \caption{\(\log \umarginal\) of \gls{epn}.}
    \end{subfigure}%
    \hspace*{\fill}%
    \caption{Embeddings and respective density for MNIST vs. KMNIST datasets.}
    \label{ch:results:sec:eval_energy_priornet:subsec:ood_detection:fig:tsne_embeddings}
\end{figure}

Finally, we qualitatively visualize the t-SNE~\cite{maatenVisualizingDataUsing2008} embeddings of the outputs of the penultimate convolutional layer of \gls{epn} together with the estimated unnormalized density in \Cref{ch:results:sec:eval_energy_priornet:subsec:ood_detection:fig:tsne_embeddings}. We observe that the individual classes of MNIST~\cite{lecun2010mnist} successfully cluster, while \gls{epn} assigns lower densities to KMNIST as the \gls{ood} dataset.

\subsection{Detecting dataset shifts}
A significant problem of neural networks in real applications is that the data distribution on which one evaluates the model might change during deployment. This phenomenon is known as \textit{dataset shift} in the machine learning literature. Dataset shift results in degraded performance, however, is hard to detect in practice since the test dataset, which one uses to determine the model to deploy, might not be updated accordingly. Further, these perturbations are often subtle and thus hard to detect.

In order to evaluate the models capability to detect dataset shifts, \citet{hendrycksBenchmarkingNeuralNetwork2019} propose \(15\) dataset shifts based on various augmentations with \(5\) different severity levels of the CIFAR-10 dataset.

\begin{figure}[ht]
    \centering
    \includegraphics{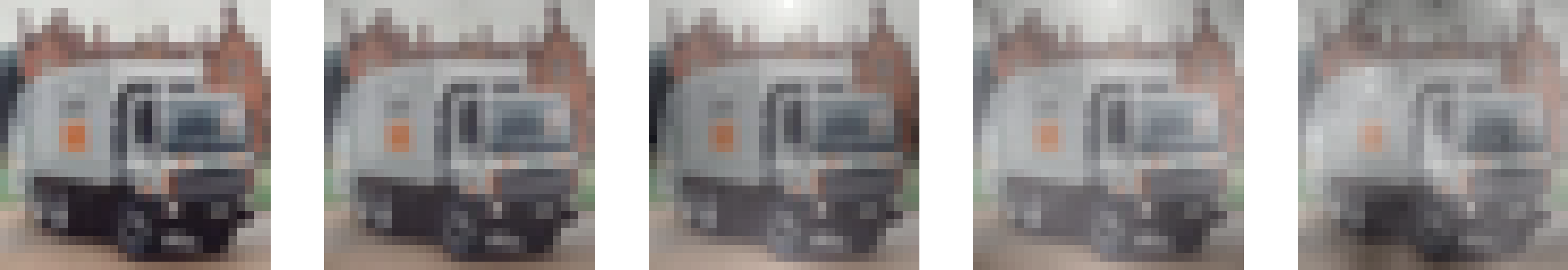}
    \caption{Example of a dataset shift of the \textit{fog} augmentation for severity \(1\) to \(5\).}
    \label{ch:results:sec:eval_energy_priornet:subsec:dataset_shift_eval:fig:dataset_shift_example}
\end{figure}

\begin{figure}
    \hspace*{3mm}%
    \begin{subfigure}[t]{0.4\linewidth}
        \includegraphics{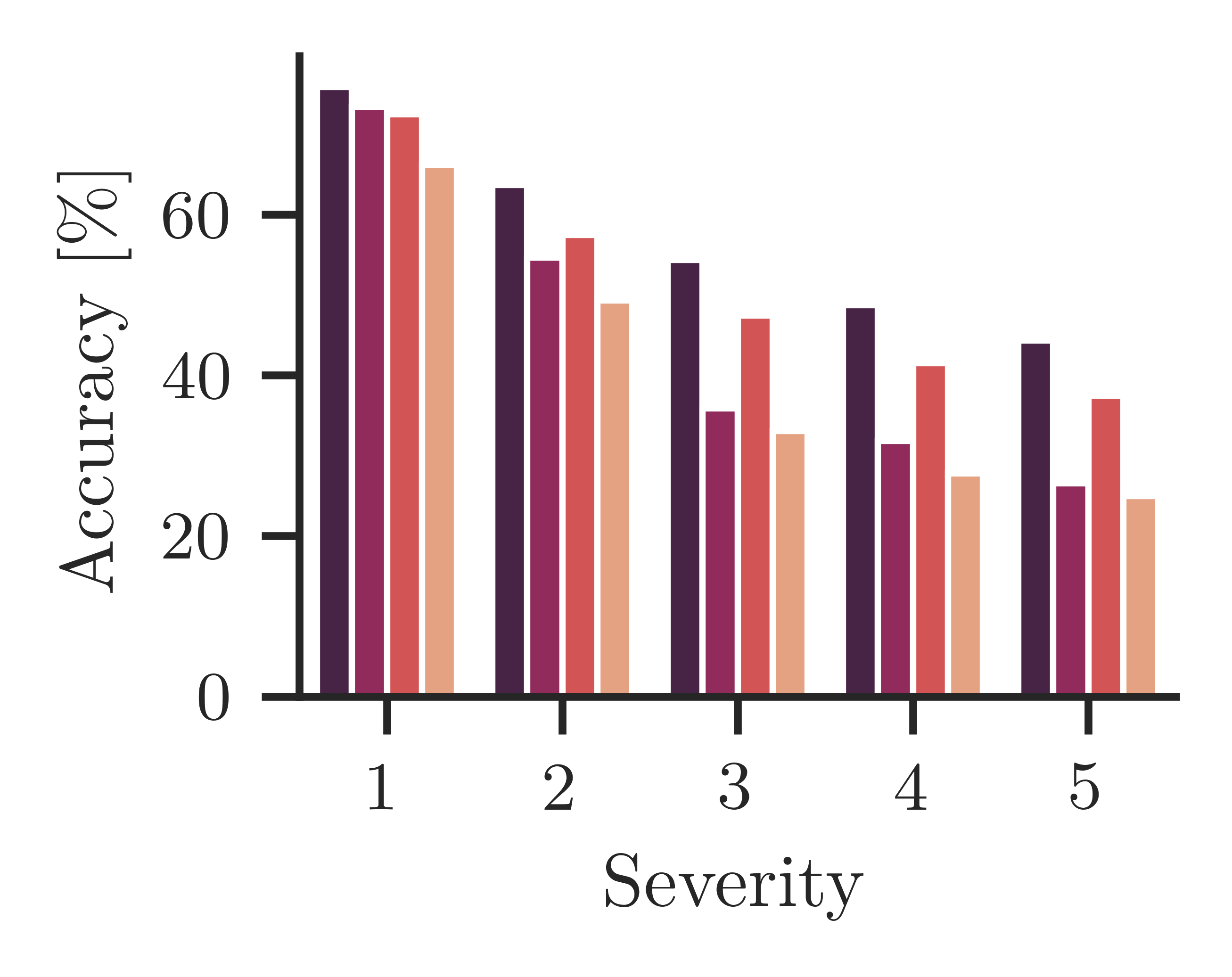}
        \caption{Accuracy under dataset shifts.}
        \label{ch:results:sec:eval_energy_priornet:fig:dataset_shift_eval:subfig:accuracy}
    \end{subfigure}%
    \hfill
    \begin{subfigure}[t]{0.45\linewidth}
        \includegraphics{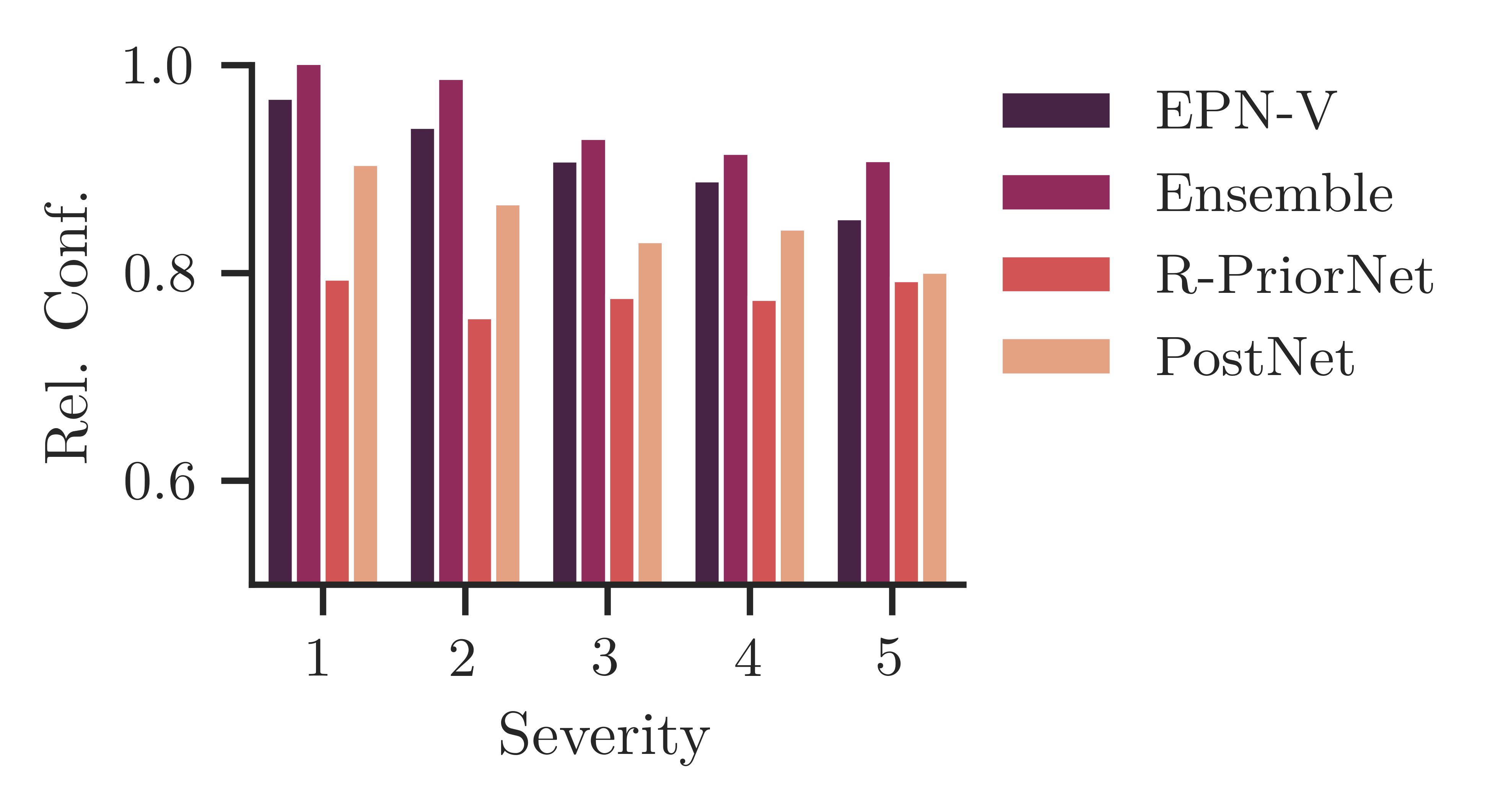}
        \caption{Confidence under dataset shifts.}
        \label{ch:results:sec:eval_energy_priornet:fig:dataset_shift_eval:subfig:confidence}
    \end{subfigure}%
    \hspace*{\fill}%
    \caption{Accuracy and relative confidence for different models under dataset shift.}
    \label{ch:results:sec:eval_energy_priornet:fig:dataset_shift_eval}
\end{figure}

We visualize the accuracy and respective relative confidence of \gls{epn} and various baselines in \Cref{ch:results:sec:eval_energy_priornet:fig:dataset_shift_eval}.  In \Cref{ch:results:sec:eval_energy_priornet:fig:dataset_shift_eval:subfig:accuracy}, we observe that \gls{epn} is more robust than the baselines w.r.t. the dataset shifts since it retains higher accuracy even under augmentations with strong severity. Further, we find that the relative confidence decreases with the increasing severity of the dataset shift in \Cref{ch:results:sec:eval_energy_priornet:fig:dataset_shift_eval:subfig:confidence}.

We conclude that \gls{epn} is able to detect dataset shifts to a similar degree as \textit{PostNet}. In particular, note that \textit{R-PriorNet} does not provide proportional decreases in confidence w.r.t. the accuracy of the model. Concerning the better robustness of the classifier to dataset shifts, we hypothesize that the \gls{ebm} training of \gls{epn} confronts the model not only with clean but also perturbed images leading to better robustness to the perturbations found in the considered dataset shifts.
 
\subsection{Detecting adversarial examples}
\label{ch:results:sec:eval_energy_priornet:subsec:adv_example_detection}
\begin{figure}[ht]
    \centering
    \hspace*{\fill}%
    \begin{subfigure}[t]{0.5\linewidth}
        \centering
        \includegraphics{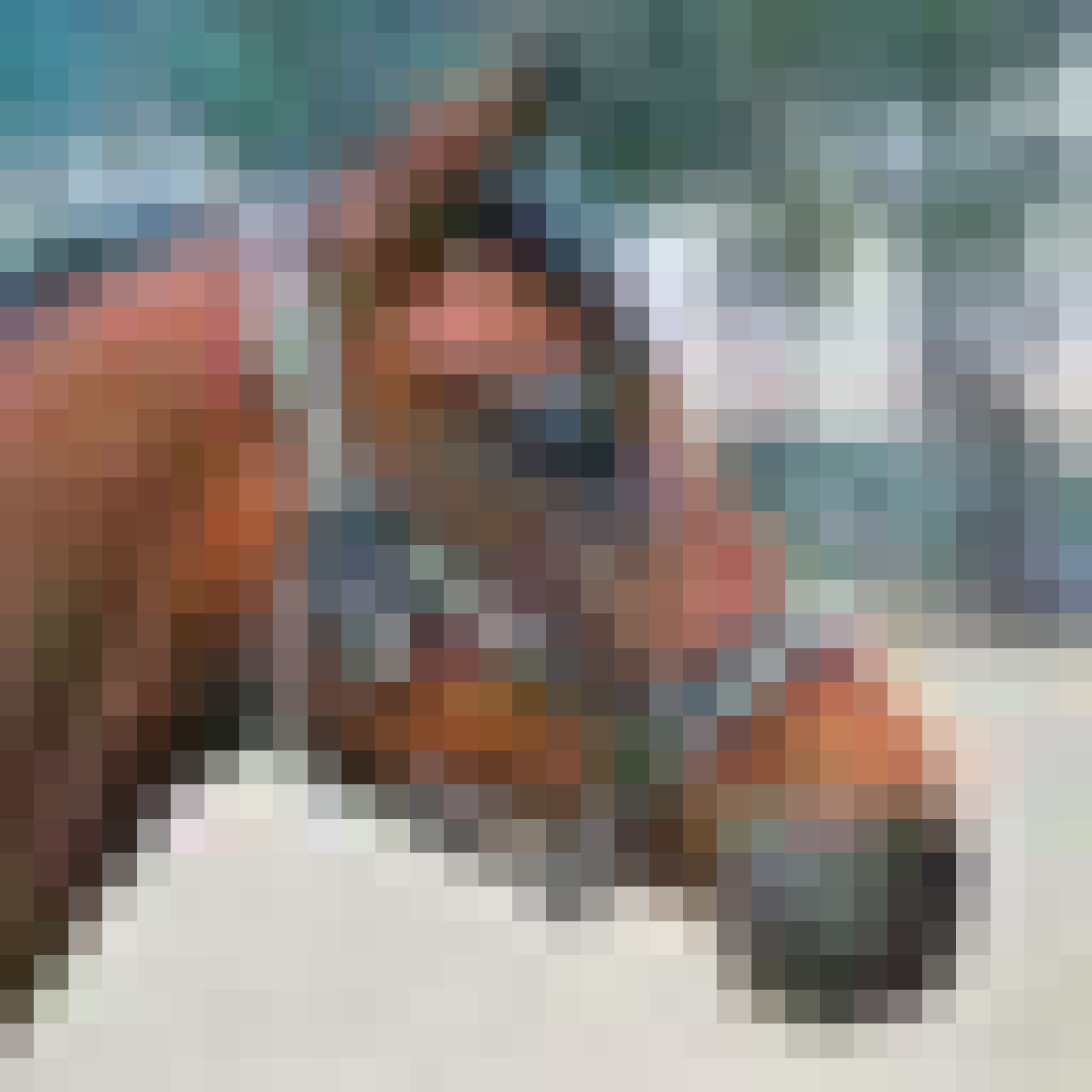}
        \caption{Prediction: \textit{Horse}}
    \end{subfigure}%
    \hfill
    \begin{subfigure}[t]{0.5\linewidth}
        \centering
        \includegraphics{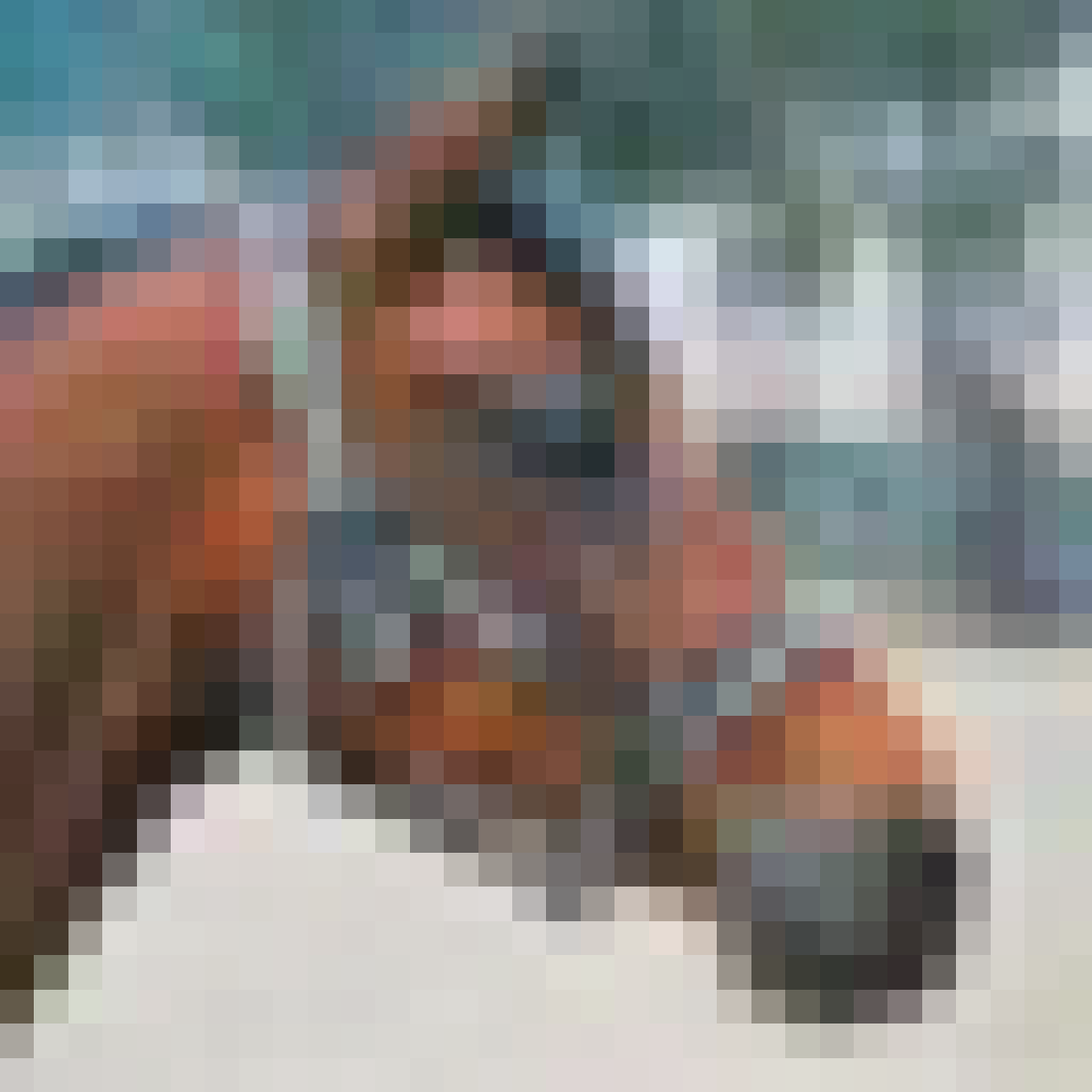}
        \caption{Prediction: \textit{Truck}}
    \end{subfigure}%
    \hspace*{\fill}%
    \caption{Example for adversarial attack with PGD on image (a) transformed to image (b).}
    \label{ch:results:sec:eval_energy_priornet:subsec:adv_example_detection:fig:example}
\end{figure}

Another interesting case is to detect when one specifically changes inputs in a way that the classifier makes wrong predictions. These perturbed inputs, called adversarial examples, are often indistinguishable from real images as shown in \Cref{ch:results:sec:eval_energy_priornet:subsec:adv_example_detection:fig:example} but cause the classifier to make decisions that might be catastrophic in security-critical applications.

We consider two common attacks, \gls{pgd}~\cite{madryDeepLearningModels2019} and \gls{fgm}~\cite{goodfellowExplainingHarnessingAdversarial2015}, with different attack budgets \(\epsilon\). Note that in general, \gls{fgm} is a weaker attack than \gls{pgd}.  In order to investigate whether \gls{epn} can detect these adversarial inputs, we treat the adversarially perturbed images for the \gls{id} dataset as an \gls{ood} dataset and report AUC-PR as before in \Cref{ch:results:sec:eval_energy_priornet:subsec:ood_detection}. Here, we consider \textit{EPN-V} on the CIFAR-10 dataset. 

\begin{figure}
    \begin{subfigure}[t]{0.48\linewidth}
        \includegraphics{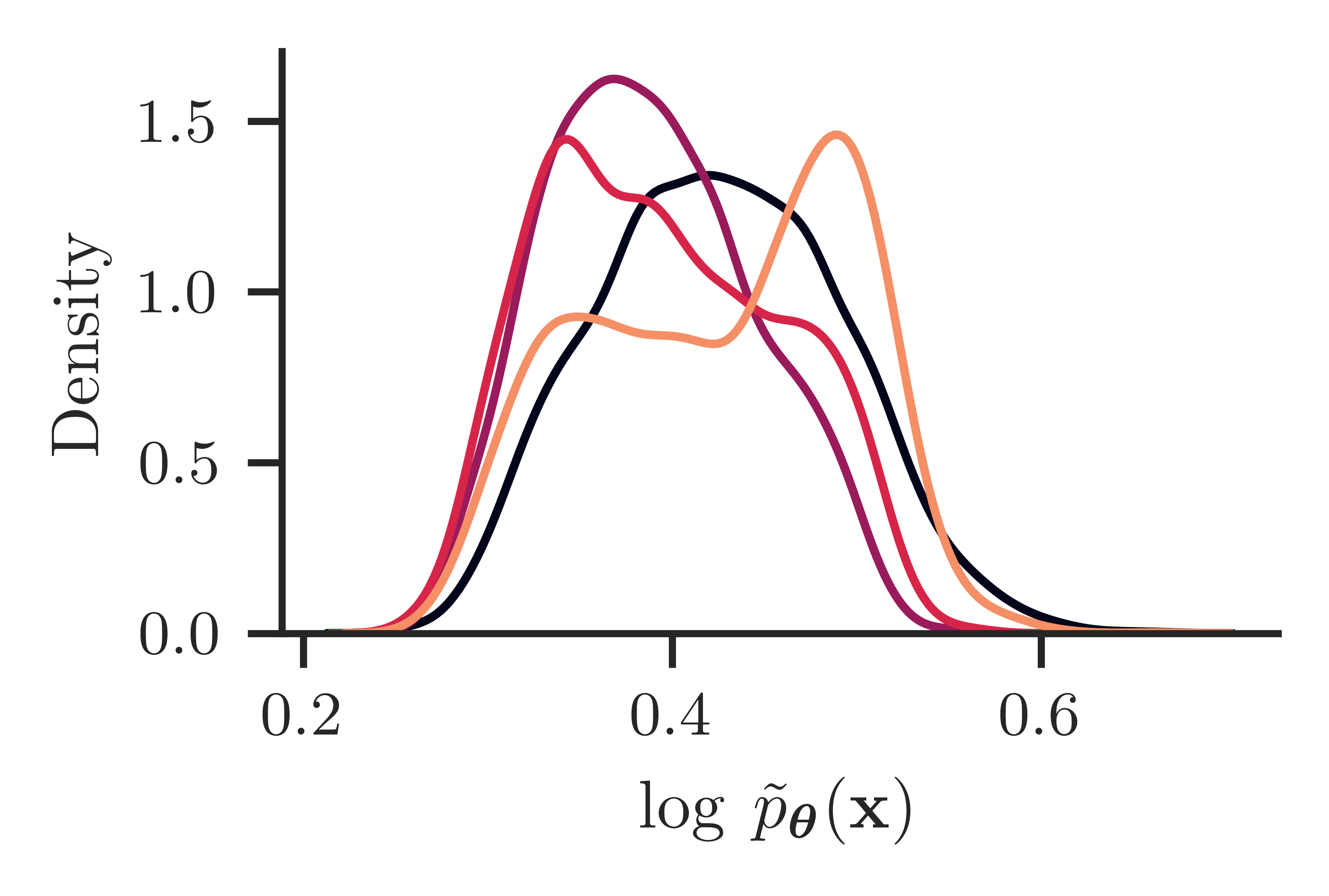}
        \caption{\(L_2\) attack.}
    \end{subfigure}%
    \hfill
    \begin{subfigure}[t]{0.5\linewidth}
        \centering
        \includegraphics{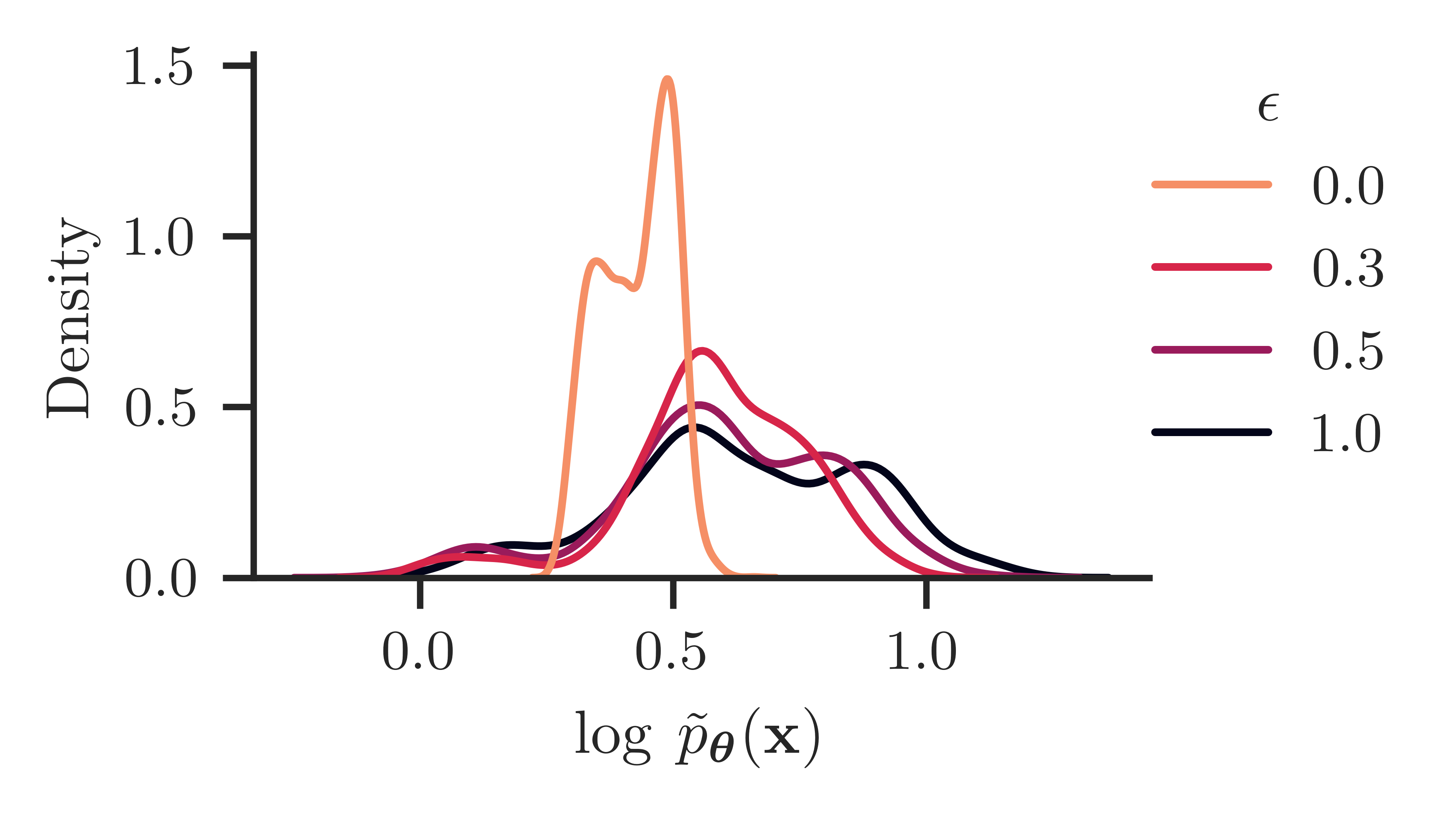}
        \caption{\(L_{\infty}\) attack.}
    \end{subfigure}%
    \hspace*{\fill}%
    \caption{Histograms of unnormalized density \(\umarginal\) for the clean dataset and with \gls{pgd} adversarially attacked samples.}
    \label{ch:results:sec:eval_energy_priornet:subsec:adv_example_detection:fig:histograms}
\end{figure}

\begin{table}
    \caption{AUC-PR for adversarial attacks generated using PGD and FGM.}
    \label{ch:results:sec:eval_energy_priornet:subsec:adv_attack:tab:adv_attack}
    \centering
    \begin{tabular}{lrrrrrr}
\toprule
 & \multicolumn{3}{c}{\(L_\infty\)} & \multicolumn{3}{c}{\(L_2\)} \\
\cmidrule(l){2-4} \cmidrule(l){5-7} 
$\epsilon$ &   0.3 &   0.5 &    1.0 &   0.3 &   0.5 &   1.0 \\
\midrule
FGM    & 95.37 & 99.97 & 100.00 & 64.29 & 69.33 & 70.31 \\
PGD    & 33.82 & 34.64 &  34.59 & 65.30 & 70.76 & 51.61 \\
\bottomrule
\end{tabular}
\end{table}

We show the marginal density \(\umarginal\) for \gls{pgd} attacked images over different attack budgets \(\epsilon\) in \Cref{ch:results:sec:eval_energy_priornet:subsec:adv_example_detection:fig:histograms}. For \(L_2\) attacks, we observe that our model is able to detect attacks with some reliability based on the marginal density of \gls{epn}. However, \(L_\infty\) attacks with increasing \(\epsilon\) lead to an increase in the density \gls{epn} assigns to adversarial samples and therefore adversarial examples become almost undetectable. 

Additionally, we include \gls{aucpr} results for \gls{fgm} and \gls{pgd} in \Cref{ch:results:sec:eval_energy_priornet:subsec:adv_attack:tab:adv_attack}. We find that \gls{epn} is able to provide a signal for detecting adversarial examples in all cases but attacks with \gls{pgd} and \(L_\infty\) norm.

\subsection{Classifier Evaluation}
\label{ch:results:sec:eval_energy_priornet:subsec:classifier_eval}
Next to the \gls{ood} detection capabilities, our model does provide competitive classification performance as well. For this, we evaluate the accuracy and calibration on the four datasets introduced in the previous section. To measure calibration, we consider the \gls{ece}~\cite{guoCalibrationModernNeural2017}. Intuitively, \gls{ece} measures whether a model reports meaningful confidences, e.g., if the model's accuracy is \(70\%\), it reports confidences of \(70\%\) on these samples.

\begin{table}
    \caption{Accuracy (\(\uparrow\)) and calibration in terms of \gls{ece} (\(\downarrow\)).}
    \label{ch:results:sec:eval_energy_priornet:subsec:classifier_eval:tab:acc_calibration}
    \centering
    \resizebox{\linewidth}{!}{%
    \setlength{\tabcolsep}{1mm}
    \begin{tabular}{lllllllll}
\toprule
 & \multicolumn{2}{c}{CIFAR-10} & \multicolumn{2}{c}{FMNIST} & \multicolumn{2}{c}{Segment} & \multicolumn{2}{c}{Sensorless} \\
\cmidrule(l){2-3} \cmidrule(l){4-5} \cmidrule(l){6-7} \cmidrule(l){8-9} 
 &                           Accuracy &                                         ECE &                          Accuracy &                                         ECE &                                     Accuracy &                                         ECE &                                     Accuracy &                                         ECE \\
\midrule
CE Baseline      &   92.26 {\footnotesize $\pm$ 0.03} &             5.69 {\footnotesize $\pm$ 1.16} &  93.99 {\footnotesize $\pm$ 0.18} &              2.3 {\footnotesize $\pm$ 0.27} &             97.22 {\footnotesize $\pm$ 0.36} &             2.39 {\footnotesize $\pm$ 0.98} &              98.6 {\footnotesize $\pm$ 0.15} &             \bfseries{0.45 {\footnotesize $\pm$ 0.14}} \\
EnergyOOD        &   90.85 {\footnotesize $\pm$ 0.42} &             5.52 {\footnotesize $\pm$ 0.78} &  93.82 {\footnotesize $\pm$ 0.02} &             2.45 {\footnotesize $\pm$ 0.49} &              97.47 {\footnotesize $\pm$ 0.0} &              1.73 {\footnotesize $\pm$ 0.5} &              96.32 {\footnotesize $\pm$ 0.5} &              2.21 {\footnotesize $\pm$ 0.0} \\
Ensemble         &                  \bfseries{93.35 } &                                      16.22  &                 \bfseries{94.39 } &                                      16.71  &                                       96.97  &                                      15.73  &                                       99.12  &                                      18.78  \\
JEM         &    88.44 {\footnotesize $\pm$ 0.1} &  \bfseries{2.58 {\footnotesize $\pm$ 0.01}} &  91.75 {\footnotesize $\pm$ 0.12} &             1.23 {\footnotesize $\pm$ 0.08} &             96.72 {\footnotesize $\pm$ 0.36} &             1.72 {\footnotesize $\pm$ 0.01} &             98.33 {\footnotesize $\pm$ 0.09} &              7.2 {\footnotesize $\pm$ 0.03} \\
MC Dropout       &   92.54 {\footnotesize $\pm$ 0.35} &             3.98 {\footnotesize $\pm$ 0.54} &  93.84 {\footnotesize $\pm$ 0.03} &             2.33 {\footnotesize $\pm$ 0.14} &             97.73 {\footnotesize $\pm$ 0.36} &  \bfseries{0.98 {\footnotesize $\pm$ 0.12}} &             96.21 {\footnotesize $\pm$ 0.43} &             2.13 {\footnotesize $\pm$ 0.04} \\
OE &   91.02 {\footnotesize $\pm$ 0.28} &             4.73 {\footnotesize $\pm$ 0.45} &  93.98 {\footnotesize $\pm$ 0.13} &             1.44 {\footnotesize $\pm$ 0.18} &   \bfseries{97.98 {\footnotesize $\pm$ 0.0}} &             1.07 {\footnotesize $\pm$ 0.29} &             99.12 {\footnotesize $\pm$ 0.06} &             0.54 {\footnotesize $\pm$ 0.05} \\
PostNet          &   90.93 {\footnotesize $\pm$ 0.06} &             2.71 {\footnotesize $\pm$ 0.19} &  93.85 {\footnotesize $\pm$ 0.36} &  \bfseries{0.75 {\footnotesize $\pm$ 0.18}} &              95.2 {\footnotesize $\pm$ 0.36} &             27.7 {\footnotesize $\pm$ 6.38} &  \bfseries{99.28 {\footnotesize $\pm$ 0.04}} &             1.08 {\footnotesize $\pm$ 0.16} \\
R-PriorNet       &   91.82 {\footnotesize $\pm$ 0.45} &             83.3 {\footnotesize $\pm$ 0.57} &  93.92 {\footnotesize $\pm$ 0.05} &           72.17 {\footnotesize $\pm$ 20.17} &             96.97 {\footnotesize $\pm$ 0.71} &            84.48 {\footnotesize $\pm$ 2.52} &             98.81 {\footnotesize $\pm$ 0.21} &            91.38 {\footnotesize $\pm$ 0.24} \\
\midrule
EPN-M           &   90.37 {\footnotesize $\pm$ 0.03} &            17.32 {\footnotesize $\pm$ 0.92} &  90.17 {\footnotesize $\pm$ 0.09} &            15.26 {\footnotesize $\pm$ 0.03} &  \bfseries{97.98 {\footnotesize $\pm$ 0.71}} &              1.58 {\footnotesize $\pm$ 0.3} &             97.27 {\footnotesize $\pm$ 0.81} &             2.78 {\footnotesize $\pm$ 0.29} \\
EPN-V            &   83.36 {\footnotesize $\pm$ 0.42} &            68.81 {\footnotesize $\pm$ 0.03} &  91.55 {\footnotesize $\pm$ 0.03} &             76.37 {\footnotesize $\pm$ 0.0} &             97.22 {\footnotesize $\pm$ 0.36} &             4.13 {\footnotesize $\pm$ 0.92} &                         92.75 {\footnotesize $\pm$ 0.18} &  12.52 {\footnotesize $\pm$ 0.47} \\
\bottomrule
\end{tabular}
    }
\end{table}

In \Cref{ch:results:sec:eval_energy_priornet:subsec:classifier_eval:tab:acc_calibration}, we report the accuracy and the calibration in terms of \gls{ece} for \gls{epn} and the baselines models.
Overall, we find that our model provides competitive discriminative results comparable to the baselines. However, \gls{epn} is not calibrated in some cases, even comparing with the baseline cross-entropy model (\textit{CE Baseline}). We investigate this behavior of \gls{epn} in the next section.

\paragraph{Tackling miscalibration.}
We find that underconfidence of \gls{epn} causes large calibration errors on some datasets. The underconfidence is a result of the normalizing constant \(Z(\params )\) of \gls{epn} being a free parameter. That is, \(Z(\params )\) can be scaled which causes the respective energy \(E_{\params}(\mathbf{x})\) to proportionally decrease or the unnormalized probability \(\umarginal{}\) to increase. This scaling affects the precision of the Dirichlet distribution, and the model becomes more or less confident depending on scaling the energy up or down.

One approach proposed in the literature to tackle this problem is temperature scaling~\cite{guoCalibrationModernNeural2017}. For this, \citet{guoCalibrationModernNeural2017} learn a single parameter \(T\) to rescale the logits in the softmax activation. We adapt this approach for our case where the expected predictive distribution in \Cref{ch:method:sec:epn:properties:eq:expected_predictive_distribution} for \(T=1\) is given as 

\begin{align}
    \label{ch:results:sec:eval_energy_priornet:subsec:classifier_eval:eq:temperature_scaled_pred}
    p_{\params}(y \mid \mathbf{x}, \params{}) 
    &= \frac{\alpha^{\text{prior}} + \hat{\alpha}_y}{\sum_{y^\prime} \alpha^{\text{prior}} + \hat{\alpha}_{y^\prime}} \\
    &= \frac{1 + \exp(\nicefrac{f_{\params}(\data)[y]}{T})}{\sum_{y^\prime} 1 + \exp(\nicefrac{f_{\params}(\data)[y^\prime]}{T})}.
\end{align}

In order to learn \(T\), we use the validation set to minimize the negative log-likelihood between the groundtruth labels and the categorical distribution defined in \Cref{ch:results:sec:eval_energy_priornet:subsec:classifier_eval:eq:temperature_scaled_pred}.

\begin{figure}[ht]
    \hspace*{\fill}%
    \begin{subfigure}[t]{0.4\linewidth}
        \includegraphics{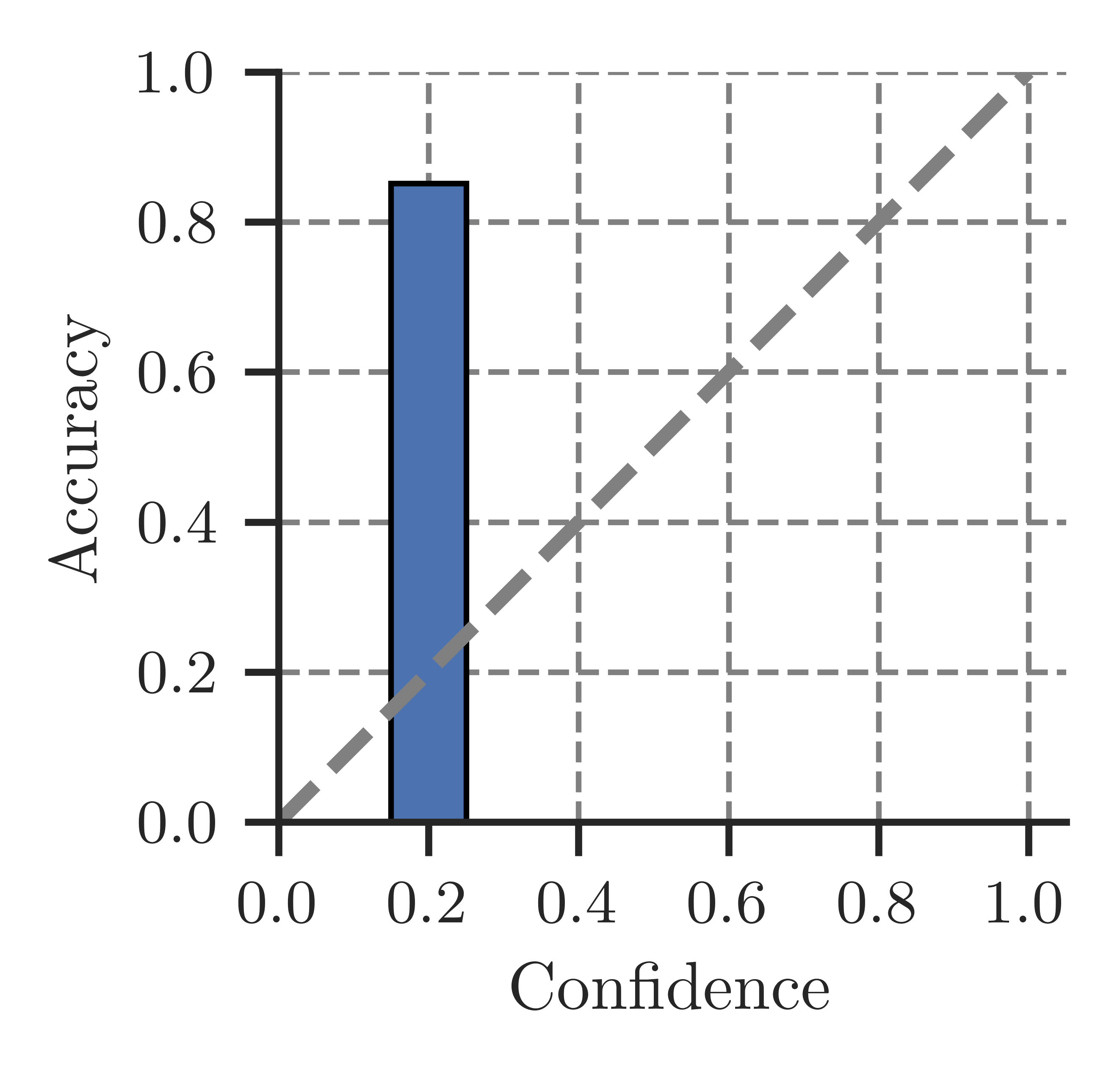}
        \caption{Calibration of \gls{epn}-V.}
    \end{subfigure}%
    \hfill
    \begin{subfigure}[t]{0.4\linewidth}
        \includegraphics{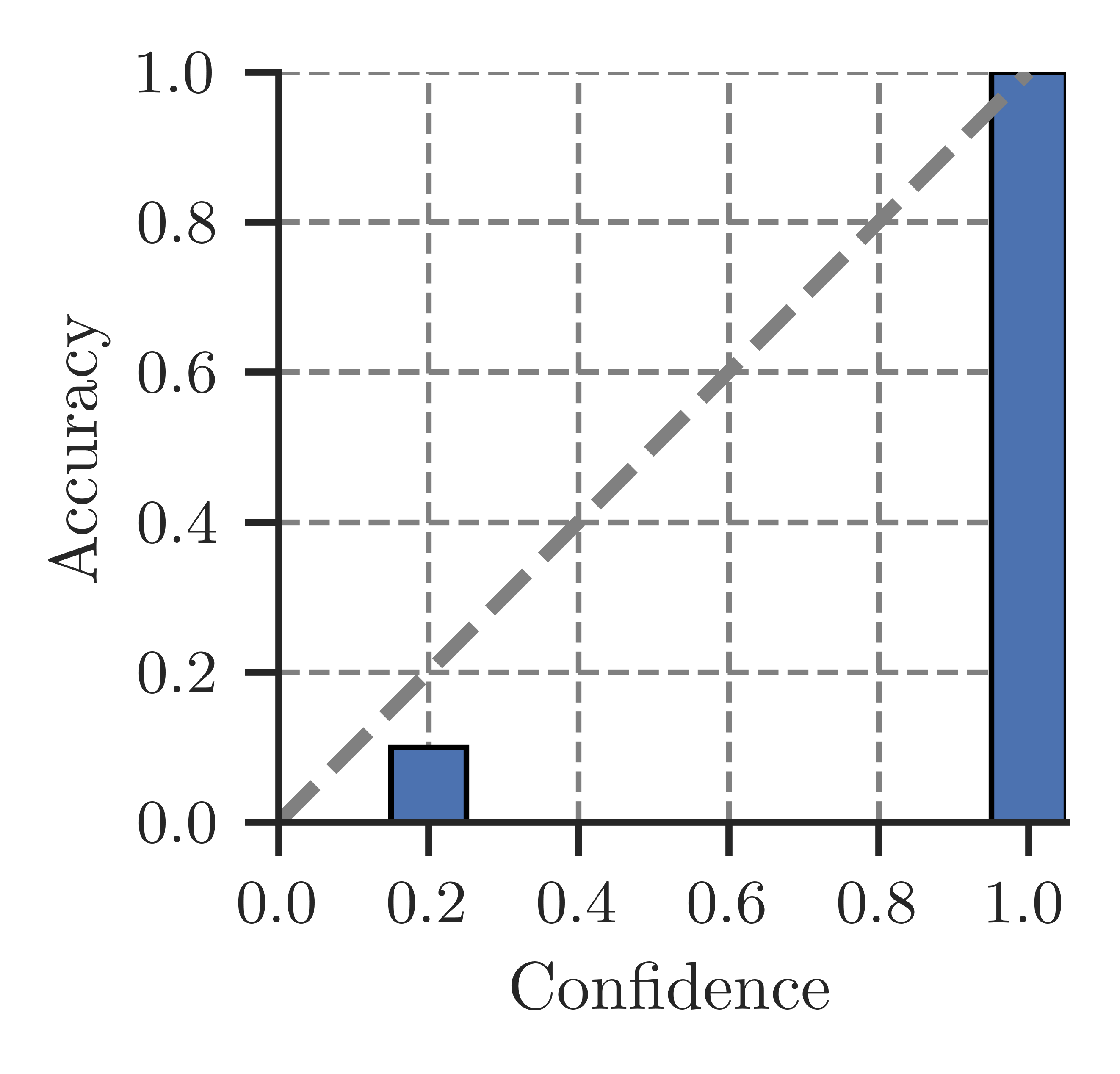}
        \caption{Calibration after temperature scaling.}
    \end{subfigure}%
    \hspace*{\fill}%
    \caption{Calibration curves of \gls{epn}-V before and after temperature scaling.}
    \label{ch:results:sec:eval_energy_priornet:subsec:classifier_eval:fig:ece_plots}
\end{figure}

In \Cref{ch:results:sec:eval_energy_priornet:subsec:classifier_eval:fig:ece_plots}, we find that temperature scaling resolves the issues of underconfidence of \gls{epn} on the CIFAR-10 dataset.

\subsection{Ablations}
To determine the merit of individual components of \gls{epn}, we consider several ablation experiments on the Sensorless dataset. 

In \Cref{ch:results:sec:eval_energy_priornet:subsec:classifier_eval:tab:ablations}, we report the accuracy as well as \gls{ood} detection performance of \gls{epn} when omitting terms of the loss function proposed in \Cref{ch:method:sec:energy-priornet:subsec:training}. We observe that when omitting the optimization of the \(p_{\params}(\mathbf{x})\) term, the \gls{ood} detection severely degrades. Similarly, \gls{epn} is not able to perform classification without the \gls{kl} term optimizing \(p_{\params}(y \mid \mathbf{x})\) introduced in \Cref{ch:method:sec:energy-priornet:eq:overall_objective}. Further, we see small gains in \gls{ood} detection when leveraging additional supervised information on the \textit{Sensorless OOD} dataset. We expand on this finding in \Cref{ch:results:sec:ood_detection_ebms}.

\begin{table}
    \caption{Ablation experiments on the Sensorless dataset. Gray cells denote significant drops in performance.}
    \label{ch:results:sec:eval_energy_priornet:subsec:classifier_eval:tab:ablations}
    \centering
    \begin{tabular}{lllll}
\toprule
 &                          Classification         &                           \multicolumn{3}{c}{OOD Detection} \\
\cmidrule(l){2-2} \cmidrule(l){3-5}
 &                          Accuracy &                          Constant &                             Noise &                    Sensorless OOD \\
\midrule
No \(p_{\params}(\mathbf{x})\)  &   95.24 {\footnotesize $\pm$ 3.6} &  \cellcolor{lightgray} 44.75 {\footnotesize $\pm$ 2.79} & \cellcolor{lightgray} 42.07 {\footnotesize $\pm$ 5.38} &  \cellcolor{lightgray} 78.89 {\footnotesize $\pm$ 3.83} \\
No \(p_{\params}(y \mid \mathbf{x})\) & \cellcolor{lightgray} 10.81 {\footnotesize $\pm$ 0.58} &   100.0 {\footnotesize $\pm$ 0.0} &   100.0 {\footnotesize $\pm$ 0.0} &  94.71 {\footnotesize $\pm$ 0.64} \\
\midrule
Full model        &  93.21 {\footnotesize $\pm$ 4.21} &   99.99 {\footnotesize $\pm$ 0.0} &   99.99 {\footnotesize $\pm$ 0.0} &  95.57 {\footnotesize $\pm$ 1.87} \\
\bottomrule
\end{tabular}
\end{table}

\subsection{Conclusion}
In this section, we evaluated our proposed model \gls{epn}. We found that \gls{epn} can detect \gls{ood} samples with high accuracy, sometimes outperforming baselines. Further, we found that \gls{epn} can identify dataset shifts and adversarial examples with some fidelity. Finally, we evaluated the discriminative capabilities of \gls{epn} and found competitive results with the baselines in this setting. Additionally, we uncovered an issue with underconfident predictions of \gls{epn} on complex datasets, which we were able to resolve in parts through post-hoc calibration.

\section{Investigation of OOD Detection with EBMs}
\label{ch:results:sec:ood_detection_ebms}

In the second part of the experiments, we shift our focus from a particular model to an investigation of the general properties that allow superior \gls{ood} detection with \glspl{ebm}. This is interesting since while \glspl{ebm}, such as \gls{epn} proposed in this work and \gls{jem}~\cite{grathwohlYourClassifierSecretly2020}, have shown good \gls{ood} detection performance based on the learned marginal density \(p(\mathbf{x})\), other generative models, e.g., Normalizing Flows, struggle with that task.

In particular, \citet{kirichenkoWhyNormalizingFlows2020, schirrmeisterUnderstandingAnomalyDetection2020} showed that Normalizing Flows are incapable of differentiating ID and \gls{ood} data on high-dimensional datasets due to learning low-level features and those features being present in all natural images. To this end we provide similar analysis for the generative model class of \glspl{ebm}  

In particular, we verify the following hypotheses improving \gls{ood} detection with \glspl{ebm} in recent works~\cite{grathwohlYourClassifierSecretly2020, grathwohlNoMCMCMe2020} compared to Normalizing Flows:

\begin{itemize}
    \item \textbf{Dimensionality reduction.} The manifold hypotheses~\cite{feffermanTestingManifoldHypothesis2013} states that high dimensional data such as images reside on a low dimensional manifold. Normalizing Flows require invertible transformations and thus operate in the original data space. We hypothesize that this hinders \gls{ood} detection as they need to model off-manifold directions. Contrarily, there is no such restriction imposed on \glspl{ebm} which allows to prune redundant dimensions without semantic content. \\ 
    \item \textbf{Supervision.} \citet{kirichenkoWhyNormalizingFlows2020, schirrmeisterUnderstandingAnomalyDetection2020} show that Normalizing Flows learn low-level features without semantic meaning (smoothness, etc.). We hypothesize that label information encourages semantic, high-level features improving \gls{ood} detection.
\end{itemize}

\paragraph{Datasets.} We use the datasets as described in \Cref{ch:results:sec:eval_energy_priornet:subsec:ood_detection}. For the analysis in this section, we split the \gls{ood} datasets into groups according to their characteristics: \textit{natural} and \textit{non-natural} \gls{ood} datasets. Natural datasets include the data with samples from a similar domain as the \gls{id}, e.g., images from classes the model was not trained on. On the other hand, \textit{non-natural} datasets contain irregular data patterns, e.g., noise. Note that the differentiation of \textit{natural} and \textit{non-natural} datasets allows evaluating distinct properties of the learned density: A model able to distinguish \textit{natural} inputs can recognize semantic features of the high-level content of images, e.g., corresponding to classes, while \textit{non-natural} inputs are easily detected semantically but lie farther away from the data manifold, thus, require the model to decrease the density when moving away from the data distribution.\\

\textbf{Architectures.}
We use a 5-layer \gls{mlp} with ReLU activations and hidden layers with dimensionality on the tabular datasets and WideResNet-10-2~\cite{zagoruykoWideResidualNetworks2017} for the image datasets.  For the Normalizing Flow models on the image datasets, we use Glow~\cite{kingmaGlowGenerativeFlow2018} with \(L=3\) layers, \(K=32\) steps, and \(C=512\) channels. For the tabular datasets, we use \(20\) stacked radial transforms~\cite{rezendeVariationalInferenceNormalizing2016}.




\subsection{Are EBMs better than baselines in general?}
\label{ch:results:sec:ood_detection_ebms:subsec:ebms_better_than_nf}
\paragraph{Experiment 1.}
We establish baseline results by training \glspl{ebm} and baseline models.
%
In \Cref{ch:results:sec:ood_detection_ebms:tab:all_results}, we find that \glspl{ebm} consistenly outperform the \textit{CE} baseline by \(62.9\%\), \(55.0\%\), and \(36.4\%\) for \textit{CD}, \textit{VERA}, and \textit{SSM} respectively. The improvements are moderate in comparison to the Normalizing Flow baselines with \(11.9\%\), \(4.3\%\), and \(-4.3\%\) respectively. Notably, improvements are mostly on \textit{natural} datasets. 

As \glspl{ebm} perform dimensionality reduction since  the \gls{ebm} is specified by a function $\mathbb{R}^D \mapsto \mathbb{R}$ mapping to the scalar energy and do not consistently outperform Normalizing Flows in this experiment across all training methods, we conclude that dimensionality reduction plays a minor role in the \gls{ood} detection performance of recent EBMs~\cite{grathwohlYourClassifierSecretly2020}. Slight improvements on \textit{natural} data can be attributed to the ability to discard non-semantic dimensions in \glspl{ebm}. 
%

\begin{table}
    \centering
    \caption{AUC-PR for \gls{ood} detection on the respective \gls{id} datasets.}
    \resizebox{\linewidth}{!}{%
    \setlength{\tabcolsep}{1mm}
    \begin{tabular}{lllllllllll}
\toprule
ID & \multicolumn{5}{c}{CIFAR-10} & \multicolumn{3}{c}{FMNIST} &                                      \multicolumn{1}{c}{Segment} &                         \multicolumn{1}{c}{Sensorless} \\
\cmidrule(l){2-6} \cmidrule(l){7-9} \cmidrule(l){10-10} \cmidrule(l){11-11} 
OOD &                                    CIFAR-100 &                            CelebA &                                         LSUN &                                          SVHN &                                      Textures &                                        KMNIST &                                        MNIST &                           NotMNIST &                                  OOD &                     OOD \\
\midrule
CE    &  \bfseries{62.76 {\footnotesize $\pm$ 1.46}} &  64.47 {\footnotesize $\pm$ 2.44} &             65.18 {\footnotesize $\pm$ 5.79} &              47.51 {\footnotesize $\pm$ 4.58} &              39.17 {\footnotesize $\pm$ 2.28} &              69.07 {\footnotesize $\pm$ 6.73} &  \bfseries{82.5 {\footnotesize $\pm$ 12.27}} &    50.9 {\footnotesize $\pm$ 6.73} &             33.35 {\footnotesize $\pm$ 1.82} &   33.02 {\footnotesize $\pm$ 1.32} \\
NF    &                                       58.34  &                 \bfseries{74.68 } &                                       62.99  &                                        31.58  &                                        50.23  &                                        62.22  &                                       49.03  &                  \bfseries{93.68 } &  \bfseries{99.45 {\footnotesize $\pm$ 0.18}} &                  \bfseries{94.35 } \\
\midrule
CD  &             50.51 {\footnotesize $\pm$ 2.13} &  43.86 {\footnotesize $\pm$ 5.85} &            54.43 {\footnotesize $\pm$ 11.37} &  \bfseries{60.72 {\footnotesize $\pm$ 24.59}} &  \bfseries{76.21 {\footnotesize $\pm$ 17.44}} &              50.52 {\footnotesize $\pm$ 9.39} &              31.69 {\footnotesize $\pm$ 0.9} &   76.85 {\footnotesize $\pm$ 2.66} &             98.18 {\footnotesize $\pm$ 2.18} &  72.83 {\footnotesize $\pm$ 16.19} \\
SSM   &             53.82 {\footnotesize $\pm$ 3.12} &   57.72 {\footnotesize $\pm$ 7.0} &             52.79 {\footnotesize $\pm$ 3.16} &              45.75 {\footnotesize $\pm$ 7.24} &              48.82 {\footnotesize $\pm$ 4.34} &              58.98 {\footnotesize $\pm$ 5.48} &             67.86 {\footnotesize $\pm$ 11.4} &  57.27 {\footnotesize $\pm$ 13.73} &            79.43 {\footnotesize $\pm$ 24.29} &  67.14 {\footnotesize $\pm$ 20.31} \\
VERA  &             55.95 {\footnotesize $\pm$ 2.68} &  73.97 {\footnotesize $\pm$ 2.63} &  \bfseries{67.39 {\footnotesize $\pm$ 2.57}} &              37.27 {\footnotesize $\pm$ 4.66} &               46.29 {\footnotesize $\pm$ 8.1} &  \bfseries{78.11 {\footnotesize $\pm$ 21.05}} &            67.53 {\footnotesize $\pm$ 21.63} &  76.22 {\footnotesize $\pm$ 22.11} &             94.63 {\footnotesize $\pm$ 7.22} &  45.66 {\footnotesize $\pm$ 10.55} \\
\bottomrule
\end{tabular}
    }
    \resizebox{\linewidth}{!}{%
    \setlength{\tabcolsep}{1mm}
    \begin{tabular}{lllllllllll}
\toprule
ID & \multicolumn{3}{c}{CIFAR-10} & \multicolumn{3}{c}{FMNIST} & \multicolumn{2}{c}{Segment} & \multicolumn{2}{c}{Sensorless} \\
\cmidrule(l){2-4} \cmidrule(l){5-7} \cmidrule(l){8-9} \cmidrule(l){10-11} 
OOD &                                      Constant &                                       Noise &                                     OODomain &                           Constant &                                       Noise &                                    OODomain &                                    Constant &                                       Noise &                                    Constant &                                       Noise \\
\midrule
CE    &               45.26 {\footnotesize $\pm$ 8.8} &           61.13 {\footnotesize $\pm$ 21.02} &              30.69 {\footnotesize $\pm$ 0.0} &    35.5 {\footnotesize $\pm$ 3.08} &           55.84 {\footnotesize $\pm$ 22.32} &            30.74 {\footnotesize $\pm$ 0.11} &           41.74 {\footnotesize $\pm$ 18.57} &            33.66 {\footnotesize $\pm$ 2.77} &            32.38 {\footnotesize $\pm$ 1.19} &            31.97 {\footnotesize $\pm$ 1.26} \\
NF    &                                        30.87  &            83.65 &                                            \bfseries{100.0} &                  \bfseries{71.07 } &            98.04 &                                           \bfseries{100.0} &            99.95  &  \bfseries{100.0} &                           \bfseries{100.0 } &  \bfseries{100.0} \\
\midrule
CD  &  \bfseries{58.75 {\footnotesize $\pm$ 28.17}} &  \bfseries{100.0 {\footnotesize $\pm$ 0.0}} &            58.41 {\footnotesize $\pm$ 37.96} &  70.59 {\footnotesize $\pm$ 12.84} &  \bfseries{100.0 {\footnotesize $\pm$ 0.0}} &  \bfseries{100.0 {\footnotesize $\pm$ 0.0}} &            95.47 {\footnotesize $\pm$ 2.34} &            95.14 {\footnotesize $\pm$ 3.71} &  \bfseries{100.0 {\footnotesize $\pm$ 0.0}} &  \bfseries{100.0 {\footnotesize $\pm$ 0.0}} \\
SSM   &             47.24 {\footnotesize $\pm$ 15.56} &           70.28 {\footnotesize $\pm$ 31.39} &  68.57 {\footnotesize $\pm$ 25.2} &  47.57 {\footnotesize $\pm$ 15.18} &           49.45 {\footnotesize $\pm$ 21.19} &            76.76 {\footnotesize $\pm$ 21.1} &           73.91 {\footnotesize $\pm$ 25.44} &            81.1 {\footnotesize $\pm$ 17.87} &            69.79 {\footnotesize $\pm$ 6.62} &           64.61 {\footnotesize $\pm$ 17.35} \\
VERA  &              31.51 {\footnotesize $\pm$ 0.66} &  \bfseries{100.0 {\footnotesize $\pm$ 0.0}} &            63.48 {\footnotesize $\pm$ 34.37} &  53.24 {\footnotesize $\pm$ 22.65} &           79.34 {\footnotesize $\pm$ 27.34} &           72.42 {\footnotesize $\pm$ 37.61} &  \bfseries{100.0 {\footnotesize $\pm$ 0.0}} &  \bfseries{100.0 {\footnotesize $\pm$ 0.0}} &  \bfseries{100.0 {\footnotesize $\pm$ 0.0}} &            99.85 {\footnotesize $\pm$ 0.31} \\
\bottomrule
\end{tabular}
    }
    \label{ch:results:sec:ood_detection_ebms:tab:all_results}
\end{table}

\begin{figure}
    \centering
    \includegraphics{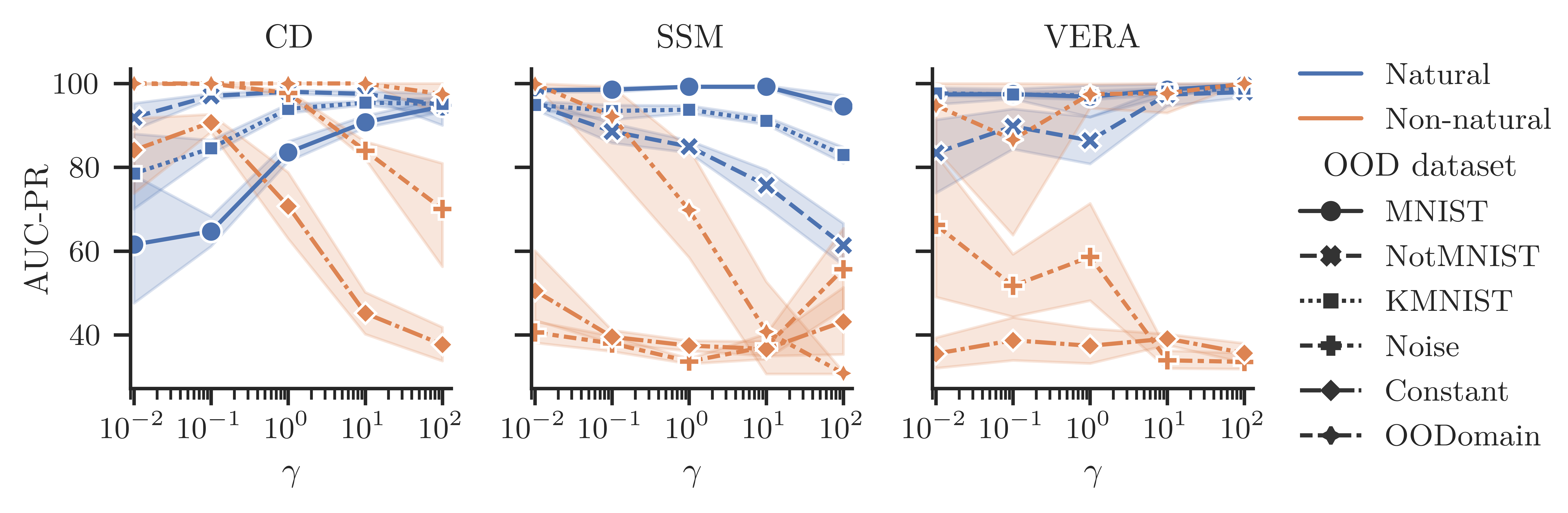}
     \caption{AUC-PR for \gls{ood} detection for different settings of the weighting hyperparameter \(\gamma\) of the cross entropy objective. We use FMNIST as the \gls{id} dataset.}
     \label{fig:fashionmnist_clf_weight}
\end{figure}

\begin{figure}
    \centering
    \includegraphics{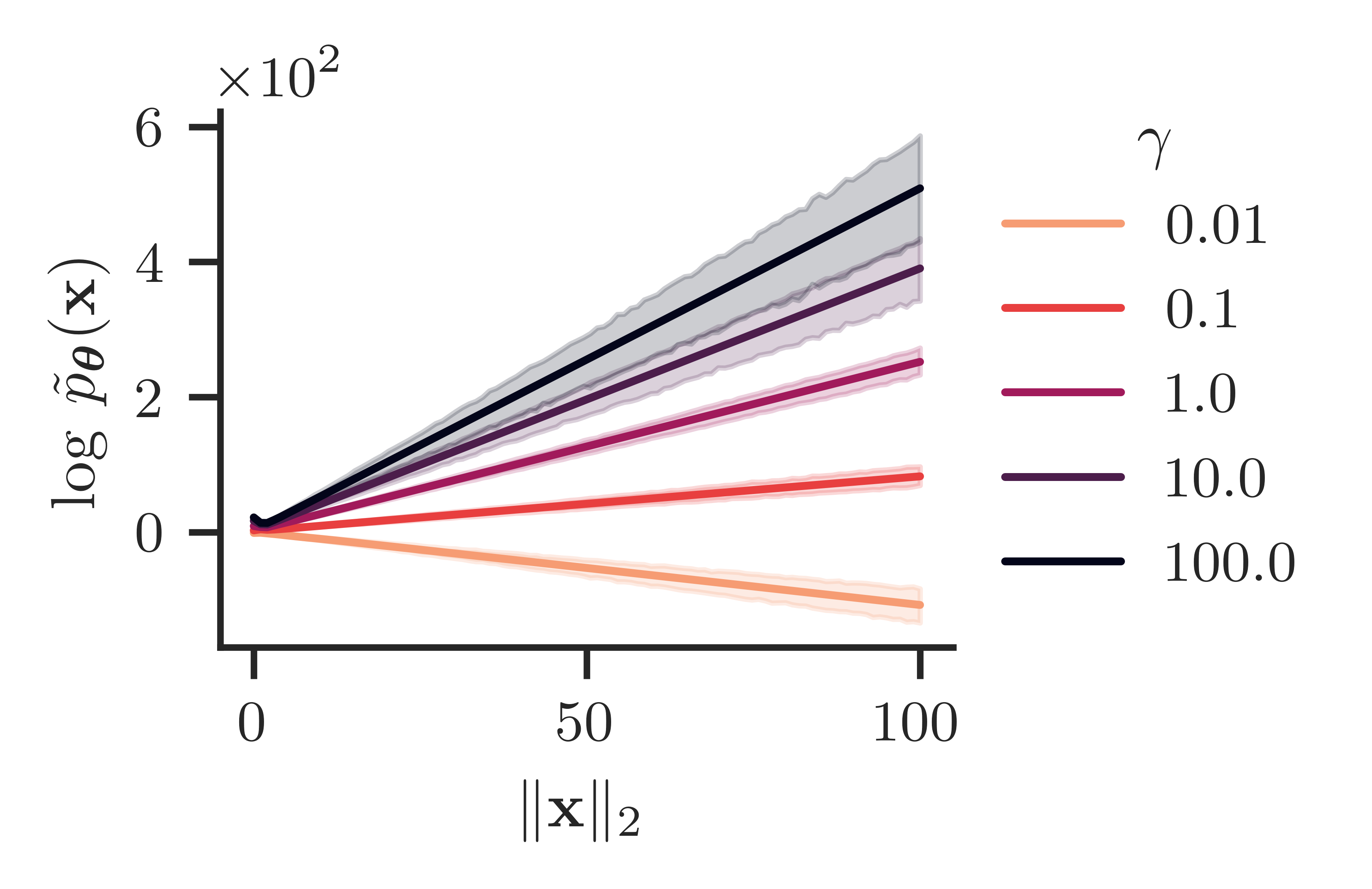}
     \caption{Unnormalized density \(\tilde{p}_{\params}(\data)\) of inputs with increasing $L_2$-norm computed with models trained with \gls{ssm} and supervision with different \(\gamma\).
     }
     \label{fig:ssm_moving_away_from_training_data}
\end{figure}
\subsection{Does supervision improve OOD detection?}
\label{ch:results:sec:ood_detection_ebms:subsec:supervision_improves_ood}
Next, we consider two ways of incorporating labels to investigate the influence of supervision. Firstly, by applying an additional loss term as in~\cite{grathwohlYourClassifierSecretly2020} which affects the optimization directly, and secondly, performing density estimation on embeddings of a classification model which incorporates supervision indirectly through class-related features.
\paragraph{Experiment 2.}
We consider \glspl{jem} as introduced in \Cref{ch:background:sec:density_estimation:subsec:ebm} and apply a cross entropy objective with weighting hyperparameter \(\gamma\) optimizing \(p_{\params}(y \mid \data)\).

In \Cref{tab:supervision_improvement}, we find substantial improvements in \gls{ood} detection on most datasets compared to the baseline models. Using label information within the model encourages discriminative features relevant for classification improving detection of \textit{natural}, \gls{ood} inputs by $29.61\%$. These results indicate that \gls{ebm} training tends to assign high-likelihood to all natural, structured images, an issue also observed in other generative models~\cite{renLikelihoodRatiosOutofDistribution2019}. Note however that supervision decreases performance on some \textit{natural} datasets and consistently worsens results at differentiating \textit{non-natural} inputs ($-11.74\%$). 
%
%
\begin{table}
    \centering
    \caption{\% relative improvement in AUC-PR for \gls{ood} detection when using additional supervision during training.}
    \label{tab:supervision_improvement}
    \begin{tabular}{llrr}
\toprule
 Model    &ID dataset &  Natural &  Non-natural \\
\midrule
\multirow{4}{*}{CD} & CIFAR-10 &   -10.82 &      -9.11 \\
     & FMNIST &    47.17 &       3.24 \\
     & Segment &     1.85 &       0.89 \\
     & Sensorless &    29.72 &      -0.02 \\
\midrule
\multirow{4}{*}{SSM} & CIFAR-10 &     7.33 &     -27.94 \\
     & FMNIST &    50.61 &     -20.26 \\
     & Segment &    25.89 &     -21.94 \\
     & Sensorless &    22.13 &     -40.73 \\
\midrule
\multirow{4}{*}{VERA} & CIFAR-10 &    -1.16 &      -3.00 \\
     & FMNIST &    33.66 &     -15.53 \\
     & Segment &     4.98 &      -0.57 \\
     & Sensorless &    97.93 &       0.07 \\
\bottomrule
\end{tabular}

\end{table}

\paragraph{Experiment 3.}
Investigating the difference in results between \textit{natural} and \textit{non-natural} datasets, we observe in \Cref{fig:fashionmnist_clf_weight} that the \gls{ood} detection on \textit{non-natural} images is negatively impacted by increasing the weighting \(\gamma\) of the cross-entropy objective. Additionally, we find that setting \(\gamma\) too high degrades performance at detecting \textit{natural} \gls{ood} data in some cases as well. Further, we observe in \Cref{fig:ssm_moving_away_from_training_data} that \glspl{jem} assign exponentially increasing density to data points far from the training data distribution for higher settings of \(\gamma\), similar to results for the confidence in ReLU networks~\cite{heinWhyReLUNetworks2019}. 

As a result of the increasing density, \textit{non-natural} inputs, which are farther from the training data than \textit{natural} images, become increasingly harder to detect. 
Similar to the results of \textit{EnergyOOD} in \Cref{ch:results:sec:eval_energy_priornet:subsec:ood_detection}, we conclude that since the density increases indefinitely, the learned energy function probably \textbf{does not describe a valid distribution at all} as the normalizing constant \(Z(\params)\) in the \gls{ebm} becomes \(\infty\).
Intuitively, we hypothesize that this is caused by the cross-entropy objective "focusing" on the decision boundary only (see \Cref{ch:results:sec:eval_energy_priornet:fig:ce_baseline_toy_confidence}). That is, since \(L_{\text{CE}} = \sum_y \log p_{\params}(y \mid \data) \) is independent of \(p_{\params}(\data)\). Thus, higher values of \(\lambda\) lead to relative less importance in optimizing \(p_{\params}(\data)\) which is important for reliable \gls{ood} detection. This fact is demonstrated when comparing \gls{epn} with the cross-entropy baseline in \Cref{ch:results:sec:eval_energy_priornet:subsec:toy_dataset}. 

Overall, we conclude that training with this factorization requires tuning of \(\gamma\) to achieve high \gls{ood} detection performance on both \textit{natural} and \textit{non-natural} inputs.

%

\paragraph{Tackling asymptotic behavior.}
In \Cref{ch:method:sec:epn:subsec:asymptotic_behavior}, we found that we can use a slightly adapted architecture to enforce increasing energy for \glspl{ebm} far from the training data by ensuring that the last weight matrix has component-wise negative weights. Thus, we follow \citet{meinkeProvablyRobustDetection2021} and parameterize the elements of the weight matrix as \((W)_{ck} = -e^{(V)_{ck}}\) where \(V \in \mathbb{R}^{C\times K}\). 

\begin{figure}
    \centering
    \includegraphics{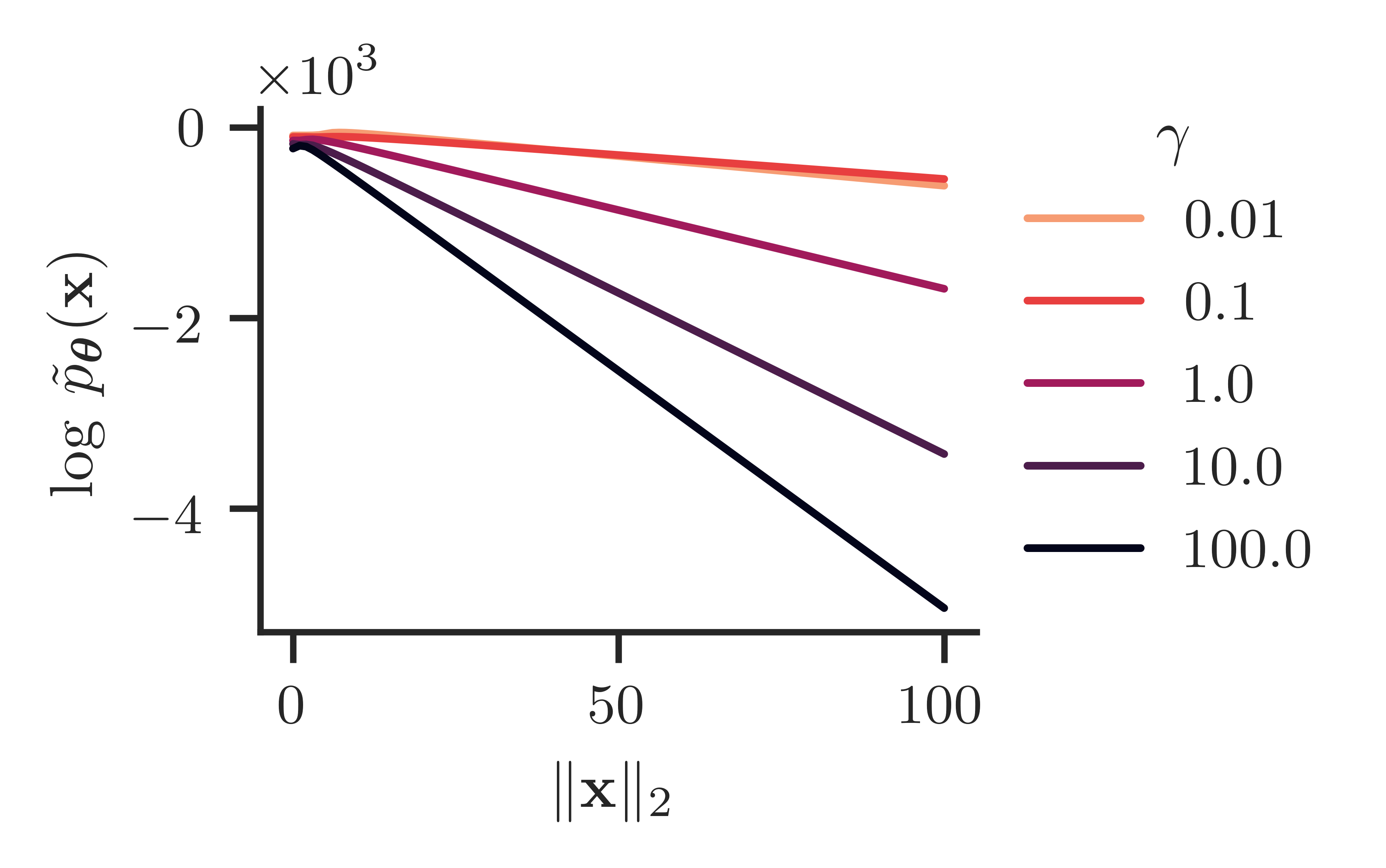}
     \caption{Unnormalized density \(\tilde{p}_{\params}(\data)\) of inputs with increasing $L_2$-norm for \glspl{ebm} with the adjusted architecture trained with different values of \(\gamma\).
     }
    \label{ch:results:sec:ood_detection_ebms:subsec:supervision_improves_ood:fig:ssm_far_from_training_data_neg_linear}
\end{figure}

We repeat \textit{Experiment 3} and train the adjusted model with different settings of the weighting hyperparameter \(\gamma\). In \Cref{ch:results:sec:ood_detection_ebms:subsec:supervision_improves_ood:fig:ssm_far_from_training_data_neg_linear}, we find that our adjustments indeed leads to decaying density far from the training data for all values of \(\gamma\) compared to the prior results in \Cref{fig:ssm_moving_away_from_training_data}.   

The guaranteed asymptotic convergence also contributes to the \gls{ood} detection results. In \Cref{tab:supervision_improvement_neg_linear}, we find that training the adjusted model with \gls{ssm} and supervision leads to large, consistent improvements on the \textit{non-natural} \gls{ood} data compared the model not leveraging \Cref{theorem1} in \Cref{tab:supervision_improvement}. We explain this result with the intuition that \textit{non-natural} data lies further away from the training data, thus, the asymptotic behavior plays a more important role.

Surprisingly, the \gls{ood} detection on CIFAR-10 improves with the adjusted model as well which cannot be explained by the asymptotic behavior alone. Thus, we hypothesize that the constraint, that enforces the asymptotic behavior, introduces some regularization into the \gls{ebm} training. We leave the investigation of this observation for future work.


\begin{table}
    \centering
    \caption{\% relative improvement in AUC-PR for \gls{ood} detection with SSM training when using supervision and the parameterization with asymptotic guarantee.}
    \label{tab:supervision_improvement_neg_linear}
    \begin{tabular}{llrr}
\toprule
Model & ID dataset &  Natural &  Non-natural \\
\midrule
\multirow{2}{*}{SSM} & CIFAR-10   &    40.96 &      48.34 \\
                     & FMNIST     &    50.48 &      78.07 \\
\bottomrule
\end{tabular}

\end{table}

\paragraph{Experiment 4.}
As an alternative, we sidestep the issue of tuning \(\gamma\) by following \citet{kirichenkoWhyNormalizingFlows2020} who noticed that training Normalizing Flows on high-level features improves \gls{ood} detection. 
As \glspl{ebm} do not require architectural restrictions, they should be able to extract these semantic features without additional processing. 
To investigate this behavior for \glspl{ebm}, we store the features from a classifier trained with cross-entropy objective after convolutional layers. Subsequently, we train \glspl{ebm} on these embeddings.
Note that the dimensionality of the embeddings is not significantly lower than that of the input images. As a result, one cannot expect significant contributions from the dimensionality reduction capabilities of \glspl{ebm}. For FMNIST, we obtain embeddings of size \(640\) while the original dimensionality is \(28 \times 28 = 784\).

In \Cref{tab:embedding_improvemnt}, we observe that density estimation on embeddings significantly improves results on \textit{natural} datasets compared to the baseline trained on images directly ($+53.65\%$). Further, performance on \textit{non-natural} datasets does not deteriorate and even increases by $10.98\%$ on average with this approach.

As training on discriminative features directly improves \gls{ood} detection, this supports our hypotheses that \glspl{ebm} trained on high-dimensional data such as images struggle to learn semantic features by themselves. 

\begin{table}
    \centering
    \caption{\% relative improvement in AUC-PR for \gls{ood} detection when training on embeddings.}
    \label{tab:embedding_improvemnt}
    \begin{tabular}{llrr}
\toprule
 Model    & ID dataset &  Natural & Non-natural \\
\midrule
\multirow{2}{*}{CD} & CIFAR-10 &    48.60 &       3.37 \\
     & FMNIST &    95.79 &     -13.52 \\
\midrule
\multirow{2}{*}{SSM} & CIFAR-10 &    53.84 &      -2.31 \\
     & FMNIST &    58.40 &      59.59 \\
\midrule
\multirow{2}{*}{VERA} & CIFAR-10 &    50.16 &      16.97 \\
     & FMNIST &    15.12 &       1.80 \\
\bottomrule
\end{tabular}

\end{table}

\begin{table}
    \centering
    \caption{\% relative improvement in AUC-PR for \gls{ood} detection after introducing bottlenecks.}
    \label{tab:dim_red_improvement}
    \begin{tabular}{llrr}
\toprule
 Model    & ID dataset &  Natural &  Non-natural \\
\midrule
\multirow{2}{*}{CD} & CIFAR-10 &    20.18 &        20.38 \\
     & FMNIST &    67.95 &        10.88 \\
\midrule
\multirow{2}{*}{SSM} & CIFAR-10 &    14.76 &        33.34 \\
     & FMNIST &     1.75 &        -5.92 \\
\midrule
\multirow{2}{*}{VERA} & CIFAR-10 &    19.66 &        33.22 \\
     & FMNIST &    26.84 &        32.94 \\
\bottomrule
\end{tabular}

\end{table}

\subsection{Can we encourage semantic features?}
\label{ch:results:sec:ood_detection_ebms:subsec:can_encourage_semantic_features}

While \glspl{ebm} inherently perform dimensionality reduction, the previous experiments suggest that this is insufficient to capture semantic features within the data. 
As shown by \citet{kirichenkoWhyNormalizingFlows2020}, introducing a bottleneck in the coupling transforms of Normalizing Flows enforces the network to learn semantic features improving \gls{ood} detection. One can also interpret this change in the frame of compression~\cite{serraInputComplexityOutofdistribution2020}, i.e., one removes redundant information.

\paragraph{Experiment 5.}
To study the effect, we introduce bottlenecks after every block of the Wide-ResNet through a set of $\smash{1\times1}$ convolutions mapping to $0.2 \times$ the original dimensionality. We experiment with different values of the downscaling factor and find $0.2$ to yield the best results. However, other settings did improve \gls{ood} detection as well. We present the full results for other choices of the bottleneck factor in \Cref{tab:bottleneck_full}.

In \Cref{tab:dim_red_improvement}, we observe that this simple adjustment yields improvements in \gls{ood} detection on \textit{natural} images for all training methods. 

The bottlenecks force the network to compress the features, removing redundant information. Thus, the bottlenecks enable improved \gls{ood} detection and support the hypotheses that generic \glspl{ebm} retain non-semantic features. We provide further investigation on low-level features in \Cref{sec:additional_results:low_level_features}.

\subsection{Low-level features in EBMs}
\label{sec:additional_results:low_level_features}
In \Cref{ch:results:sec:ood_detection_ebms:subsec:supervision_improves_ood} and \Cref{ch:results:sec:ood_detection_ebms:subsec:can_encourage_semantic_features}, we argue that supervision and bottlenecks encourage semantic features, while unsupervised \glspl{ebm} learn generic local pixel correlations (low-level features) common to all \textit{natural} images as shown by \citet{schirrmeisterUnderstandingAnomalyDetection2020} which results in worse \gls{ood} detection.

\begin{figure}[ht]
    \centering
    \hspace*{\fill}%
    \includegraphics[width=0.12\linewidth]{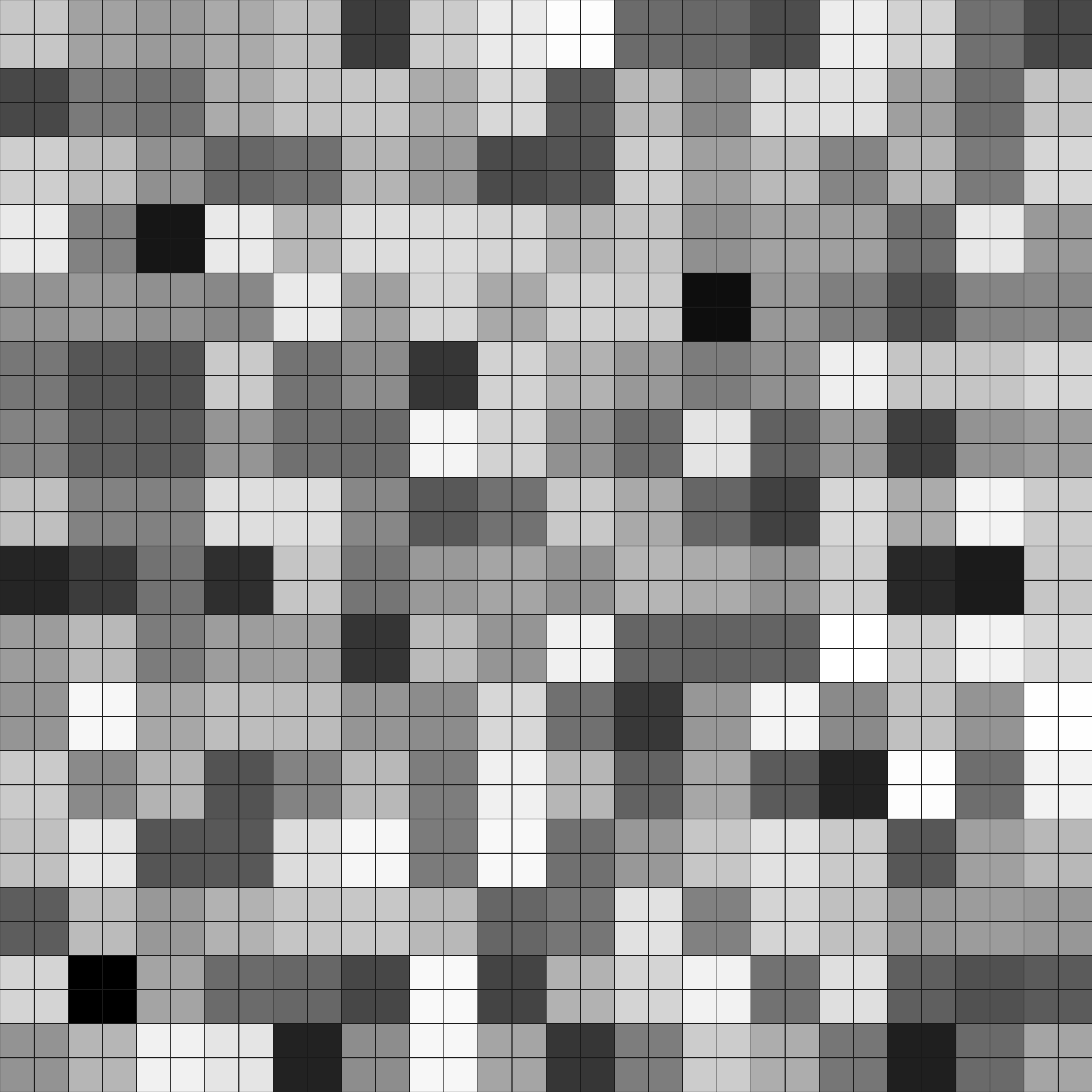}
    \hfill
    \includegraphics[width=0.12\linewidth]{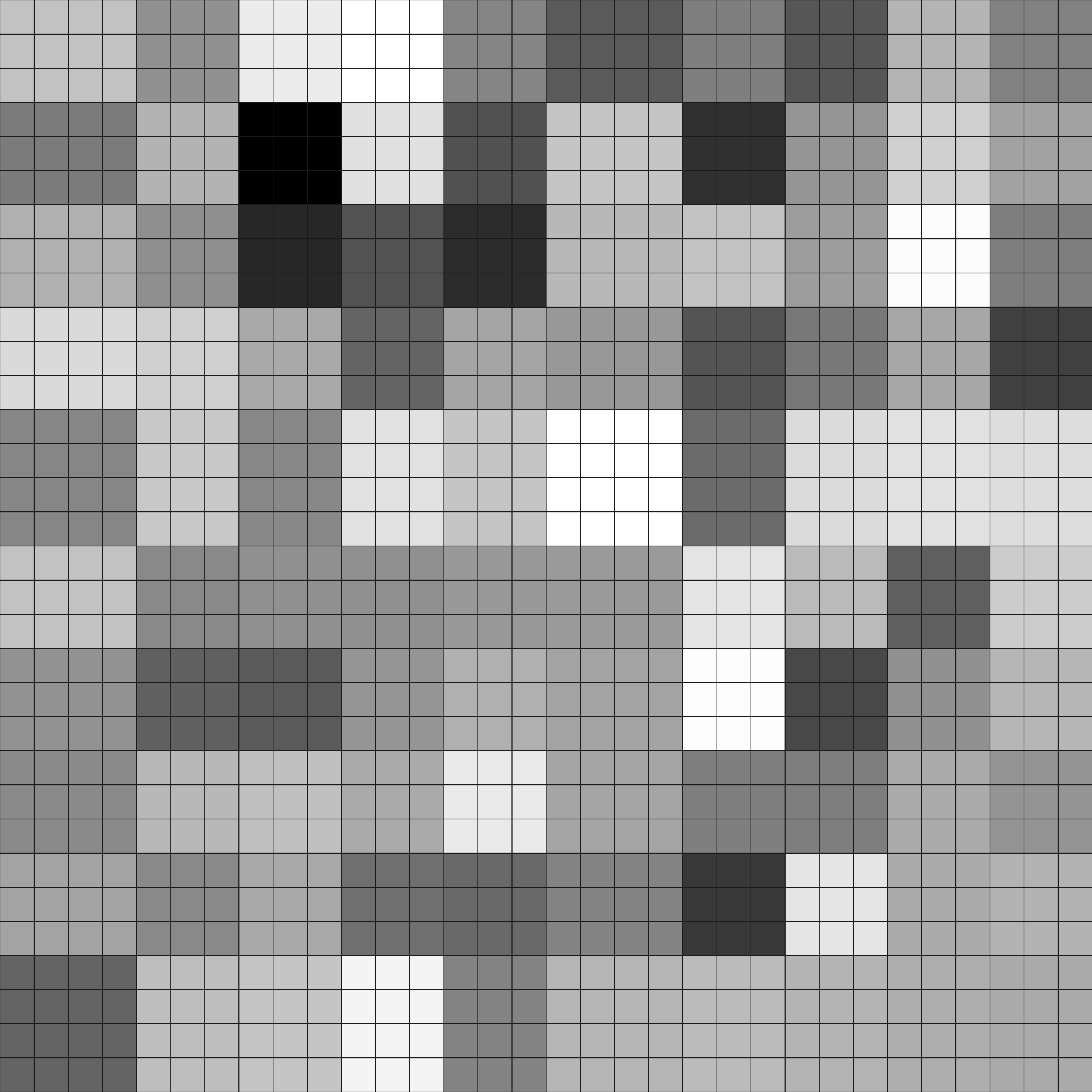}
    \hfill
    \includegraphics[width=0.12\linewidth]{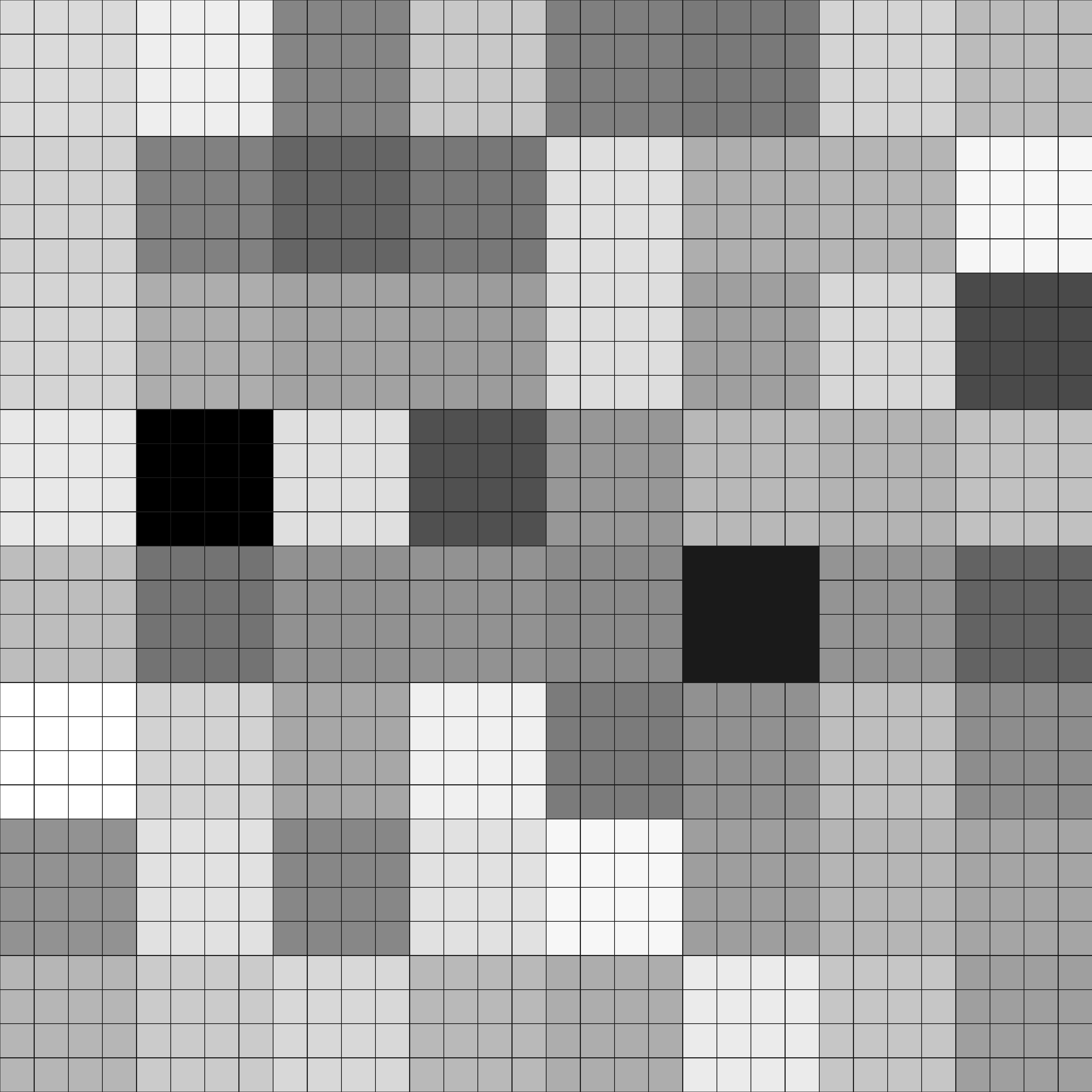}
    \hfill
    \includegraphics[width=0.12\linewidth]{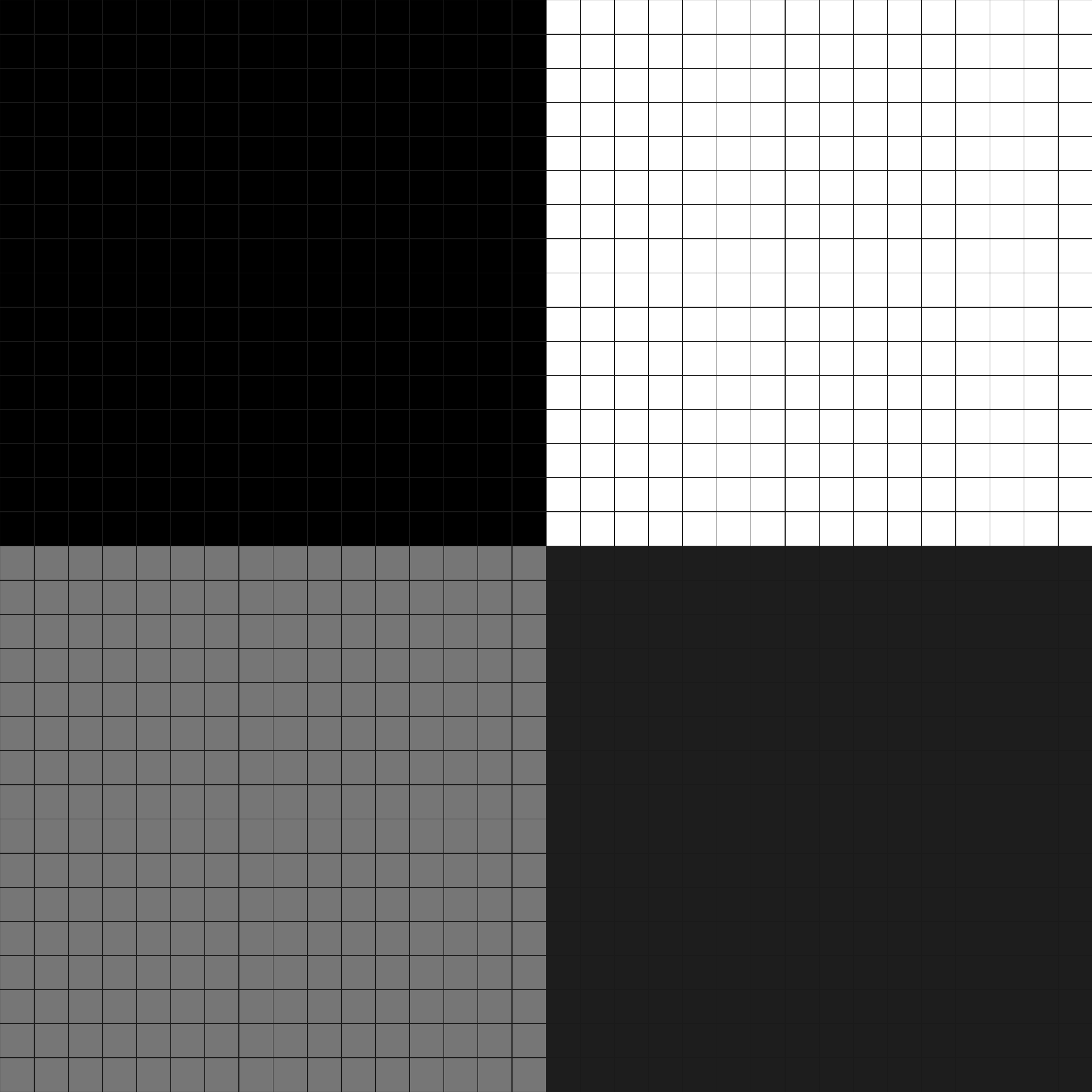}
    \hspace*{\fill}%
    \caption{Example images generated with pooling sizes 2, 3, 4, and 16. Note that the images become smoother with higher pooling sizes.}
    \label{fig:example_input_images_smooth}
\end{figure}

\begin{figure}
    \hspace*{5mm}%
    \begin{subfigure}[t]{0.4\linewidth}
        \includegraphics{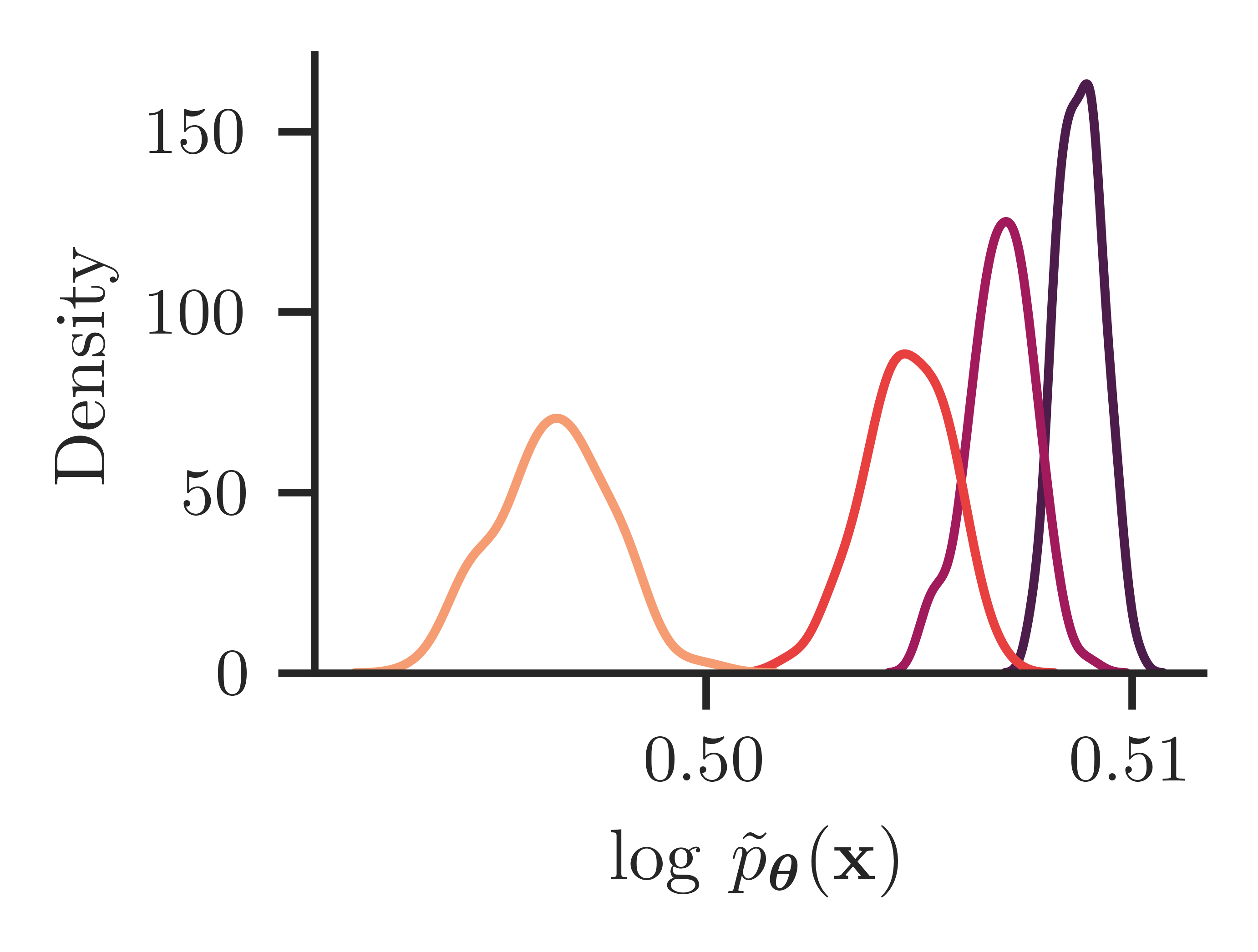}
        \caption{Unsupervised \gls{ebm}.}
        \label{fig:density_histogram_smoothness_unsupervised}
    \end{subfigure}%
    \hfill
    \begin{subfigure}[t]{0.43\linewidth}
        \includegraphics{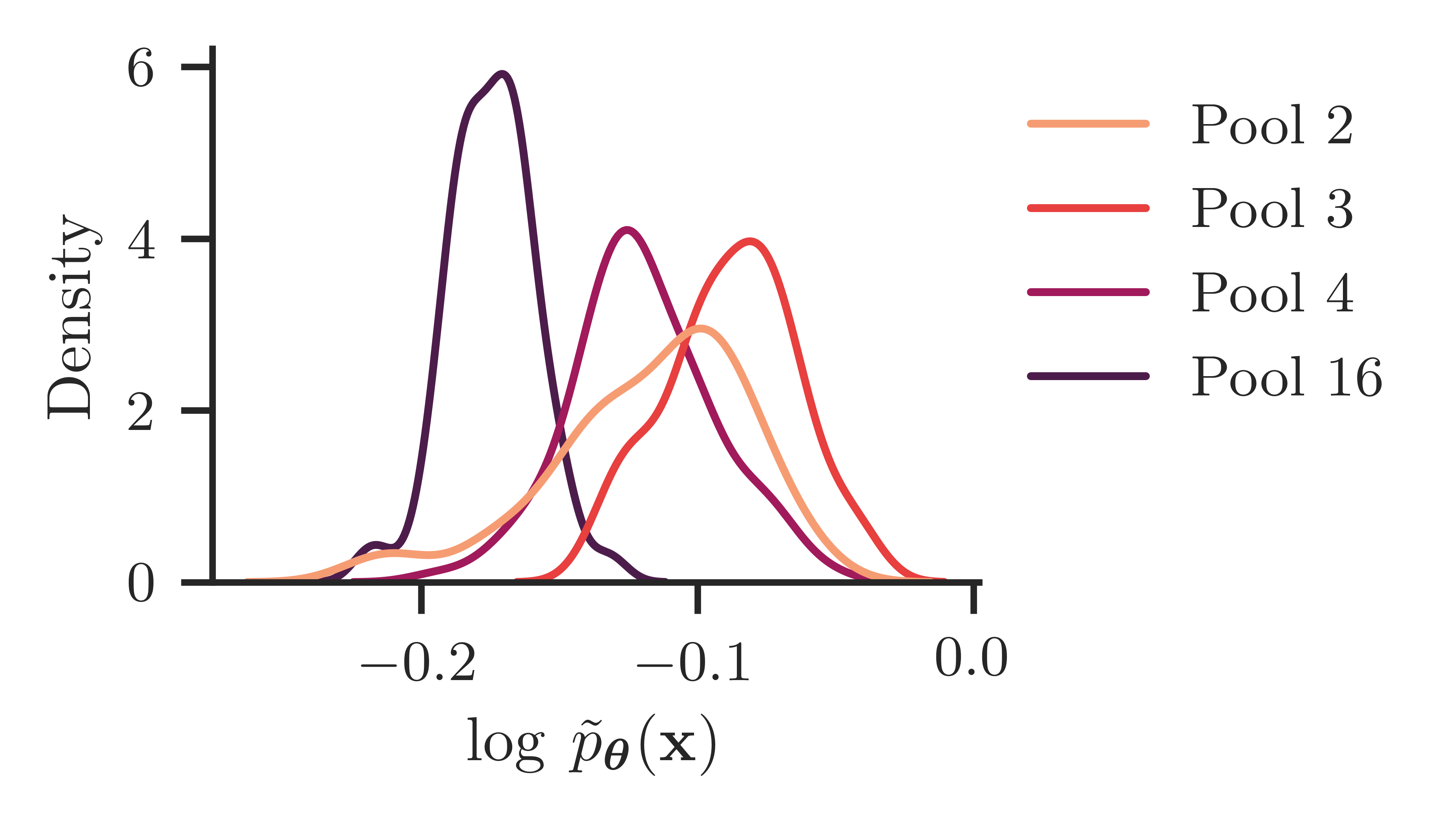}
        \caption{Supervised \gls{ebm}.}
        \label{fig:density_histogram_smoothness_supervised}
    \end{subfigure}%
    \hspace*{\fill}%
    \caption{Density histograms of generated dataset of noise images with different smoothness under different \glspl{ebm}. Higher pooling corresponds to higher smoothness of the input images as visualized in \Cref{fig:example_input_images_smooth}.}
\end{figure}

\begin{figure}
    \hspace*{5mm}%
    \begin{subfigure}[t]{0.4\linewidth}
        \includegraphics{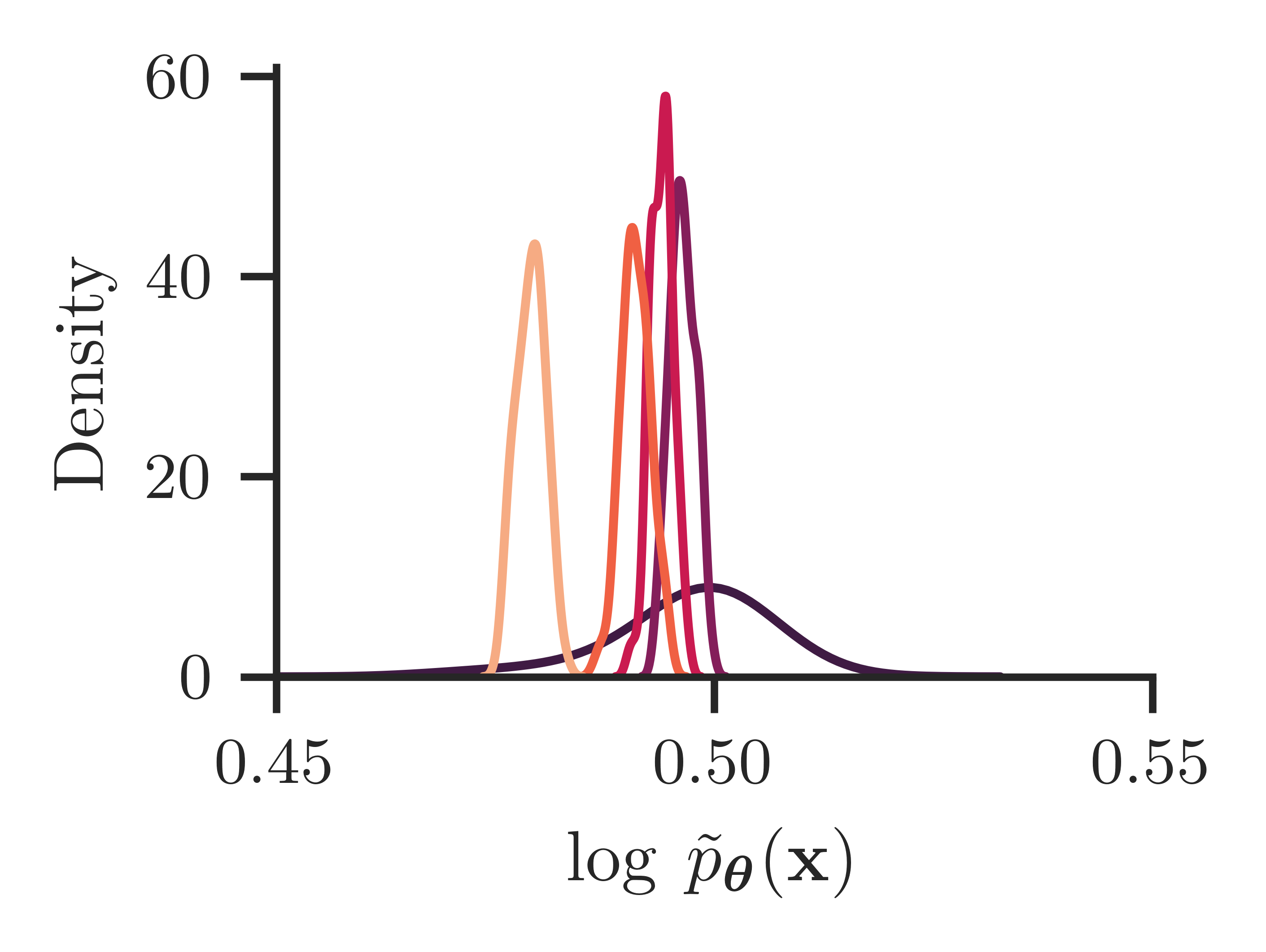}
        \caption{\gls{ebm} without bottleneck.}
        \label{fig:density_histogram_no_bottleneck}
    \end{subfigure}%
    \hfill
    \begin{subfigure}[t]{0.45\linewidth}
        \includegraphics{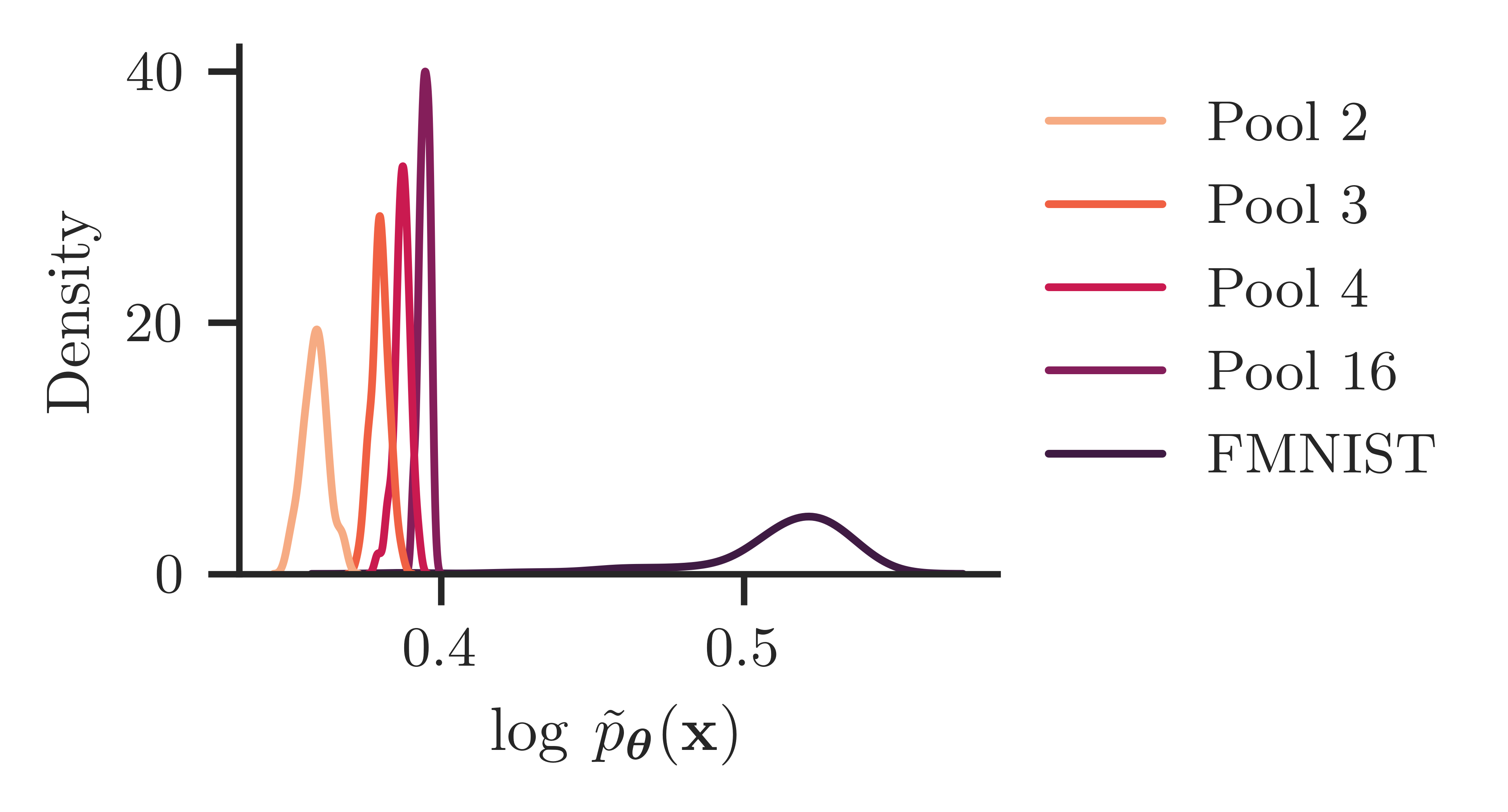}
        \caption{\gls{ebm} with bottleneck.}
        \label{fig:density_histogram_with_bottleneck}
    \end{subfigure}%
    \hspace*{\fill}%
    \caption{Comparison of density histogram of ID test set of FMNIST vs. low-level features for unsupervised \glspl{ebm} with and without bottleneck.}
\end{figure}

\paragraph{Experiment 6.}
To provide further evidence for our observation that low-level features affect the likelihood of unsupervised EBMs, we include density histograms for datasets with varying low-level features. We take inspiration from \citet{serraInputComplexityOutofdistribution2020} and generate images with varying smoothness properties, which \citet{serraInputComplexityOutofdistribution2020} show to affect the likelihood of samples in other generative models. To obtain images with different smoothness, we sample uniform noise at each pixel independently, apply average pooling with different pooling sizes, and resize to the original image dimensions with nearest neighbor upsampling. Images after this processing procedure are shown in \Cref{fig:example_input_images_smooth}. Subsequently, we estimate the density of \(1000\) images generated at each smoothing fidelity under our models. Note that we use average pooling instead of max-pooling in \citet{serraInputComplexityOutofdistribution2020} since max-pooling leads to images with different statistics (higher mean) with increasing pooling sizes. Average pooling allows us to isolate the contribution of the change of smoothness independently of image statistics. 

In \Cref{fig:density_histogram_smoothness_unsupervised}, we observe that unsupervised \glspl{ebm} assign higher likelihood to smoother versions of the dataset (corresponding to higher pooling sizes), while the supervised \gls{ebm} is not affected by the change of low-level features in \Cref{fig:density_histogram_smoothness_supervised}. 
This demonstrates that unsupervised \glspl{ebm} are susceptible to low-level features affecting the likelihood of samples, while supervised \glspl{ebm} rely on higher-level, semantic features to assign likelihoods.

In \Cref{fig:density_histogram_no_bottleneck} and \Cref{fig:density_histogram_with_bottleneck}, we investigate the effect of applying the bottleneck to the architecture of the unsupervised \glspl{ebm}. We observe that the \gls{ebm} with bottleneck assigns higher relative likelihood to the FMNST test set vs. the artificial noise datasets containing low-level features only. This supports our observation that including bottlenecks within the \gls{ebm} helps the model to learn semantic features rather than local, low-level image correlations.

\begin{figure*}
    \centering
    \begin{subfigure}[t]{0.48\linewidth}
        \centering
        \includegraphics[width=\linewidth]{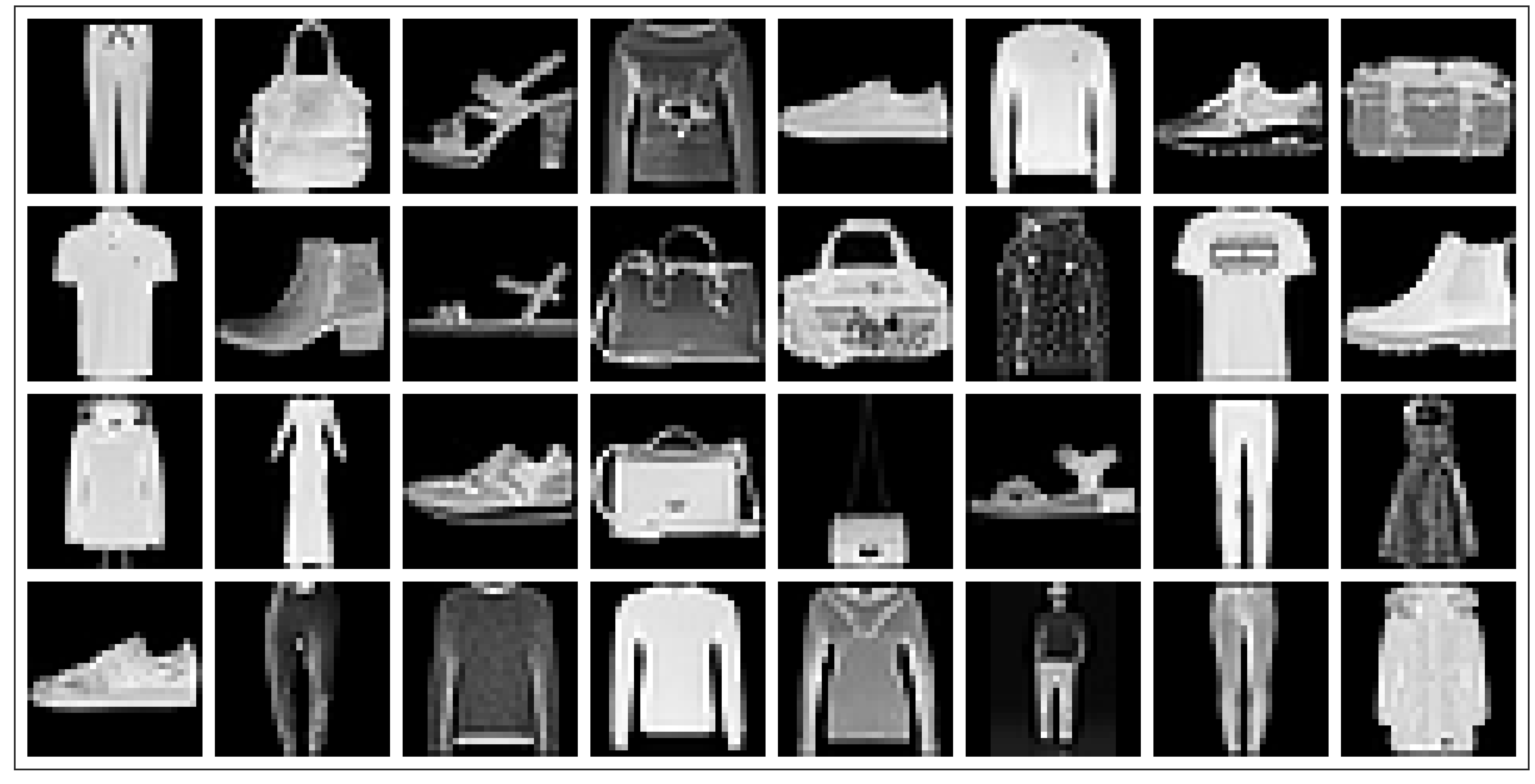}
        \caption{Dataset samples.}
        \label{fig:fmnist_samples}
    \end{subfigure}%
    \hfill
    \begin{subfigure}[t]{0.48\linewidth}
        \centering
        \includegraphics[width=\linewidth]{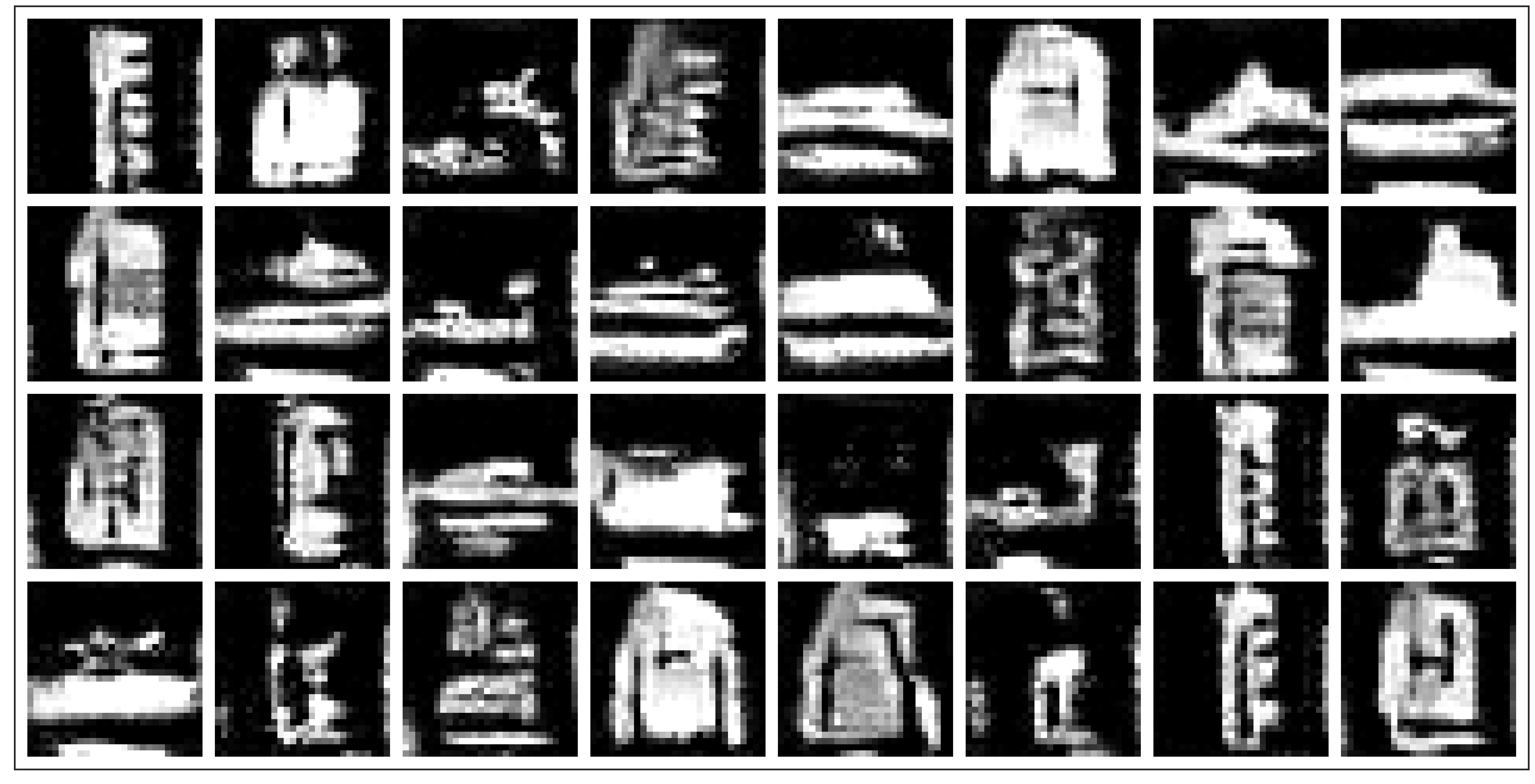}
        \caption{Optimized samples under \textbf{unsupervised} \glspl{ebm}.}
        \label{fig:fmnist_samples_unsupervised}
    \end{subfigure}%
    \vspace{3mm}
    \\
    \begin{subfigure}[t]{0.48\linewidth}
        \centering
        \includegraphics[width=\linewidth]{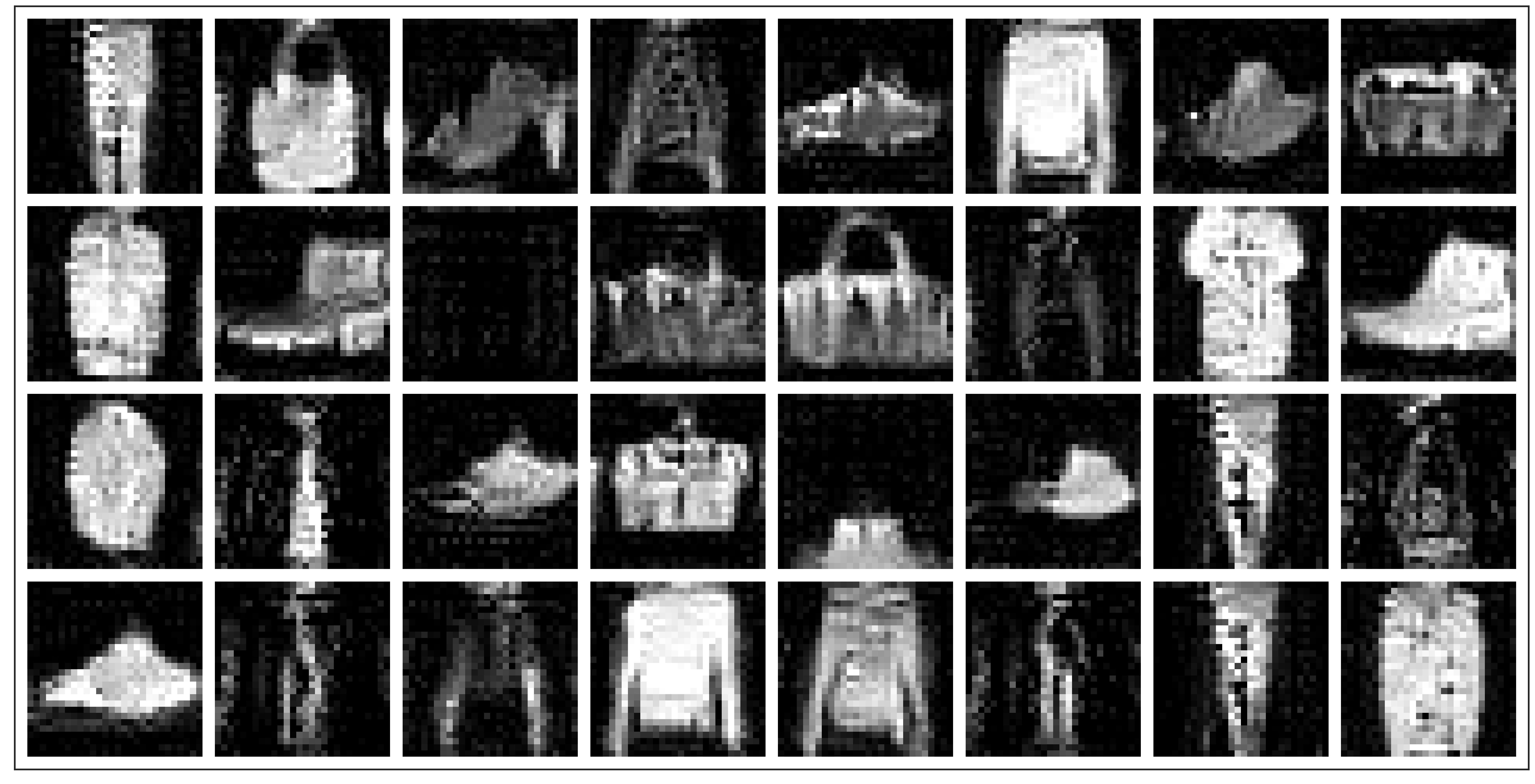}
        \caption{Optimized samples under \textbf{supervised} \glspl{ebm}.}
        \label{fig:fmnist_samples_supervised}
    \end{subfigure}
    \caption{Comparison of samples from the FMNIST dataset.}
\end{figure*}

\paragraph{Experiment 7.}
Finally, we investigate images under the learned \glspl{ebm}. Samples from the FMNST dataset can be found in \Cref{fig:fmnist_samples}. We optimize the likelihood of these samples under the model and visualize samples for unsupervised \gls{ebm} in \Cref{fig:fmnist_samples_unsupervised} and for supervised \gls{ebm} in \Cref{fig:fmnist_samples_supervised}. 

We observe that while the semantic content of samples under the unsupervised model becomes almost indistinguishable, the samples under the supervised model preserve their class semantics. 

This result once more highlights that low-level features are the driving factor for high likelihood in unsupervised \glspl{ebm}, while supervised \glspl{ebm} learn a notion of semantics.

\subsection{Conclusion}
Overall, we find that \textbf{(1)} \glspl{ebm} struggle with \gls{ood} detection on high-dimensional data but to a lower degree than Normalizing Flows, \textbf{(2)} incorporating task-specific priors such as supervision significantly improves \gls{ood} detection on \textit{natural} \gls{ood} data in line with what has been recently suggested by \citet{lanPerfectDensityModels2021, schirrmeisterUnderstandingAnomalyDetection2020} for other generative models, and \textbf{(3)} architectural modifications can be used to improve the \gls{ood} detection performance.  
\chapter{Conclusion}
\label{ch:conclusion}
We conclude this thesis by summarizing our contributions in \Cref{ch:conclusion:sec:summary} and providing ideas for future work in \Cref{ch:conclusion:sec:future_work}.

\section{Summary}
\label{ch:conclusion:sec:summary}
In this work, we derived a connection between \acrlong{ebm} and \acrfull{dpn}. We then leverage this connection to define \acrfull{epn} and an optimization scheme for training based on existing \gls{ebm} training techniques. In comparison to Prior Networks, \gls{epn} does not require an \gls{ood} dataset or additional hyperparameters defining the target Dirichlet distribution during training. We then show the efficacy of \gls{epn} in detecting \gls{ood} samples, distribution shifts, and adversarial attacks.

In a second part of experiments, we investigate the overall properties enabling better \gls{ood} detection of \glspl{ebm} versus other generative models. We find that unsupervised \glspl{ebm} suffer similar problems as Normalizing Flows, however, one can use supervision and architectural modifications to improve the \gls{ood} detection performance.


\section{Future Work}
\label{ch:conclusion:sec:future_work}

There are multiple avenues for improving or extending the ideas presented in this work. Concerning, \gls{epn} one could expand on the underconfidence issue presented in \Cref{ch:results:sec:eval_energy_priornet:subsec:classifier_eval} and why this happens only in some cases. Based on initial observations regularizing the normalizing constant \(Z(\theta)\) might provide a way forward.

Additionally, it would be interesting to investigate the issue of the energy increasing far from the training data when training with the factorization \(p_{\params{}}(\mathbf{x}, y) = p_{\params}(y \mid \mathbf{x}) \times p_{\params}(\mathbf{x})\) since both, \glspl{jem}~\cite{grathwohlYourClassifierSecretly2020} and \gls{epn}, use this type of factorization. In particular, other factorizations such as the joint \(p_{\params}(\mathbf{x}, y)\) could be a compelling direction since it might not lead to degenerate solutions. Initial results optimizing the joint distribution directly can be found in \citet{kelly2021no}. In a similar vein, one could explore the observation in \Cref{ch:results:sec:ood_detection_ebms:subsec:supervision_improves_ood} that enforcing the asymptotic constraint on the energy using \Cref{theorem1} improves the \gls{ood} detection with \glspl{ebm} not only on \textit{non-natural} but also \textit{natural} \gls{ood} data.

Since \gls{epn} only relies on a point estimate of \(p(\params)\), estimating epistemic uncertainty is only possible on a limited basis. Thus, one could extend \gls{epn} with a full Bayesian treatment of \(p(\params)\) such as deep ensembles~\cite{lakshminarayananSimpleScalablePredictive2017} or \gls{bnn}~\cite{maddoxSimpleBaselineBayesian2019}.

Finally, regarding our findings in \Cref{ch:results:sec:ood_detection_ebms:subsec:can_encourage_semantic_features}, investigating which and why some architectural modifications yield improved \gls{ood} detection in \glspl{ebm} might yield further insights for learning high-level features in an unsupervised setting with \glspl{ebm}.

\appendix{}

\chapter{General Addenda}
In this chapter, we present the proof of \Cref{theorem1} in \Cref{ch:appendix:sec:proof}, additional details on the experiments in \Cref{ch:appendix:sec:training_details}, complementary results in \Cref{ch:appendix:sec:additonal_results}, and other research directions with initial results in \Cref{ch:appendix:sec:other_directions}. 

\section{Proof of Theorem 1}
\label{ch:appendix:sec:proof}

\setcounter{theorem}{0}
\begin{theorem}
Let \(\data \in \mathbb{R}^D, \data \neq \mathbf{0}\) and let \(g:\mathbb{R}^D \mapsto \mathbb{R}^K\) be a ReLU-network with parameters \(\boldsymbol{\phi}\) where the last layer explicitly contains a ReLU activation. By \Cref{lemma1}, there exists a finite set of polytopes \(\{Q_r\}_r\) on which \(g\) is affine. Denote by \(Q_t\) the polytope such that \(\beta \data \in Q_t\) for all \(\beta \geq \alpha\) with \(\alpha \in \mathbb{R}\) and by \(g(\mathbf{z}) = U\mathbf{z}+\mathbf{d}\) with \(U \in \mathbb{R}^{K\times D}\) and \(\mathbf{d} \in \mathbb{R}^K\) be the output of \(g\) for \(\mathbf{z} \in Q_t\). Finally, let \(f:\mathbb{R}^D \mapsto \mathbb{R}^C, f(\data) = W g(\data) + \mathbf{b}\) 
be the Energy-based model where the last weight matrix \(W \in \mathbb{R}^{C\times K}\) is component-wise negative and \(\mathbf{b} \in \mathbb{R}^C\) .
If \(U\data \neq \mathbf{0}\), then

\begin{equation}
    \lim_{\beta \mapsto \infty} \efunc(\beta \data) = \lim_{\beta \mapsto \infty} -\log \sum_{c=1}^C e^{(f(\beta \data))_c} = \infty \nonumber
\end{equation}

where \(\params = \{\boldsymbol{\phi}, W, \mathbf{b}\}\) are the learnable parameters.

Furthermore, \(\lim_{\beta \mapsto \infty} p_{\params}(\beta \data) = 0\).
\end{theorem}

The proof follows a very similar structure as the result in \citet{meinkeProvablyRobustDetection2021}.

\begin{proof}
By \Cref{lemma1}, it holds that \(\beta \data \in Q_t\) for \(\beta \geq \alpha\). Thus, we can write the marginal energy as

\begin{align}
    E_{\params}(\beta \data) 
    \label{ch:appendix:sec:proof:eq:marginal_energy}
    &= -\log \sum_{c=1}^C e^{(f(\beta \data))_c} \\
    &= -\log \sum_{c=1}^C e^{\langle (W)_c, U \beta \data + \mathbf{d} \rangle + (\mathbf{b})_c}
\end{align}

where \((W)_c\) denotes the \(c\)-th row of \(W\).

We have \((g(\data))_i \geq 0\) for all \(\data \in \mathbb{R}^D\) since the last layer of \(g\) contains a ReLU activation. Thus, it has to hold \((U \beta \data + \mathbf{d})_i \geq 0\) for all \(\beta \geq \alpha\) and \(i \in {1, \dots, K}\). Therefore, \((U\data)_i \geq 0\) and as \(U\data \neq 0\) there needs to exist a \(i^*\) with \((Ux)_{i^*} > 0\).
Thus, for all \(\beta \geq \alpha\) and \(c \in \{1, \dots, C\}\) it holds

\begin{align}
    (f(\beta \data))_c
    &= \langle (W)_c, U \beta \data + \mathbf{d} \rangle + (\mathbf{b})_c \\
    &= \beta \left\langle (W)_c, U\data \right\rangle + \left\langle (W)_c, \mathbf{d} \right\rangle + (\mathbf{b})_c
\end{align}

As \((W)_{ck} < 0\), i.e., \(W\) has strictly negative components, and \(U\data > 0\), we have \(\langle (W)_c, U\data \rangle < 0\) and thus 

\begin{equation}
    \lim_{\beta \mapsto \infty} (f(\beta \data))_c = -\infty.
\end{equation}

Plugging into \Cref{ch:appendix:sec:proof:eq:marginal_energy} yields the result.

As \(\lim_{\beta \mapsto \infty} \efunc(\beta \data) = 0\), we have

\begin{equation}
    \lim_{\beta \mapsto \infty} p_{\params}(\data) = \lim_{\beta \mapsto \infty} \frac{e^{-\efunc(\beta \data)}}{Z(\params)} = 0
\end{equation}

\end{proof}

\section{Training details}
\label{ch:appendix:sec:training_details}
In this section, we provide further details on the training procedures and hyperparameters for individual methods in \Cref{ch:results}. Unless we specify otherwise, we use the Adam optimizer with default parameters \(\beta_1=0.9\) and \(\beta_2=0.999\). Further, we use learning rate warm-up with \(2500\) steps across all models. We train the models on the tabular datasets for \(10,000\) steps and on the image datasets for \(50\) and \(100\) epochs for the FMNIST and CIFAR-10 datasets, respectively. For the experiments in \Cref{ch:results:sec:ood_detection_ebms}, we perform model selection based on the \gls{aucpr} on an \gls{ood} validation dataset. For CIFAR-10, we use the validation sets of CelebA and CIFAR-100, while for FMNIST we use the validation sets of MNIST and KMNIST. For the tabular dataset, we use \(10\%\) of the hold-out \gls{ood} data for model selection. To obtain the results in \Cref{ch:results:sec:eval_energy_priornet}, we use the accuracy for model selection. For all datasets, we use the original train/validation/test splits if available. In the other cases, we use \(10\%\) of the training set as the validation set.

\paragraph{Contrastive Divergence.}
Following~\cite{grathwohlYourClassifierSecretly2020, duImplicitGenerationGeneralization2020}, we use persistent contrastive divergence~\cite{tielemanTrainingRestrictedBoltzmann2008} which significantly reduces computation time compared to seeding new chains at every iteration as in \citet{nijkampLearningNonConvergentNonPersistent2019}. For the parameters of the \gls{sgld} sampler, we use the settings of \citet{grathwohlYourClassifierSecretly2020}, set the step size \(\alpha\) to \(1\), and reinitialize samples from the replay buffer with probability \(0.05\). The size of the buffer is \(10000\). In contrast to~\cite{grathwohlYourClassifierSecretly2020}, we find that training with \(20\) \gls{sgld} steps consistently diverges, thus, we set the number of \gls{sgld} steps to \(100\) for tabular datasets and \(30\) for image datasets which leads to stable convergence in our experiments. Further, we set the initial learning rate to \(0.001\). Following \citet{duImplicitGenerationGeneralization2020, nijkampLearningNonConvergentNonPersistent2019}, we add additive Gaussian noise with variance \(0.1\) to the inputs to stabilize training.

\paragraph{VERA.}
We use the default hyperparameters proposed in \citet{grathwohlNoMCMCMe2020} and initialize the variance of the variational approximation \(\eta\) with \(0.1\) while clamping \(\eta\) into the range \([0.01, 0.3]\) during training. We perform a grid search for the entropy regularizer in \([10^{-4}, 1]\) and find \(10^{-4}\) to yield the best results in terms for training stability.  
Further, we set the learning rate of the \gls{ebm} to \(3 \cdot 10^{-4}\) and the learning rate of the generator to \(6 \cdot 10^{-4}\). We use the Adam optimizer with parameters \(\beta_1=0.0\) and \(\beta_2=0.9\) to train the generator.

For the generator architecture, we use a 5-layer \gls{mlp} with hidden dimension \(100\) and leaky ReLU activations with slope \(0.2\) for the tabular datasets. The latent distribution is a \(16\)-dim. isotropic Normal distribution. For the image datasets, we follow~\cite{grathwohlNoMCMCMe2020} and use the generator from~\cite{miyatoSpectralNormalizationGenerative2018} based on ResNet~\cite{heDeepResidualLearning2015} blocks with latent dimension \(128\).

\paragraph{Sliced Score Matching.}
We set the distribution $p_v$ to a multivariate Rademacher distribution which enable the use of the variance reduced objective (SSM-VR)~\cite{songSlicedScoreMatching2019}. That is, since one can compute the expectation \(\mathbb{E}_{p_\mathbf{v}} \left[ \mathbf{v}^T \boldsymbol{\psi}_{\params}(\data))^2 \right] = \Vert \boldsymbol{\psi}_{\params}(\data) \Vert_2^2\) analytically~\cite{songSlicedScoreMatching2019}. During training, we use a single projection vector \(\mathbf{v}\) from $p_{\mathbf{v}}$ to compute the objective. 

\paragraph{Normalizing Flow.}
We train Normalizing Flow models with maximum likelihood and learning rate \(10^{-3}\). We perform early stopping based on the log-likelihood of the validation set with patience \(10\).

\paragraph{Cross-entropy classifier.}
We train the cross-entropy baseline with learning rate \(10^{-3}\) on the tabular and \(10^{-4}\) on the image datasets. Further, we use weight decay with weight \(5 \times 10^{-4}\) and perform early stopping based on the accuracy of the model. 

\paragraph{EnergyOOD.}
For the \textit{EnergyOOD} model, we use \(m_{\text{in}} = -23\) and \(m_{\text{out}} = -5\) as suggested in \citet{liuEnergybasedOutofdistributionDetection2020} and set the weighting for the margin loss \(L_{\text{Energy}}\) to \(0.1\). We first train models based on a cross-entropy objective. Then, we finetune the model for \(10\) epochs on the image datasets with learning rate \(10^{-4}\) and \(10000\) iterations with learning rate \(10^{-3}\) on the tabular datasets. We perform model selection based on the the overall combined loss of cross-entropy and \(L_{\text{Energy}}\).

\paragraph{Outlier-Exposure.}
For the \textit{OE} model, we use a weighting of \(0.5\) for the \(L_{\text{OE}}\) loss term. Similar to \textit{EnergyOOD}, we finetune models trained with cross-entropy loss for \(10\) epochs on the image datasets and \(10000\) iterations on the tabular datasets with the same learning rates.

\paragraph{\acrlong{dpn}.}
For \gls{dpn}, we use learning rate \(10^{-4}\) on the image and \(10^{-3}\) on the tabular datasets. Further, we set the concentration of the target class \(\beta_y = 100\) and use the reverse \gls{kl} divergence training~\cite{malininReverseKLDivergenceTraining2019}. Finally, we use a weighting of \(1\) to balance the \gls{id} and \gls{ood} loss terms.

\paragraph{Posterior Network.}
We use learning rate \(10^{-4}\) and use the class counts of the training dataset for the budget function. Further, we set the weight for the entropy regularizer to \(10^{-5}\). We set the dimensionality of the latent space to \(10\) and use \(8\) stacked radial transforms~\cite{rezendeVariationalInferenceNormalizing2016} for the Normalizing Flows. We perform model selection based on the accuracy.

\paragraph{\acrlong{epn}.}
For \gls{epn}, we use learning rate \(3 \cdot 10^{-5}\) on the image datasets and \(10^{-3}\) on the tabular datasets. We set the weighting of the \gls{kl} term to \(1.0\) for the \textit{EPN-M} and \(100.0\) for the \textit{EPN-V} model after performing grid search in \(\{0.1, 1.0, 10, 100\}\). Further, we weight the entropy regularizer with \(10^{-4}\). For the \gls{ebm} optimization, we use the default parameters of \gls{vera} for \textit{EPN-V} and of contrastive divergence for \textit{EPN-M}, as described before. We perform model selection based on the accuracy on the validation set.

\subsection{Implementation \& Configuration}
We implement all models with PyTorch~\cite{paszkePyTorchImperativeStyle2019} and train on a single GPU with either \textit{Nvidia GeForce RTX 2080Ti} or \textit{Nvidia GeForce GTX 1080Ti}. We provide the full implementation of all models together with training configurations for reproducing the results in this work at \url{https://github.com/selflein/MA-EBM}.

\section{Additional results}
\label{ch:appendix:sec:additonal_results}
In this section, we add complementary results for the \gls{ood} detection with \gls{epn} in \Cref{ch:results:sec:eval_energy_priornet:subsec:ood_detection}. \gls{ood} detection results for the Sensorless dataset, we add in \Cref{ch:results:sec:eval_energy_priornet:subsec:ood_detection:tab:sensorless_ood} and for the CIFAR-10 dataset in \Cref{ch:results:sec:eval_energy_priornet:subsec:ood_detection:tab:cifar10}. \gls{epn} remains competitive, however, we find that the \gls{epn}-V variant starts to outperform the \gls{epn}-M variant on the CIFAR-10 dataset. We explain this observation with better capabilities of \gls{vera} for estimating the marginal data distribution compared to \gls{cd} training with \gls{mcmc}. These results are consistent with results in \citet{grathwohlNoMCMCMe2020} for \gls{jem} models.

For completeness in \Cref{ch:results:sec:ood_detection_ebms}, we report the tables measuring \gls{ood} detection on \textit{natural} datasets in \Cref{tab:full_overall_table} and on \textit{non-natural} datasets in \Cref{tab:full_overall_table_other}. Models with \textit{-E} suffix correspond to models trained on classifier embeddings, while models with the \textit{-S} suffix correspond to models trained with additional supervision in the form of cross-entropy objective weighted with parameter $\gamma=1$.  We also present the full results of \Cref{ch:results:sec:ood_detection_ebms:subsec:can_encourage_semantic_features} for different choices of the bottleneck size in \Cref{tab:bottleneck_full}.

\begin{figure*}
    \centering
    \includegraphics{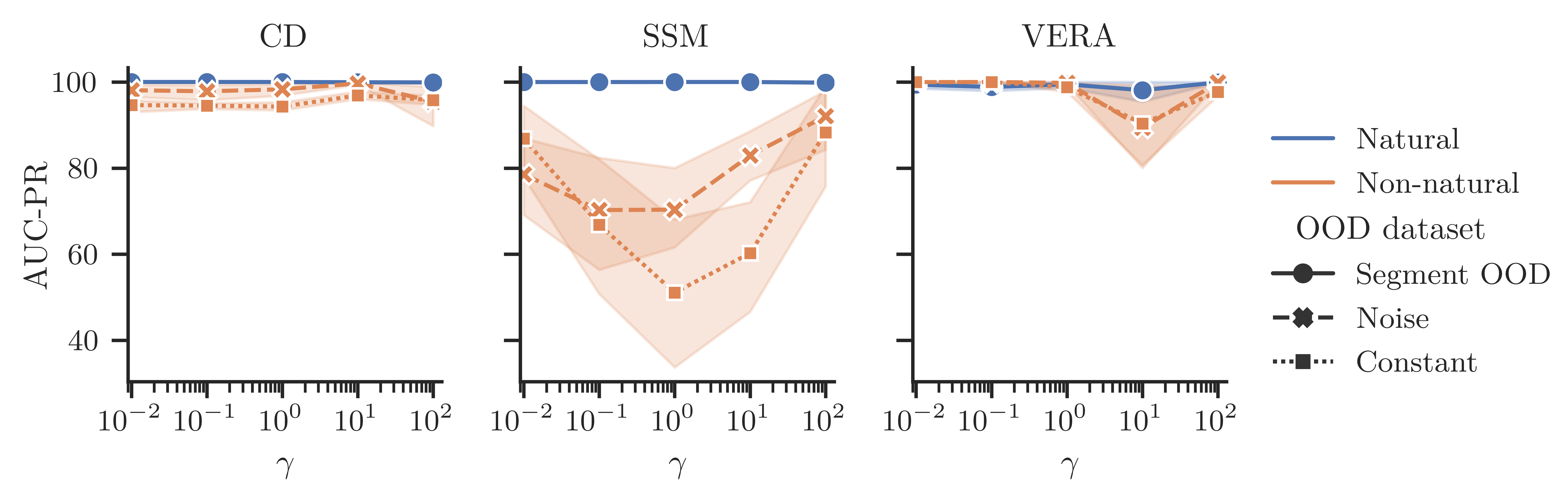}
    \caption{AUC-PR for OOD detection for different settings of the weighting hyperparameter \(\gamma\) of the cross entropy objective. Segment is used as the in-distribution dataset.}
    \label{fig:segment_clf_weight}
\end{figure*}

\begin{figure*}
    \centering
    \includegraphics{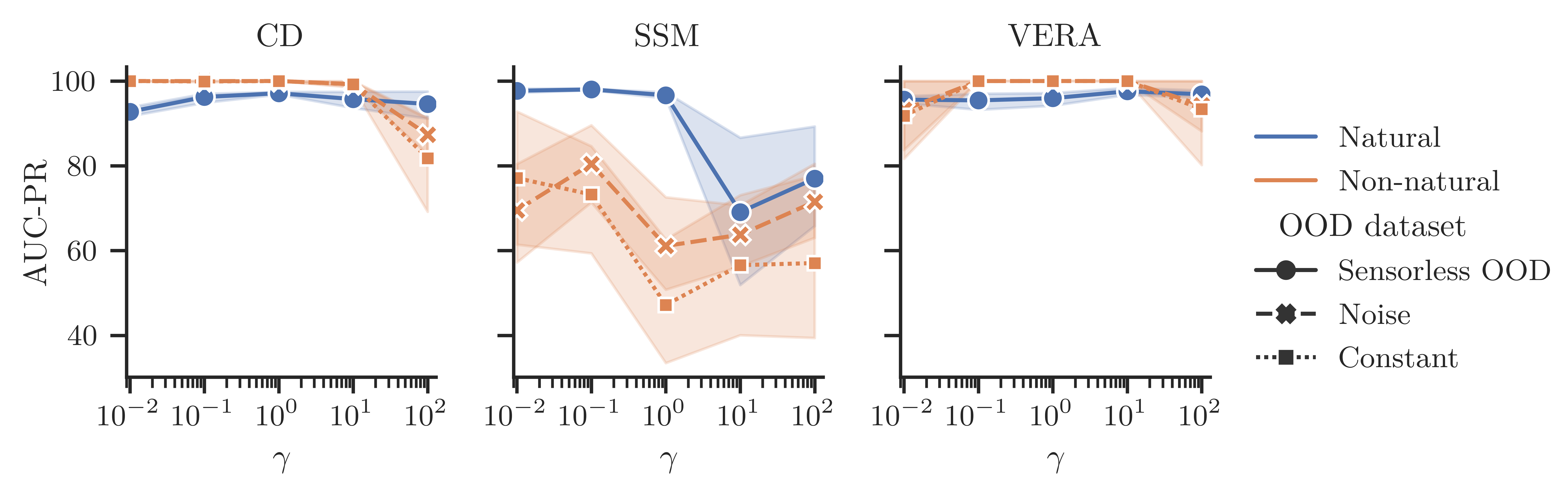}
    \caption{AUC-PR for OOD detection for different settings of the weighting hyperparameter \(\gamma\) of the cross entropy objective. Sensorless is used as the in-distribution dataset.}
    \label{fig:sensorless_clf_weight}
\end{figure*}

\begin{figure*}
    \centering
    \includegraphics{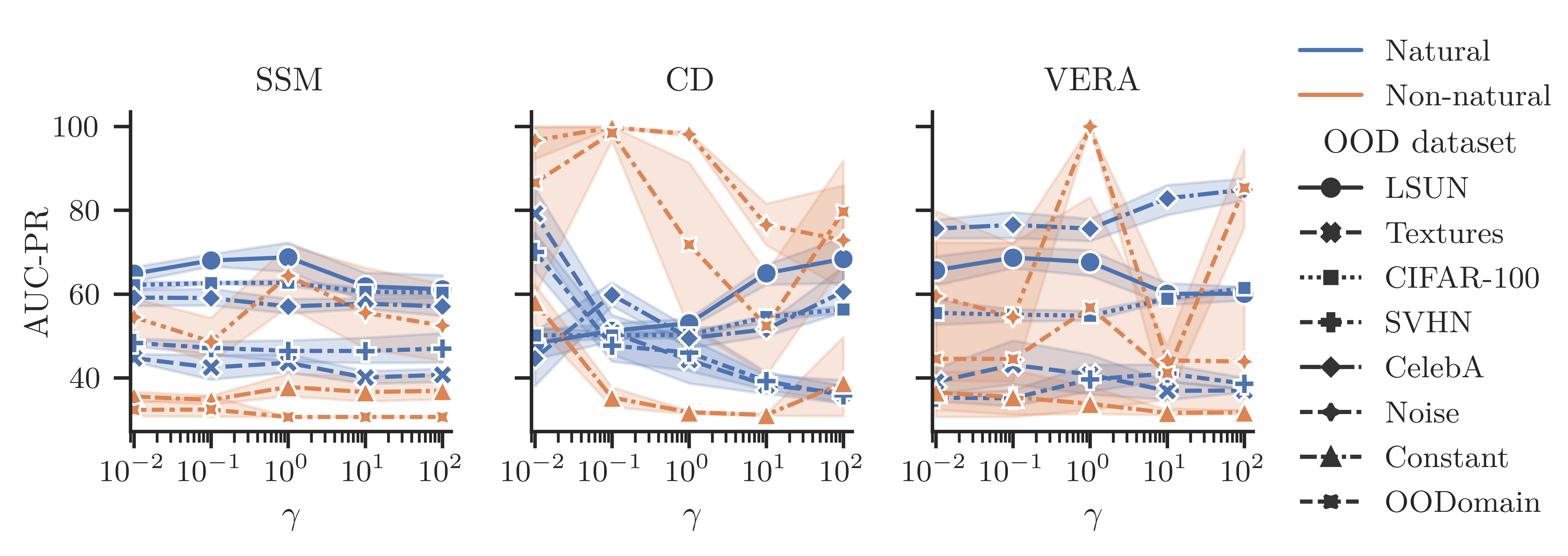}
    \caption{AUC-PR for OOD detection for different settings of the weighting hyperparameter \(\gamma\) of the cross entropy objective. CIFAR-10 is used as the in-distribution dataset.}
    \label{fig:cifar10_clf_weight}
\end{figure*}

In addition to the results for the effect of \(\gamma\) on \gls{ood} detection on FMNIST in \Cref{ch:results:sec:ood_detection_ebms:subsec:supervision_improves_ood}, we add results for the Segment dataset in \Cref{fig:segment_clf_weight}, the Sensorless dataset in \Cref{fig:sensorless_clf_weight} and CIFAR-10 in \Cref{fig:cifar10_clf_weight}. Our main finding holds that the choice of \(\gamma\) heavily affects the \gls{ood} detection performance, particularly on high-dimensional datasets.

Finally, we report the full \gls{ood} detection results in \Cref{ch:results:sec:eval_energy_priornet:subsec:ood_detection:tab:ssm_comparison} for the experiment where we enforce the asymptotic convergence of the energy using \Cref{theorem1} in \Cref{ch:results:sec:ood_detection_ebms:subsec:supervision_improves_ood}. We add the baselines of \gls{ssm} training (\textit{SSM}) and \gls{ssm} training with additional supervision (\textit{SSM-S}) for reference. As we mentioned before, the asymptotic guarantee leads to surprising improvements on the \textit{natural} datasets.

\begin{table}
    \caption{AUC-PR for \gls{ood} detection on the Sensorless dataset.}
    \label{ch:results:sec:eval_energy_priornet:subsec:ood_detection:tab:sensorless_ood}
    \centering
    \begin{tabular}{lllll}
\toprule
           & OOD dataset &                                    Constant &                                       Noise &                             Sensorless OOD \\
Model & Score &                                             &                                             &                                            \\
\midrule
CE Baseline & $\max p(y \mid \mathbf{x})$ &            35.93 {\footnotesize $\pm$ 5.09} &            33.52 {\footnotesize $\pm$ 1.38} &           40.02 {\footnotesize $\pm$ 5.41} \\
EnergyOOD & $p_{\boldsymbol{\theta}}(\mathbf{x})$ &  \bfseries{100.0 {\footnotesize $\pm$ 0.0}} &  \bfseries{100.0 {\footnotesize $\pm$ 0.0}} &  \bfseries{97.7 {\footnotesize $\pm$ 1.2}} \\
Ensemble & Variance &                                      99.32  &            94.82 {\footnotesize $\pm$ 0.59} &                                     77.72  \\
JEM & $p_{\boldsymbol{\theta}}(\mathbf{x})$ &  \bfseries{100.0 {\footnotesize $\pm$ 0.0}} &  \bfseries{100.0 {\footnotesize $\pm$ 0.0}} &          77.89 {\footnotesize $\pm$ 12.39} \\
MC Dropout & Variance &             64.27 {\footnotesize $\pm$ 9.1} &            52.45 {\footnotesize $\pm$ 0.76} &           34.57 {\footnotesize $\pm$ 0.17} \\
OE & $\max p(y \mid \mathbf{x})$ &            99.96 {\footnotesize $\pm$ 0.01} &  \bfseries{100.0 {\footnotesize $\pm$ 0.0}} &          60.93 {\footnotesize $\pm$ 17.18} \\
PostNet & $\max p(y \mid \mathbf{x})$ &            99.99 {\footnotesize $\pm$ 0.01} &            99.99 {\footnotesize $\pm$ 0.01} &           80.05 {\footnotesize $\pm$ 1.45} \\
R-PriorNet & $\mathcal{H}(p(\boldsymbol{\mu} \mid \mathbf{x}, \boldsymbol{\theta}))$ &             99.99 {\footnotesize $\pm$ 0.0} &             99.99 {\footnotesize $\pm$ 0.0} &           85.18 {\footnotesize $\pm$ 12.3} \\
\midrule
\multirow{2}{*}{EPN-M} & $\mathcal{H}(p(\boldsymbol{\mu} \mid \mathbf{x}, \boldsymbol{\theta}))$ &             99.99 {\footnotesize $\pm$ 0.0} &             99.99 {\footnotesize $\pm$ 0.0} &           94.88 {\footnotesize $\pm$ 2.82} \\
           & $p_{\boldsymbol{\theta}}(\mathbf{x})$ &             99.99 {\footnotesize $\pm$ 0.0} &             99.99 {\footnotesize $\pm$ 0.0} &           95.57 {\footnotesize $\pm$ 1.87} \\
\midrule
\multirow{2}{*}{EPN-V} & $\mathcal{H}(p(\boldsymbol{\mu} \mid \mathbf{x}, \boldsymbol{\theta}))$ &             99.99 {\footnotesize $\pm$ 0.0} &            98.05 {\footnotesize $\pm$ 3.61} &           60.43 {\footnotesize $\pm$ 0.91} \\
           & $p_{\boldsymbol{\theta}}(\mathbf{x})$ &             99.99 {\footnotesize $\pm$ 0.0} &             98.2 {\footnotesize $\pm$ 3.36} &          82.48 {\footnotesize $\pm$ 15.26} \\
\bottomrule
\end{tabular}
\end{table}

\begin{table}
    \caption{AUC-PR for \gls{ood} detection on the CIFAR-10 dataset.}
    \label{ch:results:sec:eval_energy_priornet:subsec:ood_detection:tab:cifar10}
    \centering
    \resizebox{\linewidth}{!}{%
    \setlength{\tabcolsep}{1mm}
    \begin{tabular}{llllllllll}
\toprule
                 & OOD dataset &                                         SVHN &                                         LSUN &                                    CIFAR-100 &                                      CelebA &                                    Textures &                                       Noise &                                     Constant &                                     OODomain \\
Model & Score &                                              &                                              &                                              &                                             &                                             &                                             &                                              &                                              \\
\midrule
CE Baseline & $\max p(y \mid \mathbf{x})$ &             87.54 {\footnotesize $\pm$ 0.11} &             85.94 {\footnotesize $\pm$ 0.53} &             98.66 {\footnotesize $\pm$ 0.14} &             91.9 {\footnotesize $\pm$ 0.41} &              98.4 {\footnotesize $\pm$ 0.9} &            38.12 {\footnotesize $\pm$ 0.56} &             94.08 {\footnotesize $\pm$ 0.52} &             95.72 {\footnotesize $\pm$ 0.74} \\
EnergyOOD & $p_{\boldsymbol{\theta}}(\mathbf{x})$ &             84.34 {\footnotesize $\pm$ 0.34} &              85.3 {\footnotesize $\pm$ 4.15} &  \bfseries{99.94 {\footnotesize $\pm$ 0.08}} &            91.75 {\footnotesize $\pm$ 1.45} &  \bfseries{100.0 {\footnotesize $\pm$ 0.0}} &             76.8 {\footnotesize $\pm$ 7.21} &  \bfseries{98.01 {\footnotesize $\pm$ 1.71}} &             93.38 {\footnotesize $\pm$ 5.67} \\
Ensemble & Variance &                                       56.33  &                                       67.99  &                                        34.7  &                                      61.46  &           68.35 {\footnotesize $\pm$ 31.47} &                                      99.62  &                                       56.45  &                                        41.3  \\
JEM & $p_{\boldsymbol{\theta}}(\mathbf{x})$ &             55.67 {\footnotesize $\pm$ 1.14} &             79.66 {\footnotesize $\pm$ 2.73} &             31.96 {\footnotesize $\pm$ 0.07} &            65.66 {\footnotesize $\pm$ 1.75} &            31.13 {\footnotesize $\pm$ 0.38} &            31.45 {\footnotesize $\pm$ 0.37} &             60.64 {\footnotesize $\pm$ 2.56} &              42.61 {\footnotesize $\pm$ 5.1} \\
MC Dropout & Variance &             87.43 {\footnotesize $\pm$ 0.02} &  \bfseries{89.98 {\footnotesize $\pm$ 0.93}} &             89.54 {\footnotesize $\pm$ 0.05} &            90.67 {\footnotesize $\pm$ 0.03} &            92.31 {\footnotesize $\pm$ 0.27} &            36.37 {\footnotesize $\pm$ 0.71} &              89.9 {\footnotesize $\pm$ 0.15} &             89.99 {\footnotesize $\pm$ 0.53} \\
OE & $\max p(y \mid \mathbf{x})$ &  \bfseries{87.55 {\footnotesize $\pm$ 0.14}} &              89.4 {\footnotesize $\pm$ 0.34} &             99.17 {\footnotesize $\pm$ 0.99} &  \bfseries{92.6 {\footnotesize $\pm$ 0.39}} &  \bfseries{100.0 {\footnotesize $\pm$ 0.0}} &            36.18 {\footnotesize $\pm$ 1.43} &             93.88 {\footnotesize $\pm$ 0.17} &  \bfseries{96.59 {\footnotesize $\pm$ 0.57}} \\
PostNet & $\max p(y \mid \mathbf{x})$ &             83.34 {\footnotesize $\pm$ 0.88} &   \bfseries{89.98 {\footnotesize $\pm$ 0.5}} &             97.66 {\footnotesize $\pm$ 1.11} &            87.49 {\footnotesize $\pm$ 0.91} &            99.24 {\footnotesize $\pm$ 0.24} &  \bfseries{100.0 {\footnotesize $\pm$ 0.0}} &              92.1 {\footnotesize $\pm$ 0.52} &             91.54 {\footnotesize $\pm$ 2.43} \\
R-PriorNet & $\mathcal{H}(p(\boldsymbol{\mu} \mid \mathbf{x}, \boldsymbol{\theta}))$ &             79.41 {\footnotesize $\pm$ 0.55} &             79.02 {\footnotesize $\pm$ 1.59} &             94.47 {\footnotesize $\pm$ 0.93} &            84.42 {\footnotesize $\pm$ 1.82} &  \bfseries{100.0 {\footnotesize $\pm$ 0.0}} &            85.1 {\footnotesize $\pm$ 11.29} &             86.91 {\footnotesize $\pm$ 0.98} &             77.99 {\footnotesize $\pm$ 2.37} \\
\midrule
\multirow{2}{*}{EPN-M} & $\mathcal{H}(p(\boldsymbol{\mu} \mid \mathbf{x}, \boldsymbol{\theta}))$ &            66.29 {\footnotesize $\pm$ 15.88} &             63.98 {\footnotesize $\pm$ 1.33} &             83.01 {\footnotesize $\pm$ 5.52} &           70.48 {\footnotesize $\pm$ 11.38} &  \bfseries{100.0 {\footnotesize $\pm$ 0.0}} &  \bfseries{100.0 {\footnotesize $\pm$ 0.0}} &            76.85 {\footnotesize $\pm$ 12.55} &             85.16 {\footnotesize $\pm$ 2.44} \\
                 & $p_{\boldsymbol{\theta}}(\mathbf{x})$ &             53.53 {\footnotesize $\pm$ 0.97} &             58.83 {\footnotesize $\pm$ 0.73} &             67.31 {\footnotesize $\pm$ 7.38} &            54.76 {\footnotesize $\pm$ 0.23} &  \bfseries{100.0 {\footnotesize $\pm$ 0.0}} &  \bfseries{100.0 {\footnotesize $\pm$ 0.0}} &             65.38 {\footnotesize $\pm$ 6.66} &             84.04 {\footnotesize $\pm$ 3.24} \\
\midrule
\multirow{2}{*}{EPN-V} & $\mathcal{H}(p(\boldsymbol{\mu} \mid \mathbf{x}, \boldsymbol{\theta}))$ &             80.07 {\footnotesize $\pm$ 0.06} &              87.04 {\footnotesize $\pm$ 0.2} &             84.21 {\footnotesize $\pm$ 5.75} &             83.77 {\footnotesize $\pm$ 0.2} &  \bfseries{100.0 {\footnotesize $\pm$ 0.0}} &            69.88 {\footnotesize $\pm$ 3.62} &              85.8 {\footnotesize $\pm$ 1.15} &             73.25 {\footnotesize $\pm$ 0.71} \\
                 & $p_{\boldsymbol{\theta}}(\mathbf{x})$ &             79.58 {\footnotesize $\pm$ 0.11} &             87.22 {\footnotesize $\pm$ 0.26} &             78.96 {\footnotesize $\pm$ 6.87} &             83.6 {\footnotesize $\pm$ 0.31} &  \bfseries{100.0 {\footnotesize $\pm$ 0.0}} &            30.74 {\footnotesize $\pm$ 0.04} &             83.85 {\footnotesize $\pm$ 1.85} &             69.18 {\footnotesize $\pm$ 0.45} \\
\bottomrule
\end{tabular}
    }
\end{table}

\begin{table}
    \caption{AUC-PR for \gls{ood} detection with EBMs trained using SSM. \textit{SSM} uses plain SSM training, \textit{SSM-S} uses an additional cross-entropy loss, and \textit{SSM-S-A} additionally leverages the parameterization with asymptotic guarantee following \Cref{theorem1}.}
    \label{ch:results:sec:eval_energy_priornet:subsec:ood_detection:tab:ssm_comparison}
    \centering
    \resizebox{\linewidth}{!}{%
    \setlength{\tabcolsep}{1mm}
    \begin{tabular}{lrrrrrrrrrrrrrr}
\toprule
ID & \multicolumn{8}{c}{CIFAR-10} & \multicolumn{6}{c}{FMNIST} \\
\cmidrule(l){2-9} \cmidrule(l){10-15} 
OOD &              SVHN &              LSUN &            CelebA &         CIFAR-100 &         Textures &             Noise &          Constant &          OODomain &            KMNIST &             MNIST &          NotMNIST &             Noise &          Constant &          OODomain \\
\midrule
SSM               &             45.75 &             52.79 &             57.72 &             53.82 &            48.82 &             70.23 &              47.2 &             68.57 &             58.98 &             67.86 &             57.27 &              49.5 &             47.59 &             76.76 \\
SSM-S     &             46.54 &  \bfseries{68.77} &             57.15 &             62.71 &            43.51 &             64.15 &             37.83 &             30.69 &             93.77 &  \bfseries{99.22} &             84.93 &              33.7 &             37.52 &             70.03 \\
SSM-S-A &  \bfseries{70.95} &             68.14 &  \bfseries{81.93} &  \bfseries{66.22} &  \bfseries{76.0} &  \bfseries{84.23} &  \bfseries{84.61} &  \bfseries{100.0} &  \bfseries{96.51} &             89.06 &  \bfseries{89.66} &  \bfseries{99.69} &  \bfseries{96.39} &  \bfseries{100.0} \\
\bottomrule
\end{tabular}
    }
\end{table}

\begin{table*}
    \centering
    \caption{AUC-PR for OOD detection on the natural datasets when trained on the respective in-distribution dataset.}
    \label{tab:full_overall_table}
    \setlength{\tabcolsep}{1mm}
    \resizebox{\linewidth}{!}{%
    \begin{tabular}{lllllllllll}
\toprule
ID & \multicolumn{5}{c}{CIFAR-10} & \multicolumn{3}{c}{FMNIST} &                                     \multicolumn{1}{c}{Segment} &                                   \multicolumn{1}{c}{Sensorless} \\
\cmidrule(l){2-6} \cmidrule(l){7-9} \cmidrule(l){10-10} \cmidrule(l){11-11} 
OOD &                                   CIFAR-100 &                                       CelebA &                                        LSUN &                                         SVHN &                                    Textures &                                       KMNIST &                                        MNIST &                                     NotMNIST &                                 OOD &                               OOD \\
\midrule
CE     &            62.76 {\footnotesize $\pm$ 1.46} &             64.47 {\footnotesize $\pm$ 2.44} &            65.18 {\footnotesize $\pm$ 5.79} &             47.51 {\footnotesize $\pm$ 4.58} &            39.17 {\footnotesize $\pm$ 2.28} &             69.07 {\footnotesize $\pm$ 6.73} &             82.5 {\footnotesize $\pm$ 12.27} &              50.9 {\footnotesize $\pm$ 6.73} &            33.35 {\footnotesize $\pm$ 1.82} &             33.02 {\footnotesize $\pm$ 1.32} \\
NF     &                                      58.34  &                                       74.68  &                                      62.99  &                                       31.58  &                                      50.23  &                                       62.22  &                                       49.03  &                                       93.68  &                                      99.12  &                                       94.35  \\
\midrule
CD   &            50.51 {\footnotesize $\pm$ 2.13} &             43.86 {\footnotesize $\pm$ 5.85} &           54.43 {\footnotesize $\pm$ 11.37} &            60.72 {\footnotesize $\pm$ 24.59} &           76.21 {\footnotesize $\pm$ 17.44} &             50.52 {\footnotesize $\pm$ 9.39} &              31.69 {\footnotesize $\pm$ 0.9} &             76.85 {\footnotesize $\pm$ 2.66} &            98.18 {\footnotesize $\pm$ 2.18} &            72.83 {\footnotesize $\pm$ 16.19} \\
CD-E &            77.88 {\footnotesize $\pm$ 1.61} &             66.21 {\footnotesize $\pm$ 1.94} &            80.61 {\footnotesize $\pm$ 4.26} &  \bfseries{97.38 {\footnotesize $\pm$ 1.15}} &  \bfseries{98.6 {\footnotesize $\pm$ 0.56}} &             90.53 {\footnotesize $\pm$ 3.38} &             93.05 {\footnotesize $\pm$ 2.88} &              88.05 {\footnotesize $\pm$ 4.5} &                                           - &                                            - \\
CD-S &            49.81 {\footnotesize $\pm$ 1.15} &             50.15 {\footnotesize $\pm$ 4.24} &            53.09 {\footnotesize $\pm$ 1.04} &             45.14 {\footnotesize $\pm$ 7.55} &            46.55 {\footnotesize $\pm$ 4.37} &             93.88 {\footnotesize $\pm$ 1.44} &             83.47 {\footnotesize $\pm$ 2.93} &  \bfseries{98.03 {\footnotesize $\pm$ 0.53}} &  \bfseries{100.0 {\footnotesize $\pm$ 0.0}} &  \bfseries{94.48 {\footnotesize $\pm$ 2.11}} \\
\midrule
SSM    &            53.82 {\footnotesize $\pm$ 3.12} &              57.72 {\footnotesize $\pm$ 7.0} &            52.79 {\footnotesize $\pm$ 3.16} &             45.75 {\footnotesize $\pm$ 7.24} &            48.82 {\footnotesize $\pm$ 4.34} &             58.98 {\footnotesize $\pm$ 5.48} &             67.86 {\footnotesize $\pm$ 11.4} &            57.27 {\footnotesize $\pm$ 13.73} &           79.43 {\footnotesize $\pm$ 24.29} &            67.13 {\footnotesize $\pm$ 20.31} \\
SSM-E  &            84.73 {\footnotesize $\pm$ 0.67} &             78.62 {\footnotesize $\pm$ 2.23} &  \bfseries{89.4 {\footnotesize $\pm$ 1.18}} &             74.69 {\footnotesize $\pm$ 2.76} &             69.8 {\footnotesize $\pm$ 2.72} &             97.88 {\footnotesize $\pm$ 0.66} &             95.57 {\footnotesize $\pm$ 0.97} &             96.44 {\footnotesize $\pm$ 0.83} &                                           - &                                            - \\
SSM-S  &            62.71 {\footnotesize $\pm$ 0.98} &             57.15 {\footnotesize $\pm$ 2.18} &             68.77 {\footnotesize $\pm$ 4.5} &             46.54 {\footnotesize $\pm$ 3.54} &            43.51 {\footnotesize $\pm$ 2.74} &             93.77 {\footnotesize $\pm$ 1.08} &             99.22 {\footnotesize $\pm$ 0.19} &             84.93 {\footnotesize $\pm$ 2.07} &  \bfseries{100.0 {\footnotesize $\pm$ 0.0}} &            81.99 {\footnotesize $\pm$ 21.79} \\
\midrule
VERA   &            55.95 {\footnotesize $\pm$ 2.68} &             73.97 {\footnotesize $\pm$ 2.63} &            67.39 {\footnotesize $\pm$ 2.57} &             37.27 {\footnotesize $\pm$ 4.66} &             46.29 {\footnotesize $\pm$ 8.1} &            78.11 {\footnotesize $\pm$ 21.05} &            67.53 {\footnotesize $\pm$ 21.63} &            76.22 {\footnotesize $\pm$ 22.11} &            94.63 {\footnotesize $\pm$ 7.22} &            45.66 {\footnotesize $\pm$ 10.55} \\
VERA-E &            76.66 {\footnotesize $\pm$ 3.23} &             73.41 {\footnotesize $\pm$ 6.68} &            81.31 {\footnotesize $\pm$ 3.92} &              83.6 {\footnotesize $\pm$ 7.45} &            78.52 {\footnotesize $\pm$ 7.98} &             85.8 {\footnotesize $\pm$ 15.18} &            88.52 {\footnotesize $\pm$ 13.91} &            79.58 {\footnotesize $\pm$ 15.61} &                                           - &                                            - \\
VERA-S &            61.37 {\footnotesize $\pm$ 0.74} &  \bfseries{85.02 {\footnotesize $\pm$ 2.38}} &            58.91 {\footnotesize $\pm$ 3.66} &             38.35 {\footnotesize $\pm$ 1.08} &            36.68 {\footnotesize $\pm$ 0.52} &  \bfseries{98.89 {\footnotesize $\pm$ 0.61}} &  \bfseries{99.64 {\footnotesize $\pm$ 0.53}} &             97.75 {\footnotesize $\pm$ 1.72} &            99.35 {\footnotesize $\pm$ 1.08} &             90.38 {\footnotesize $\pm$ 3.78} \\
\bottomrule
\end{tabular}
    }
\end{table*}

\begin{table*}
    \centering
    \caption{AUC-PR for OOD detection on the non-natural datasets when trained on the respective in-distribution dataset.}
    \label{tab:full_overall_table_other}
    \setlength{\tabcolsep}{1mm}
    \resizebox{\linewidth}{!}{%
    \begin{tabular}{lllllllllll}
\toprule
ID & \multicolumn{3}{c}{CIFAR-10} & \multicolumn{3}{c}{FMNIST} & \multicolumn{2}{c}{Segment} & \multicolumn{2}{c}{Sensorless} \\
\cmidrule(l){2-4} \cmidrule(l){5-7} \cmidrule(l){8-9} \cmidrule(l){10-11} 
OOD &                                     Constant &                                       Noise &                                    OODomain &                                     Constant &                                       Noise &                                    OODomain &                                    Constant &                                       Noise &                                    Constant &                                        Noise \\
\midrule
CE     &              45.26 {\footnotesize $\pm$ 8.8} &           61.13 {\footnotesize $\pm$ 21.02} &             30.69 {\footnotesize $\pm$ 0.0} &              35.5 {\footnotesize $\pm$ 3.08} &           55.84 {\footnotesize $\pm$ 22.32} &            30.74 {\footnotesize $\pm$ 0.11} &            42.57 {\footnotesize $\pm$ 18.3} &            33.82 {\footnotesize $\pm$ 3.14} &            32.42 {\footnotesize $\pm$ 1.04} &             31.96 {\footnotesize $\pm$ 1.28} \\
NF     &                                       30.87  &            83.65  &                                          \bfseries{100.0} &                                       71.07  &            98.04  &                                          \bfseries{100.0} &                                      99.97  &  \bfseries{100.0} &                           \bfseries{100.0 } &   \bfseries{100.0} \\
\midrule
CD   &            58.75 {\footnotesize $\pm$ 28.17} &  \bfseries{100.0 {\footnotesize $\pm$ 0.0}} &           58.41 {\footnotesize $\pm$ 37.96} &            70.59 {\footnotesize $\pm$ 12.84} &  \bfseries{100.0 {\footnotesize $\pm$ 0.0}} &  \bfseries{100.0 {\footnotesize $\pm$ 0.0}} &            96.13 {\footnotesize $\pm$ 2.55} &            95.43 {\footnotesize $\pm$ 3.58} &  \bfseries{100.0 {\footnotesize $\pm$ 0.0}} &   \bfseries{100.0 {\footnotesize $\pm$ 0.0}} \\
CD-E &  \bfseries{99.92 {\footnotesize $\pm$ 0.07}} &            87.5 {\footnotesize $\pm$ 24.31} &             30.69 {\footnotesize $\pm$ 0.0} &  \bfseries{96.63 {\footnotesize $\pm$ 7.53}} &            86.7 {\footnotesize $\pm$ 26.75} &             35.83 {\footnotesize $\pm$ 7.8} &                                           - &                                           - &                                           - &                                            - \\
CD-S &             31.32 {\footnotesize $\pm$ 0.29} &            98.01 {\footnotesize $\pm$ 0.93} &           70.86 {\footnotesize $\pm$ 30.57} &            71.17 {\footnotesize $\pm$ 10.16} &            97.79 {\footnotesize $\pm$ 1.02} &  \bfseries{100.0 {\footnotesize $\pm$ 0.0}} &            94.57 {\footnotesize $\pm$ 2.11} &             98.68 {\footnotesize $\pm$ 1.9} &            99.97 {\footnotesize $\pm$ 0.06} &             99.98 {\footnotesize $\pm$ 0.03} \\
\midrule
SSM    &            47.24 {\footnotesize $\pm$ 15.56} &           70.28 {\footnotesize $\pm$ 31.39} &            68.57 {\footnotesize $\pm$ 25.2} &            47.57 {\footnotesize $\pm$ 15.18} &           49.45 {\footnotesize $\pm$ 21.19} &            76.76 {\footnotesize $\pm$ 21.1} &           74.86 {\footnotesize $\pm$ 23.37} &           80.35 {\footnotesize $\pm$ 17.38} &            69.66 {\footnotesize $\pm$ 6.12} &            64.81 {\footnotesize $\pm$ 17.03} \\
SSM-E  &             65.64 {\footnotesize $\pm$ 3.66} &             64.5 {\footnotesize $\pm$ 4.05} &            42.74 {\footnotesize $\pm$ 6.73} &             82.51 {\footnotesize $\pm$ 4.95} &            87.54 {\footnotesize $\pm$ 2.17} &            98.49 {\footnotesize $\pm$ 1.94} &                                           - &                                           - &                                           - &                                            - \\
SSM-S  &              37.8 {\footnotesize $\pm$ 3.49} &           64.24 {\footnotesize $\pm$ 12.88} &             30.69 {\footnotesize $\pm$ 0.0} &             37.61 {\footnotesize $\pm$ 1.83} &            33.71 {\footnotesize $\pm$ 1.07} &           70.03 {\footnotesize $\pm$ 17.45} &           51.57 {\footnotesize $\pm$ 25.04} &           70.09 {\footnotesize $\pm$ 17.57} &            37.5 {\footnotesize $\pm$ 13.23} &             41.6 {\footnotesize $\pm$ 12.05} \\
\midrule
VERA   &             31.51 {\footnotesize $\pm$ 0.66} &  \bfseries{100.0 {\footnotesize $\pm$ 0.0}} &           63.48 {\footnotesize $\pm$ 34.37} &            53.24 {\footnotesize $\pm$ 22.65} &           79.34 {\footnotesize $\pm$ 27.34} &           72.42 {\footnotesize $\pm$ 37.61} &  \bfseries{100.0 {\footnotesize $\pm$ 0.0}} &  \bfseries{100.0 {\footnotesize $\pm$ 0.0}} &  \bfseries{100.0 {\footnotesize $\pm$ 0.0}} &             99.88 {\footnotesize $\pm$ 0.25} \\
VERA-E &             83.95 {\footnotesize $\pm$ 8.71} &            36.19 {\footnotesize $\pm$ 4.43} &             30.69 {\footnotesize $\pm$ 0.0} &            77.82 {\footnotesize $\pm$ 21.24} &           60.28 {\footnotesize $\pm$ 10.11} &            60.29 {\footnotesize $\pm$ 28.5} &                                           - &                                           - &                                           - &                                            - \\
VERA-S &              31.7 {\footnotesize $\pm$ 0.55} &           45.32 {\footnotesize $\pm$ 25.14} &  92.1 {\footnotesize $\pm$ 8.59} &             36.01 {\footnotesize $\pm$ 4.24} &            33.73 {\footnotesize $\pm$ 3.21} &  \bfseries{100.0 {\footnotesize $\pm$ 0.0}} &            99.01 {\footnotesize $\pm$ 1.45} &              99.9 {\footnotesize $\pm$ 0.3} &  \bfseries{100.0 {\footnotesize $\pm$ 0.0}} &  \bfseries{100.0 {\footnotesize $\pm$ 0.01}} \\
\bottomrule
\end{tabular}
    }
\end{table*}

\begin{table*}
    \centering
    \caption{AUC-PR for OOD detection of EBMs with different dimensions of the bottleneck introduced into the WideResNet-10-2 architecture.}
    \label{tab:bottleneck_full}
    \setlength{\tabcolsep}{1mm}
    \resizebox{\linewidth}{!}{%
    \begin{tabular}{lcrrrrrrrrrrrrrr}
\toprule
     & ID & \multicolumn{8}{c}{CIFAR-10} & \multicolumn{6}{c}{FashionMNIST} \\
\cmidrule(l){3-10} \cmidrule(l){11-16} 
     & OOD &     SVHN &   LSUN & CelebA & CIFAR-100 & Textures &  Noise & OODomain & Constant &       KMNIST &  MNIST & NotMNIST &  Noise & OODomain & Constant \\
Model & Bottleneck &          &        &        &           &          &        &          &          &              &        &          &        &          &          \\
\midrule
\multirow{4}{*}{CD} & 0.05 &    81.29 &  42.16 &  42.82 &     53.09 &    59.13 &  100.0 &    76.77 &     78.4 &        65.27 &  43.26 &    80.72 &  100.0 &    100.0 &    73.18 \\
     & 0.10 &    67.93 &   48.3 &  52.71 &     53.47 &    61.99 &  100.0 &    62.19 &    70.85 &         73.8 &  53.23 &    83.25 &  100.0 &    100.0 &    78.27 \\
     & 0.20 &    77.97 &   43.8 &  42.19 &     51.85 &    63.53 &  100.0 &    85.99 &    79.71 &        73.46 &   51.8 &     81.2 &  99.99 &    100.0 &    64.49 \\
     & 1.00 &    60.72 &  54.43 &  43.86 &     50.51 &    76.21 &  100.0 &    58.41 &    58.75 &        50.52 &  31.69 &    76.85 &  100.0 &    100.0 &    70.59 \\
\cline{1-16}
\multirow{4}{*}{SSM} & 0.05 &     53.6 &  49.44 &  54.06 &     53.47 &     43.3 &  51.57 &     89.8 &     67.7 &        56.79 &  62.09 &    49.05 &  42.48 &     77.2 &    46.42 \\
     & 0.10 &    52.51 &   49.4 &   57.4 &     51.54 &    40.74 &  40.81 &    87.05 &    62.91 &        58.32 &  69.04 &    48.75 &  46.52 &    65.72 &    42.82 \\
     & 0.20 &    52.69 &  49.29 &  59.96 &     52.37 &    49.31 &  49.26 &    71.39 &    57.82 &        62.31 &  53.76 &     68.8 &  67.07 &    65.84 &    56.88 \\
     & 1.00 &    45.75 &  52.79 &  57.72 &     53.82 &    48.82 &  70.28 &    68.57 &    47.24 &        58.98 &  67.86 &    57.27 &  49.45 &    76.76 &    47.57 \\
\cline{1-16}
\multirow{4}{*}{VERA} & 0.05 &    33.34 &  89.65 &  82.69 &     55.22 &    58.13 &  72.28 &    86.14 &    30.92 &        87.52 &  80.64 &    75.22 &  74.43 &    100.0 &    33.09 \\
     & 0.10 &    34.52 &  80.64 &   79.4 &     54.44 &    54.84 &  45.02 &    84.56 &    33.65 &        86.85 &  85.65 &     66.3 &  75.58 &    96.28 &    31.54 \\
     & 0.20 &    33.95 &  73.29 &   76.1 &     54.03 &    45.01 &  48.29 &    73.57 &    31.31 &         88.7 &  89.06 &     60.7 &  59.56 &    100.0 &    36.01 \\
     & 1.00 &    37.27 &  67.39 &  73.97 &     55.95 &    46.29 &  100.0 &    63.48 &    31.51 &        78.11 &  67.53 &    76.22 &  79.34 &    72.42 &    53.24 \\
\bottomrule
\end{tabular}
    }
\end{table*}

\clearpage

\section{Other directions}
\label{ch:appendix:sec:other_directions}
In addition to the work in the main thesis, we pursued several other research directions which do not fit into the existing thesis structure. For completeness, we report the ideas and results in the following. 

\subsection{Noise Contrastive Estimation for Energy-based model training}
There are several other methods proposed in the literature for training \glspl{ebm} based on contrastive learning. \gls{nce}~\cite{gutmannNoisecontrastiveEstimationNew} uses a noise distribution for training the \gls{ebm}. However to be effective, \gls{nce} relies on the noise distribution being close to the data distribution. Obviously, this is problematic since we want to estimate the data distribution in the first place.

Instead, Conditional \gls{nce}~\cite{ceylanConditionalNoiseContrastiveEstimation2018} conditions the noise distribution on the data samples, thus reducing the amount of assumptions necessary to obtain a noise distribution similar to the data distribution. 

We train \glspl{ebm} with Conditional \gls{nce} with the noise distribution set to a Gaussian \(N(\mathbf{x} \mid \sigma \mathbb{I}_D) \) centered at the datapoint \(\mathbf{x} \in \mathbb{R}^D\) on the CIFAR-10 dataset. 

In \Cref{ch:appendix:sec:other_directions:tab:conditional_nce_ood}, we report \gls{ood} detection results for different settings of \(\sigma\). We find that Conditional \gls{nce} does not yield \glspl{ebm} with satisfying \gls{ood} detection results, especially on SVHN, which are comparable to other \gls{ebm} training methods on high-dimensional datasets. Therefore, we omitted \gls{nce} and Conditional \gls{nce} in further experiments. 

\begin{table}[ht]
\caption{AUC-PR for \gls{ood} detection with \glspl{ebm} trained with conditional \gls{nce}.}
\label{ch:appendix:sec:other_directions:tab:conditional_nce_ood}
\centering
\begin{tabular}{@{}lrrrr@{}}
\toprule
  & SVHN  & LSUN  & Noise  & OODomain \\ \midrule
Empirical variance & 16.49 & 68.21 & 100.00 & 100.00   \\
$\sigma = 1.0$            & 17.68 & 67.68 & 100.00 & 100.00   \\
$\sigma = 0.1$            & 19.52 & 82.43 & 100.00 & 99.08    \\
$\sigma = 0.0001$         & 22.95 & 64.24 & 100.00 & 99.83    \\
$\sigma = 10^{-6}$        & 31.53 & 54.84 & 93.79  & 24.61    \\ \bottomrule
\end{tabular}%
\end{table}

\subsection{Role of transformation in Normalizing Flows for out-of-distribution detection}
In addition to the work in \Cref{ch:results:sec:ood_detection_ebms}, we also investigate certain hypotheses potentially influencing the \gls{ood} detection with Normalizing Flows. In particular, we consider dimensionality reduction and the expressiveness of transformations in Normalizing Flows in the following.

\paragraph{Dimensionality reduction.}
To investigate whether the gap in \gls{ood} detection between \glspl{ebm} and Normalizing Flows is caused by \glspl{ebm} being able to project to lower dimensions while Normalizing Flows operate in the original data space \(\mathbb{R}^D\), we considered \(\mathcal{M}\)-Flows~\cite{brehmerFlowsSimultaneousManifold2020}. \(\mathcal{M}\)-Flows are able to simultaneously learn a lower-dimensional manifold of the data and a density on that manifold with a Normalizing Flow. 

For this, \citet{brehmerFlowsSimultaneousManifold2020} introduce a diffeomorphism between the latent space and data space as

\begin{align}
    f: U \times V \mapsto X, \qquad \mathbf{u}, \mathbf{v} \rightarrow f(\mathbf{u}, \mathbf{v})
\end{align}

They then define the manifold \(\mathcal{M}\) as a level set

\begin{align}
    g: U \mapsto \mathcal{M} \subset X, \qquad \mathbf{u} \rightarrow g(\mathbf{u}) = f(\mathbf{u}, \mathbf{0})
\end{align}

For training, they minimize the reconstruction error \(\| \data - g(g^{-1}(\data)) \|\) to learn the manifold while training a Normalizing Flow on the subspace \(U\) with maximum likelihood.

We train \(\mathcal{M}\)-Flow on the CIFAR-10 dataset with latent dimension \(\dim(U) = 128\) and report \gls{ood} detection results using the density \(p_{\params}(\mathbf{x})\) and reconstruction error \(\| \data^\prime - \data \|\) in \Cref{ch:appendix:sec:other_directions:tab:mflow_ood}. 

Comparing with a vanilla Normalizing Flow, which does not leverage dimensionality reduction before density estimation, in \Cref{ch:results:sec:ood_detection_ebms:tab:all_results}, we observe no significant improvements. This further validates our hypotheses in \Cref{ch:results:sec:ood_detection_ebms:subsec:ebms_better_than_nf} that the ability of a model to perform dimensionality reduction does not necessarily directly lead to improvements on the task of \gls{ood} detection.

\begin{table}
\caption{AUC-PR for OOD detection with \(\mathcal{M}\)-Flow}
\label{ch:appendix:sec:other_directions:tab:mflow_ood}
\centering
\begin{tabular}{@{}lrrrrrr@{}}
\toprule
Score      & CIFAR-100 & CelebA & LSUN & SVHN & Textures & Noise \\ \midrule
\(p_{\params}(\mathbf{x})\)       & 53.2      & 61     & 75.2 & 22.5 & 50       & 100   \\
\(\| x^\prime - x \|\) & 49.6      & 51.7   & 71.4 & 15.5 & 45.7     & 100   \\ \bottomrule
\end{tabular}%
\end{table}

\paragraph{Expressiveness of transformations.}
Another advantage of \glspl{ebm} over Normalizing Flows is that \glspl{ebm} are not restricted in their transformations. As introduced in \Cref{ch:background:sec:density_estimation:subsec:normalizing_flow}, Normalizing Flows rely on invertible transformations where the determinant of the Jacobian \(\frac{\partial \mathbf{f}}{\partial \mathbf{z}}\) is easy to compute, reducing the expressiveness of individual transformations. We hypothesize that the lower expressiveness reduces the capability of Normalizing Flows to fit the true density, in turn worsening \gls{ood} detection.

We investigate the hypotheses by comparing with Normalizing Flow variants which have less restrictions on individual transformations. \citet{behrmannInvertibleResidualNetworks2019} propose \textit{iResNet} introducing a ResNet architecture~\cite{heDeepResidualLearning2015} with invertible transformations. In particular, iResNet contains general transformations of ResNets, while invertibility is enforced during training. 

Additionally, we propose a variant of Normalizing Flows where we add a regularization term on the weight matrices $W \in \mathbb{R}^{D\times D}$ to be orthogonal 

\begin{equation}
L_{reg} = \| \mathbb{I}_D - W^TW \|_F
\end{equation}

where \(\| \cdot \|_F\) denotes the Frobenius norm. This is based on a similar regularization in \citet{qiPointNetDeepLearning2017}. It holds that \(W \text{ is orthogonal} \Leftrightarrow L_{reg} = 0\). Thus, the Jacobian determinant is given as $| \det (\frac{df}{dz}) | \approx 1$ given \(L_{reg} \approx 0\) for $f(x) = Wx + b$ and $f^{-1}(z) = W^Tz - b$. We call this variant \textit{Approximate Flow} since it does not provide exact likelihoods depending on the convergence of the regularization term \(L_{reg}\). In this model, the individual transformations are also freely specified, however, during training the weight matrices are pushed to be orthogonal. We introduce leaky ReLU activations as non-linearities  in between the layers since they are invertible.  

We compare against traditional Normalizing Flow models where the transformations are defined in a way such that \(\det (\frac{df}{d\mathbf{z}})\) is tractable (\textit{Radial}, \textit{IAF}, \textit{Planar})~\cite{rezendeVariationalInferenceNormalizing2016, kingmaImprovingVariationalInference2017}. Further, we consider the two Normalizing Flow models above where the transformations are freely specified (\textit{iResNet}, \textit{Approx. Flow}). For all models, we consider \(20\) stacked transformations.

In \Cref{ch:appendix:sec:other_directions:tab:log_likelihoods_toy_dataset}, we report the log-likelihood of the test set under the different models. We observe that there is no significant difference between models with restricted and free transformations. In some cases the models with restricted transformations even outperform non-restricted models.

We conclude that the expressiveness of individual transformations is not significant when the overall network is sufficiently deep. 

\begin{table}
\caption{Log-likelihood of Normalizing Flow models on different datasets. Rows with $^*$ denote log-likelihoods estimated with numerical integration.}
\label{ch:appendix:sec:other_directions:tab:log_likelihoods_toy_dataset}
\centering
\begin{tabular}{@{}lrrrr@{}}
\toprule
Model                & 3-Gaussians & Two Moons & Checkerboard & Swiss Roll \\ \midrule
NF (Radial)          & -2.16       & -2.53     & -4.23        & -4.22      \\
NF (IAF)             & -0.88       & -2.42     & -3.76        & -3.69      \\
NF (Planar)          & -1.34       & -2.42     & -3.77        & -3.30      \\
\midrule
iResNet              & -1.34       & -2.53     & -3.95        & -4.18      \\
Approx. Flow$^*$         & -3.13       & -3.34    & -4.51        & -5.35      \\ \bottomrule
\end{tabular}%
\end{table}

\subsection{Generating low-entropy samples using SGLD}

\citet{heinWhyReLUNetworks2019} attempted to fix the issue of overly confident predictions in ReLU networks by adding an additional loss term enforcing high-entropy predictions for samples randomly obtained from a noise distribution covering the domain of the classifier, e.g., \(\mathcal{U}(-1, 1)\). We deem this approach to be inefficient since it causes the classifier to be evaluated even in regions outside that training data which might already have high-entropy predictions.  

Thus, we propose to extend this approach by guiding the sampling into low-entropy regions.
For this, we define the entropy function \(\mathcal{H}\) over the domain of the classifier \(p_{\params}(y \mid \mathbf{x})\) as 

\begin{equation}
    \mathcal{H}(p_\theta(y|x)) = - \sum_y p_\theta(y|x) \log p_\theta(y|x) 
\end{equation}

We then leverage \gls{sgld} to obtain low-entropy samples from \(\mathcal{H}\)

\begin{equation}
    x_0 \sim p_0(x), \qquad x_{i+1} = x_i - \frac{\alpha}{2} \frac{\partial H_\theta}{\partial x_i} + \epsilon, \qquad \epsilon \sim N(0, \alpha) 
\end{equation}

similar to what we introduced in \Cref{ch:background:sec:density_estimation:subsec:ebm} for \gls{ebm} training. For our experiments we consider \(p_0 = \mathcal{U}(-10, 10)\).

We then follow \citet{heinWhyReLUNetworks2019} and maximize the entropy at these positions

\begin{equation}
    \max_\theta \mathbb{E}_{\mathbf{x} \sim \mathcal{H}} - \sum_y p_{\params}(y|\mathbf{x}) \log p_{\params}(y|\mathbf{x})
\end{equation}

as additional loss term.

\begin{figure}[ht]
    \centering
    \begin{subfigure}[t]{0.3\linewidth}
        \centering
        \includegraphics{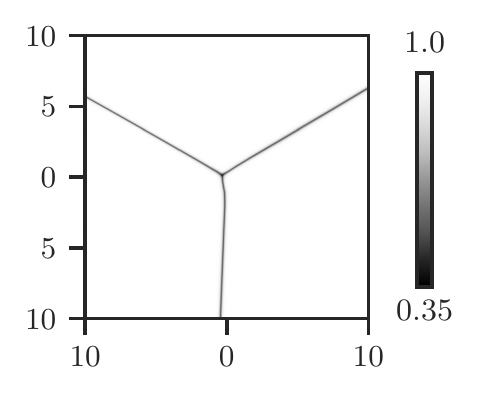}
        \caption{}
        \label{ch:appendix:sec:max_ent:subfig:baseline}
    \end{subfigure}%
    \hfill
    \begin{subfigure}[t]{0.3\linewidth}
        \centering
        \includegraphics{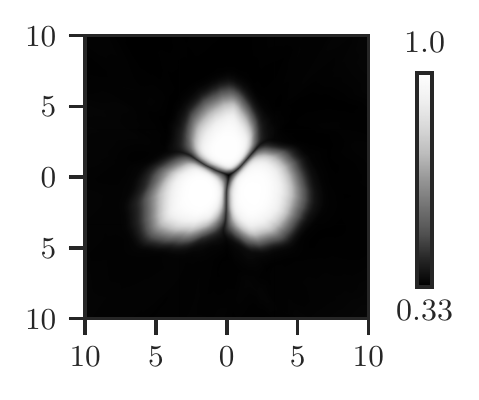}
        \caption{}
        \label{ch:appendix:sec:max_ent:subfig:max_ent_-10_10}
    \end{subfigure}%
    \hfill
    \begin{subfigure}[t]{0.3\linewidth}
        \centering
        \includegraphics{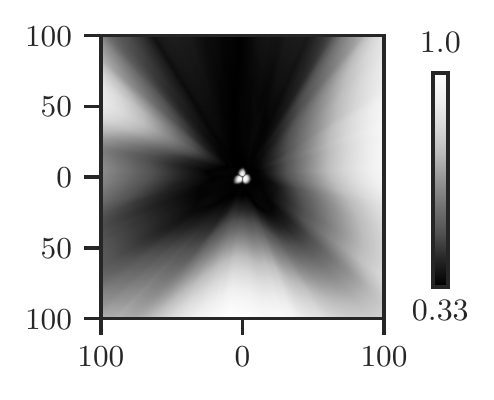}
        \caption{}
        \label{ch:appendix:sec:max_ent:subfig:max_ent_-1000_1000}
    \end{subfigure}%
    \caption{(a) Confidence of the cross-entropy model. (b, c) Confidence on two domains for the model trained with additional entropy objective.}
    \label{ch:appendix:sec:max_ent:fig:confidence}
\end{figure}

From \Cref{ch:appendix:sec:max_ent:fig:confidence}, we observe that indeed this additional term can ensure that the model makes confident predictions only close to the training data. However, when considering out-of-domain inputs as it can be seen in \Cref{ch:appendix:sec:max_ent:subfig:max_ent_-1000_1000}, the confidence again tends to \(1\) for samples far from the training data resulting in possibly confident wrong predictions.

\microtypesetup{protrusion=false}
\listoffigures{}
\listoftables{}
\microtypesetup{protrusion=true}
\printglossaries
\printbibliography{}

\end{document}